\documentclass[mnsc,nonblindrev]{informs3-VVM}

\SingleSpacedXI

\usepackage[colorinlistoftodos, textwidth=\marginparwidth, textsize = footnotesize]{todonotes}

\usepackage{natbib}
 \bibpunct[, ]{(}{)}{,}{a}{}{,}%
 \def\bibfont{\small}%

 \usepackage{rotating}

\usepackage{float}
\usepackage{amsmath}
\usepackage{amssymb}
\usepackage{mathtools}
\usepackage{tikz}
\usepackage{algorithm}
\usepackage[noend]{algpseudocode}
\usetikzlibrary{calc,arrows.meta,positioning}
\usepackage{arydshln}
\usepackage{subcaption}
\usepackage{xcolor}
\usepackage{forest}
\usepackage{siunitx}
\usepackage{booktabs}
\usepackage{pgfplots}
\usepackage{ctable}
\usepackage{multirow}
\pgfplotsset{compat=newest}

\theoremstyle{definition}
\newtheorem{exmp}{Example}

\def\Ab{\mathbf{A}}

\def\ob{\mathbf{o}}
\def\Pb{\mathbf{P}}
\def\rb{\mathbf{r}}
\def\ub{\mathbf{u}}
\def\vb{\mathbf{v}}
\def\wb{\mathbf{w}}
\def\yb{\mathbf{y}}
\def\zb{\mathbf{z}}

\def\zerob{\mathbf{0}}
\def\oneb{\mathbf{1}}

\def\Ncal{\mathcal{N}}
\def\Scal{\mathcal{S}}
\def\Tcal{\mathcal{T}}

\def\Ibb{\mathbb{I}}

\def\alphab{\boldsymbol \alpha}
\def\lambdab{\boldsymbol \lambda}
\def\xib{\boldsymbol \xi}
\def\epsilonb{\boldsymbol \epsilon}

\def\mub{\boldsymbol \mu}
\def\psib{\boldsymbol \psi}
\def\Deltab{\boldsymbol \Delta}
\def\Rb{\mathbf{R}}

\def\Fcal{\mathcal{F}}
\def\Ycal{\mathcal{Y}}
\def\Pbb{\mathbb{P}}
\def\Rbb{\mathbb{R}}
\def\Exp{\mathbb{E}}

\def\ispreferredto{\succ}

\def\splitterset{\Xi}
\def\condsetsize{\Gamma}

\def\NumLeaves{L}
\def\Depth{\mathrm{Depth}}
\def\leaves{\mathbf{leaves}}
\def\splits{\mathbf{splits}}

\def\EstLO{\textsc{EstLO}}
\usepackage{color}

\def\CSsize{\tau}

\TheoremsNumberedThrough     %
\ECRepeatTheorems

\EquationsNumberedThrough    %

\MANUSCRIPTNO{MS-OPT-19-01028.R1}

\begin{document}

\RUNAUTHOR{Chen and Mi\v{s}i\'{c}}

\RUNTITLE{Decision Forest: A Nonparametric Approach to Modeling Irrational Choice}

\TITLE{Decision Forest: A Nonparametric Approach to Modeling Irrational Choice}

\ARTICLEAUTHORS{%
\AUTHOR{Yi-Chun Chen}
\AFF{UCLA Anderson School of Management, University of California, Los Angeles, California 90095, United States, \EMAIL{yi-chun.chen.phd@anderson.ucla.edu}}
\AUTHOR{Velibor V. Mi\v{s}i\'{c}}
\AFF{UCLA Anderson School of Management, University of California, Los Angeles, California 90095, United States, \EMAIL{velibor.misic@anderson.ucla.edu}} %

} %

\ABSTRACT{%
Customer behavior is often assumed to follow weak rationality, which implies that adding a product to an assortment will not increase the choice probability of another product in that assortment. However, an increasing amount of research has revealed that customers are not necessarily rational when making decisions. In this paper, we propose a new nonparametric choice model that relaxes this assumption and can model a wider range of customer behavior, such as decoy effects between products. In this model, each customer type is associated with a binary decision tree, which represents a decision process for making a purchase based on checking for the existence of specific products in the assortment. Together with a probability distribution over customer types, we show that the resulting model -- a {\it decision forest} -- is able to represent {\it any} customer choice model, including models that are inconsistent with weak rationality. We theoretically characterize the depth of the forest needed to fit a data set of historical assortments and prove that with high probability, a forest whose depth scales logarithmically in the number of assortments is sufficient to fit most data sets. We also propose two practical algorithms -- one based on column generation and one based on random sampling -- for estimating such models from data. Using synthetic data and real transaction data exhibiting non-rational behavior, we show that the model outperforms both rational and non-rational benchmark models in out-of-sample predictive ability.
}%

\HISTORY{First version: April 22, 2019. Second version: May 18, 2020. This version: July 25, 2021. Forthcoming in \emph{Management Science}.}

\maketitle

\section{Introduction}
\label{sec:intro}

A common problem in business is to decide which products to offer to customers by using historical sales data. The problem can be generally stated as follows: a firm offers a set of products ({\it an assortment}) to a group of customers. Each customer makes a decision to either purchase one of the products or not purchase any of the products. The goal of the firm is to decide which products to offer, so as to maximize the expected revenue when customers exercise their preferences.

In order to make such decisions, it is critical to have access to a model for predicting customer choices. %
{\it Customer choice models} have been used to model and predict the substitution behavior of customers when they are offered different assortments of products. In general, a choice model can be thought of as a conditional probability distribution over all purchase options given an assortment that is offered. A rich literature spanning marketing, psychology, economics, and operations management has contributed to the understanding of choice models. 

A widely-used assumption is that customers are rational, i.e., choice models are assumed to follow {\it rational choice theory} and are based on the {\it random utility maximization} (RUM) principle. The RUM principle requires that each product is endowed with a stochastic utility. When a customer encounters the assortment and needs to make a purchase decision, all utilities are realized and the customer will choose the product from the assortment with the highest realized utility. A consequence of the RUM principle is that whenever we add a product to an assortment, the choice probability of each incumbent product either stays the same or decreases. This property is known as \emph{regularity} or \emph{weak rationality}. 

However, customers are not always rational. There is an increasing body of experimental evidence, arising in the fields of marketing, economics, and psychology, which suggests that the aggregate choice behavior of individuals is not always consistent with the RUM principle and often violates the weak rationality property. A well-known example is the experiment involving subscriptions to \emph{The Economist} magazine from \cite{ariely2008predictably}, which is re-created in Table \ref{tb:economist}. One hundred MIT students were asked to make decisions given two different assortments of subscription options. In the first assortment in Table~\ref{tb:economist}, two subscription options are given: ``Internet-Only'' (\$59.00) and ``Print-\&-Internet'' (\$125.00). The first option is chosen by the majority of the students (68 out of 100). In the second assortment, the students are given one more option: ``Print-Only'' (\$125.00). For this second assortment, due to the obvious advantage in ``Print-\&-Internet'' over ``Print-Only'', no one chose the latter option. But with the addition of the the ``Print-Only'' option, the number of subscribers of the ``Print-\&-Internet'' option actually \emph{increased} from $32$ to $84$, thus demonstrating a violation of the weak rationality property. Here, the option ``Print-Only'' serves as a decoy or an anchor: its presence can influence an individual's preference over the two other options ``Internet-Only'' and  ``Print-\&-Internet''. {While this example comes from a classroom experiment, there has been an extensive peer-reviewed research literature on this phenomenon, known as the \emph{decoy} or \emph{attraction} effect, since the seminal work of \cite{huber1982adding}.
	
	\begin{table}[]
		\caption{Behavioral experiment involving subscriptions to {\it The Economist}; reproduced from \cite{ariely2008predictably}}
		\label{tb:economist}
		\centering
		\small
		\begin{tabular}{ccc}
			\toprule
			Option          & Price & Num. of Subscribers  \\ \midrule
			Internet-Only          & \$59.00    & 68                  \\
			Print-\&-Internet         & \$125.00   & 32                  \\ \bottomrule
			&       &                     \\ \toprule
			Option          & Price & Num. of Subscribers \\ \midrule
			Internet-Only          & \$59.00    & 16                  \\
			Print-Only           & \$125.00   & 0                   \\
			Print-\&-Internet & \$125.00   & 84                 \\ \bottomrule
		\end{tabular}
	\end{table}

	The example that we have described above is important for two reasons. First, even for this very simple example, no choice model based on RUM can perfectly capture the subscribers' observed behaviors; as such, choice predictions based on RUM models will be inherently biased if customers do not behave according to a RUM model. Second, the presence of irrationality in customer choice behavior can have significant operational implications on which products should be offered. As a concrete example, observe that in Table \ref{tb:economist}, assuming that customers have no outside option, the expected per-customer revenue arising from the first assortment is \$80.12, whereas the expected per-customer revenue of the second assortment is \$114.44 -- an increase of more than 40\%! Indeed, outside of experimental settings (as in the above example), deviations from rational behavior have been observed -- and exploited -- in business practice. For example, when Williams-Sonoma observed low sales of a bread bakery machine priced at \$275, it introduced a larger and more expensive version priced at \$429; few customers bought the new model, but sales of the original model almost doubled \citep{poundstone2010priceless}.\footnote{We gratefully acknowledge the paper of \cite{golrezaei2014real} for bringing this example to our attention.} %

	In this paper, we propose a new type of choice model, called the \emph{decision forest} model, that is flexible enough to model non-rational choice behavior, i.e., choice behavior that is inconsistent with the RUM principle. In this choice model, one assumes that the customer population can be described as a finite collection of customer types, where each customer type is associated with a binary decision tree, together with a probability distribution over those types. Each decision tree defines a sequence of queries that the customer follows in order to reach a purchase decision, where each query involves checking whether a particular product is contained in the assortment or not. 
	
	We make the following specific contributions:
	
	\begin{enumerate}
		
		\item \textbf{Model}: We propose a new model for customer choice based on representing the customer population as a probability distribution over decision trees. We provide several examples of how well-known behavioral anomalies, such as the decoy effect and the preference cycle, can be represented by this model. We also prove a key theoretical result: \emph{any choice model}, whether it obeys the RUM property or not, can be represented as a probability distribution over binary decision trees. As a result, our model can be regarded as a nonparametric model for general choice behavior. 
		
		\item \textbf{Model complexity guarantees}: We consider the problem of how to estimate our forest model from data and establish two guarantees on the complexity of the trees required to learn the model. Our first result states that for any data set of $M$ historical assortments for $N$ products there exists a decision forest model consisting of $M(N+1)+1$ trees, each of depth at most $\min\{M+1,N+1\}$ and with at most $2M$ leaves, that perfectly fits the data. Our second result states that with very high probability over the sample of $M$ historical assortments, forests consisting of trees of depth scaling logarithmically with the number of assortments $M$ are sufficient to perfectly fit the data. Thus, when data is limited, we can use less complex (simpler) models to fit the data. 
		
		\item  \textbf{Estimation methods}: We formulate the problem of estimating the decision forest model from data as an optimization problem and propose two solution methods. The first method, based on column generation, involves sequentially adding new trees to a growing collection. While this approach guarantees optimality when the column generation subproblem is solved exactly as an integer program, it is not computationally scalable. We thus propose a top-down learning algorithm for heuristically solving the subproblem, leading to a heuristic column generation approach. The second method, called randomized tree sampling, is based on randomly sampling a large number of trees and then finding the corresponding probability distribution over these trees by solving an optimization problem. This method removes the computational effort needed to search for decision trees by generating them through a simple and efficient randomization scheme. We provide a theoretical result to justify the usage of this method showing that the training error of the obtained model is bounded by the error of a model defined relative to the sampling distribution plus a term that decays with rate $1/\sqrt{K}$, where $K$ is the number of sampled trees.

		\item \textbf{Practical performance}: We evaluate the performance of our proposed model using real sales data from the IRI Academic Data Set \citep{bronnenberg2008database} and compare to other methods in the literature. We show that decision forest models lead to a significant improvement in out-of-sample prediction, as measured by Kullback-Leibler divergence, over the multinomial logit (MNL) model, the latent-class MNL model, the ranking-based model, and the HALO-MNL model \citep{maragheh2018customer} across a large range of product categories. We also demonstrate how the decision forest model can be used to extract insights about substitution and complementarity effects and identify interesting customer behaviors within a specific product category.
	\end{enumerate}
	
	The rest of this paper is organized as follows. In Section~\ref{sec:literature_review}, we review the relevant literature in rational and non-rational choice modeling. In Section~\ref{sec:model}, we present our decision forest model and theoretically characterize its expressive power. In Section~\ref{sec:simple_trees}, we present our theoretical results on model complexity. In Section~\ref{sec:model_estimation_methods}, we present our two estimation methods. In Section~\ref{sec:experiment_IRI}, we numerically show the effectiveness of our approach on real-world data. In Section~\ref{sec:conclusions}, we conclude. All proofs are provided in the electronic companion, along with additional numerical experiments based on real and synthetic data.

\section{Literature Review}
\label{sec:literature_review}

In this section, we review the relevant literature. We first review prior work in rational choice modeling (Section~\ref{subsec:literature_review_rational}), followed by prior research in non-rational choice modeling (Section~\ref{subsec:literature_review_nonrational}). As a key contribution of this paper is the universality property of the decision forest model (see Theorem~\ref{thm:universal_choice_model} in Section~\ref{subsec:model_universal}), in Section~\ref{subsec:literature_review_universality}, we review two classes of choice models that also share this property and compare them with our model. Finally, we relate our proposed model to other research areas in Section~\ref{subsec:literature_review_other}. 

\subsection{Rational Choice Modeling}
\label{subsec:literature_review_rational}

Numerous discrete choice models have been proposed based on the RUM principle, such as the multinomial logit (MNL), latent-class MNL (LC-MNL), and nested logit (NL) model; we refer the reader to \cite{ben1985discrete} and \cite{train2009discrete} for more details.

There has been a significant effort to develop ``universal'' choice models. A well-known universality result in choice modeling comes from the paper of \cite{mcfadden2000mixed}, which showed that any RUM choice model can be approximated to an arbitrary precision by a mixture of MNL models. Outside of logit models, earlier research proposed the ranking-based model (also known as the \emph{stochastic preference} model), in which one represents a choice model as a probability distribution over rankings; \cite{block1959random} showed that the class of RUM choice models is equivalent to the class of ranking-based models. Later, the seminal paper of \cite{farias2013nonparametric} developed a data-driven approach for making revenue predictions via the ranking-based model; specifically, the method involves computing the worst-case revenue of a given assortment over all ranking-based models that are consistent with the available choice data. Subsequent research on ranking-based models has studied other estimation approaches \citep{van2014market,misic2016data,jagabathula2016nonparametric,jagabathula2018limit}, as well as methods for obtaining optimal or near-optimal assortments \citep{aouad2015assortment,aouad2018approximability,feldman2018assortment,bertsimas2017exact}. Another model is the Markov chain model of customer choice \citep{blanchet2016markov}. By modeling substitution behavior between products as transitions between states in the Markov chain, the model provides a good approximation to any choice model based on the RUM principle. Since the original paper of \cite{blanchet2016markov}, later research has considered other methods of estimating such models from limited data \citep{simsek2018expectation} as well as methods for solving core revenue management problems under such models \citep{feldman2017revenue,desir2015capacity,desir2015robust}. In a different direction, the papers of \cite{natarajan2009persistency} and \cite{mishra2014theoretical} proposed the marginal distribution model, which is the choice model obtained by finding the joint distribution of errors in the random utility model that is consistent with given marginal error distributions and maximizes the customer's expected utility.

As discussed before, there exist many experimental and empirical examples of choice behavior that deviates from RUM, and thus cannot be modeled using the ranking-based model. Our model, on the other hand, is general enough to represent models that cannot be represented using RUM. Our model can be regarded as a natural extension of the nonparametric model of \cite{farias2013nonparametric} to the realm of non-rational choice models.

\subsection{Non-rational Choice Modeling}
\label{subsec:literature_review_nonrational}

The study of non-rational choice has its roots in the seminal work of \cite{kahneman1979prospect}, which demonstrated how expected utility theory fails to explain certain choice phenomena, and proposed prospect theory as an alternative model. Since this paper, significant research effort has been devoted to the study of non-rational decision making. Within this body of research, our work relates to the significant empirical and theoretical work in behavioral economics on context-dependent choice \citep{tversky1993context}, which includes important context effects such as the compromise effect \citep{simonson1989choice}, the attraction effect \citep{huber1982adding}, and the similarity effect \citep{tversky1972elimination}.

Recently, new choice models have been proposed for modeling behavior outside of the RUM class. Within behavioral economics, examples include the generalized Luce model \citep{echenique2015general} and the perception-adjusted Luce model (PALM) \citep{echenique2018perception}. The focus of these papers is descriptive, in that they develop axiomatic theories for new models. In contrast, the focus of our paper is prescriptive, as we develop optimization-based methods for estimating our decision forest models from limited data.

Within operations management, examples of new choice models include the general attraction model (GAM) \citep{gallego2014general}, the HALO-MNL model \citep{maragheh2018customer} and the generalized stochastic preference (GSP) model \citep{berbeglia2018generalized}. The main difference between our model and these prior models is in expressive power. As we will show in Section~\ref{subsec:model_universal}, our choice model is universal and is able to represent \emph{any} discrete choice model that may or may not be in the RUM class; in contrast, for each of the generalized Luce model, PALM, GAM, HALO-MNL model and GSP model, there either exist choice models that do not obey the RUM principle and cannot be represented by the model, or the representational power of the model is unknown.

Lastly, the recent paper of \cite{jagabathula2018limit} introduced the concept of {\it loss of rationality}. The loss of rationality is defined as the lowest possible information loss attained by fitting an RUM model to a given choice dataset; in other words, it is the lack of fit between the given data and the entire RUM class. By computing this lack of fit, one can determine whether it is necessary to consider models outside of the RUM class. The focus of our paper is different, in that we propose a specific answer to the question of how one should model customer behavior that may be irrational, and as such is complementary to that of \cite{jagabathula2018limit}.

\subsection{Previous Universality Results}
\label{subsec:literature_review_universality}
	
Within the economics literature, there are two classes of non-rational models that have the same expressive power as the decision forest model. We review these two classes of models in detail, followed by our contribution to this ``universality'' paradigm.

The first is the class of game tree models and randomized game tree models. The game tree model was proposed by \cite{xu2007rationalizability} as a model for how an option is deterministically chosen from a given choice set. In a game tree model, the tree encodes a hierarchy where each leaf corresponds to a product, and each non-leaf node correspond to a decision maker that is endowed with a ranking over the product universe. To make a decision, one starts at the non-leaf nodes whose children are leaves, and the decision maker at each such non-leaf node chooses its most preferred product according to its ranking. The parent nodes of those non-leaf nodes then choose from the products chosen by their children. This process repeats until reaching the decision maker at the root of the tree who makes the final decision. The model can thus be thought of as a representation of how an organization, through several rounds of decision making, reaches a decision. This type of model bears a superficial similarity to ours in that both models involve trees. However, the details differ significantly: in our trees, the decision process starts at the root (rather than the leaves) and involves checking for the existence/non-existence of a product in the assortment, until reaching a leaf. In addition, the trees that we describe are always binary, whereas game trees can in general be non-binary trees. Since the paper of \cite{xu2007rationalizability}, other research has extended this type of model in different ways. For example, \cite{horan2011choice} considers game trees where all of the decision makers follow the same ranking. In the subsequent literature, the paper of \cite{li2017every} considers the randomized game tree model, where one assumes a probability distribution over the tuple of rankings for the non-leaf nodes; that paper shows that this model can represent any discrete choice model, which is similar to our universality result (Theorem~\ref{thm:universal_choice_model} in Section~\ref{subsec:model_universal}).

The second type of model that has the universality property is the pro-con model in the working paper of \cite{dogan2018choice}. In this choice model, one considers two sets of rankings: ``pro'' rankings and ``con'' rankings. Then, over the union of the pro and con rankings, one posits a signed probability distribution, where the pro rankings receive positive probabilities, and the con rankings receive negative probabilities. The choice probability of a product given an assortment is the sum of the (positive) probabilities for the pro rankings for which that product is highest ranked, plus the sum of the (negative) probabilities for the con rankings for which that product is lowest ranked. The model aims to represent the idea of a decision maker who makes decisions by listing the pros and cons of an option, adding up the pros and subtracting the cons. The main result of \cite{dogan2018choice} is the result that every choice model can be represented as a pro-con model, which is again similar to our universality result (Theorem~\ref{thm:universal_choice_model} in Section~\ref{subsec:model_universal}). While that paper develops a similar universality result, the proof requires a very careful induction argument, results from number theory and results from network flow optimization; in contrast, the proof of our universality result is simpler, requiring one to only define a specific and intuitively-chosen family of trees and to re-arrange sums.

While these two classes of models also have the universality result, to date, all research on the game tree models and the pro-con model has been theoretical and descriptive in nature, and has focused on categorizing and relating these models to existing models. These papers do not include methodological contributions: specifically, 
there has not been any research that answers the question of how to efficiently estimate these models from data, and that empirically validates these models on real data. In contrast, in our paper, we show that our proposed model can be estimated from data in two tractable ways (Section~\ref{sec:model_estimation_methods}), and its performance can be validated using real transaction data (Section~\ref{sec:experiment_IRI}). Stated more concisely, the main contribution of our paper is a universal choice model that is practical and ready to be used by practitioners for real-world applications.

\subsection{Other areas}
\label{subsec:literature_review_other}

Lastly, our model is also related to the rich literature on tree models in machine learning. Many machine learning methods construct binary tree models that can be used for classification or regression, such as ID3 \citep{quinlan1986induction}, C4.5 \citep{quinlan1993c4} and classification and regression trees (CART; \citealt{breiman1984classification}). In addition, there are also many predictive models that consist of ensembles or forests of trees, such as random forests \citep{breiman2001random} and boosted trees \citep{freund1996experiments}. The main difference between our work and prior work in machine learning is in the use of forests for discrete choice modeling, that is, using a forest to probabilistically model how customers choose from an assortment. To the best of our knowledge, the use of tree ensemble models for discrete choice modeling has not been proposed before.

We do note that some work in operations management has considered the use of tree ensemble models for demand modeling. Two examples of such papers are \cite{ferreira2015analytics} and \cite{misic2017optimization}, which use random forests to model aggregate demand or profit as a function of product prices. These papers, however, do not model substitution effects as a function of the product assortment.

\section{Decision Forest Customer Choice Model}
\label{sec:model}

In this section, we present our decision forest choice model. We begin in Section~\ref{subsec:model_decision_trees} by introducing binary decision trees and defining how customers make purchases according to such decision trees. We then define our choice model in Section~\ref{subsec:choice_model}, and compare it to the ranking-based model in Section~\ref{subsec:model_ranking}. We describe a couple of well-known examples of behavioral anomalies that can be represented by our model in Section~\ref{subsec:model_examples}. Finally, we establish our first key theoretical result, namely that decision forest models can represent \emph{any} customer choice model, in Section~\ref{subsec:model_universal}.

\subsection{Choice Modeling Background}
\label{subsec:model_background}

Consider a universe of $N$ products, denoted by the set $\Ncal \equiv \{1,2,\ldots,N\}$. The full set of purchase options is denoted by $\Ncal^+ \equiv \{ 0,1,2,
\ldots, N \}$, where $0$ corresponds to an outside or ``no-purchase'' option. An assortment $S$ is a subset of $\Ncal$. When offered the assortment $S$, the customer may choose to purchase one of the products in $S$, or choose the no-purchase option 0. 

The behavior of the customer population is represented through a \emph{discrete choice model}. A discrete choice model is defined as a conditional probability distribution $\Pb(\cdot \mid \cdot ): \Ncal^+ \times 2^{\Ncal} \rightarrow [0,1]$ that gives the probability of an option in $\Ncal^+$ being purchased when the customer is offered a particular set of products; that is, $\Pb(o \mid S)$ is the probability of the customer choosing the option $o \in S \cup \{0\}$, when offered the assortment $S \subseteq \Ncal$. Note that $\Pb(o \mid S) = 0$ whenever $o \notin S \cup \{0\}$, which models the fact that the customer cannot choose a product $o$ that is not in the assortment $S$. 

Before continuing, we pause to formally define the concepts of random utility maximization (RUM) and the regularity property. Many previously proposed discrete choice models are based on the RUM concept. In a RUM discrete choice model, each option $o \in \Ncal^+$ is associated with a random variable $V_o$, which corresponds to a stochastic utility for the option $o$. When offered the assortment $S \subseteq \Ncal$, the customer's choice is given by the random variable $\arg \max_{o \in S \cup \{0\}} V_o$; in other words, the utilities $V_0, \dots, V_N$ are realized, and the customer chooses the option from $S \cup \{ 0\}$ that offers the highest utility. Under such a choice model, the choice probabilities are given by 
\begin{equation}
\Pb( o \mid S) = \Pb( V_o > V_j \ \text{for all}\ j \in S \cup \{0\}, j \neq o ).
\end{equation}
By specifying the joint distribution of the random vector $(V_0, V_1, \dots, V_N)$, one can obtain many different types of choice models. For example, if each $V_i = U_i + \epsilon_i$ where each $U_i$ is a deterministic constant and each $\epsilon_i$ follows an independent standard Gumbel distribution, then the choice model $\Pb( \cdot \mid \cdot)$ corresponds to the standard multinomial logit model \citep{train2009discrete}.

A property that is satisfied by RUM choice models is the regularity or weak rationality property, which corresponds to the following family of inequalities:
\begin{equation}
\Pb(i \mid S \cup \{j\}) \leq \Pb(i \mid S), \quad \forall\ S \subseteq \Ncal, \ i \in S \cup \{0\},\ j \in \Ncal \setminus S. \label{eq:regularity}
\end{equation}
In words, whenever we add a new product $j$ to an assortment $S$, the choice probability of each existing product in $S$ cannot increase. Note that every RUM choice model satisfies the regularity property; however, there exist discrete choice models that satisfy the regularity property and that are outside of the RUM class \citep{block1959random}.

\subsection{Decision Trees}
\label{subsec:model_decision_trees}

The choice model that we will define is based on representing the customer population through a collection of customer \emph{types}. A customer type is associated with a purchase decision tree $t$, which is structured as a directed binary tree graph. We use $\leaves(t)$ and $\splits(t)$ to denote the sets of leaf nodes and non-leaf nodes (also called \emph{split} nodes) of decision tree $t$, respectively. For each split node $s$ in $\splits(t)$, we define $\mathbf{LL}(s)$ and $\mathbf{RL}(s)$ as the sets of leaves that belong to the left and right subtree rooted at split node $s$, respectively. Similarly, for each leaf node $\ell$, we define $\mathbf{LS}(\ell)$ and $\mathbf{RS}(\ell)$ as the sets of all split nodes for which $\ell$ is to the left or to the right, respectively.  We use $r(t)$ to denote the root node of tree $t$.
Each node in the tree, whether it is a split or a leaf, is associated with a purchase option; let $x_v$ denote the purchase option associated with node $v$.

Given an assortment $S \subseteq \Ncal$ and a customer following purchase decision tree $t$, the customer will make their purchase decision as follows: starting at the root node $r(t)$, the customer will check whether the purchase option of that node is contained in the assortment $S$ or not. If this option is a member of $S$, the customer proceeds to the left child node; otherwise, if it is not in the assortment $S$, the customer proceeds to the right child node. The process then repeats until a leaf node is reached. The purchase option $o$ that corresponds to the leaf node is then the customer's purchase decision.

Figure \ref{fig:tree_example_1} visualizes an example of a purchase decision tree. Consider a customer following the tree in Figure \ref{fig:tree_example_1}, and consider three assortments: $S_A = \{ 1,2,4 \}$, $S_B = \{  2,4 \}$, and $S_C = \{  1,3 \}$. When offered $S_A$, she will choose product $2$; when offered $S_B$, she will choose product $4$; and finally, when offered $S_C$, she will choose the no-purchase option $0$.

To ensure that a purchase decision tree is well-defined, we impose three additional requirements on it: 
\begin{enumerate}
	\item[\bfseries Requirement 1:] For each split $s \in \splits(t)$, $x_s \in \Ncal$.
	\item[\bfseries Requirement 2:] For each leaf $\ell \in \leaves(t)$, $x_\ell \in  \left( \{ 0 \} \cup  \bigcup_{s \in \mathbf{LS}(\ell) } \{x_s\} \right)$.
	\item[\bfseries Requirement 3:] For each leaf $\ell \in \leaves(t)$ and any two distinct splits $s$ and $s'$ from set $ \mathbf{LS}(\ell) \cup \mathbf{RS}(\ell)$, $x_s \neq x_{s'}$.
\end{enumerate}

Requirement 1 is needed because the no-purchase option can never belong to the assortment; thus, setting $x_s = 0$ at a particular split will force the decision process to always proceed to the right. Requirement 2 is needed to ensure that each possible purchase decision is consistent with the path followed in the tree and that the customer is only able to select products that have been observed to exist in the assortment. An example of a tree that does not satisfy the second requirement is given in Figure \ref{fig:tree_counter_example}. Observe that if the assortment $\{1,2\}$ is offered to a customer following this tree, the customer will choose to purchase product 3, which is not part of the assortment. As another example, if the assortment $\{2,3\}$ is offered, the customer would choose product $1$, which again does not exist in the assortment. Finally, Requirement 3 enforces that each product appears at most once in the split nodes on the path from the root $r(t)$ to any leaf $\ell$. This requirement ensures that for each leaf in the tree, there exists some assortment that will be mapped to it. An example of a tree that does not satisfy the requirement is given in Figure \ref{fig:tree_counter_example_2}, where product 1 appears twice on the path from the root to the third leaf node from the left. In order to reach this leaf, product 1 must simultaneously be included and not included in the assortment, which is impossible. As a result, this leaf node can never be reached given any assortment.

\begin{minipage}{\linewidth}
	\centering
	\begin{minipage}[b]{0.4\linewidth}
		\begin{figure}[H]
			\centering
			\includegraphics[width=7cm]{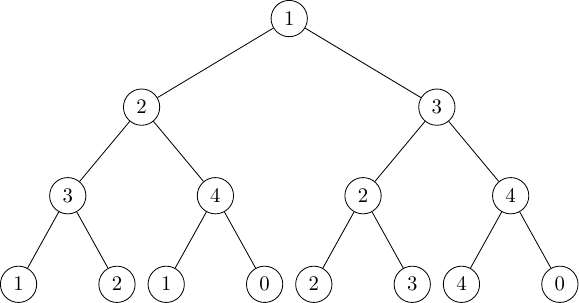}
			\caption{An example of a decision tree.}
			\label{fig:tree_example_1}
		\end{figure}
		\vskip.2\baselineskip
	\end{minipage}
	\hspace{0.05\linewidth}
	\begin{minipage}[b]{0.25\linewidth}
		\begin{figure}[H]
			\centering
			\includegraphics[width=2.5cm]{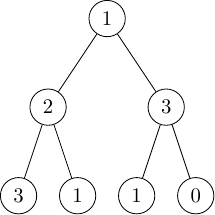}
			\caption{An example of a decision tree that does not satisfy Requirement 2. }
			\label{fig:tree_counter_example}
		\end{figure}
		\vskip.2\baselineskip
	\end{minipage}
	\hspace{0.01\linewidth}
	\begin{minipage}[b]{0.25\linewidth}
		\begin{figure}[H]
			\centering
			\includegraphics[width=2.5cm]{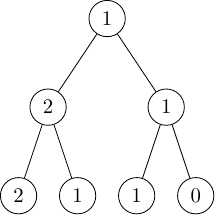}
			\caption{An example of a decision tree that does not satisfy Requirement 3.}
			\label{fig:tree_counter_example_2}
		\end{figure}
		\vskip.2\baselineskip
	\end{minipage}
\end{minipage}

Before describing our choice model, we introduce two useful definitions. We define the \emph{depth} of tree $t$ as $\text{Depth}(t) = \max \{  \textrm{dist}(r(t), \ell) + 1 \mid  \ell \in \leaves(t) \}$, where the distance $\textrm{dist}(r(t),l)$ is the number of edges connecting leaf $\ell$ and root $r(t)$. Note that our definition of depth starts at 1, i.e., a tree consisting of a single leaf would have $\text{Depth}(t) = 1$. We also say that a tree is {\it balanced} if and only if all leaves in the tree have same distance to the root. For example, the tree in Figure \ref{fig:tree_example_1} is a balanced tree of depth $4$. Lastly, notice that Requirement 3 implies that all purchase decision trees have depth at most $N+1$. Therefore, there are only finitely many purchase decision trees that satisfy Requirement 1-3.

\subsection{Decision Forest Model}
\label{subsec:choice_model}

We now present our choice model based on purchase decision trees. Consider a collection $F$ of purchase decision trees; we will refer to $F$ as a {\it decision forest}. Let ${\lambdab}: F \rightarrow [0,1]$ be a probability distribution over all decision trees in forest $F$. Each tree $t$ in the decision forest $F$ can be thought of as a customer type. For each type $t$, the probability $\lambda_t$ can be thought of as the percentage of customers in the population that behave according to the purchase decision tree $t$; alternatively, one can think of $\lambda_t$ as the probability that a random customer will choose according to tree $t$. Define $\hat{A}(S,t) \in \Ncal^+$ as the purchase option that a customer associated with decision tree $t$ would choose when an assortment $S$ is given. Therefore, for any assortment $S$, the probability that a random customer selects option $o \in \{ 0,1,2,\ldots,N\}$ is
\begin{align}
\label{eq:choice_prob_forest_weight}
\Pb^{(F,{\lambdab})}(o \mid S) = \sum_{t \in F} \lambda_t \cdot \Ibb\{ o = \hat{A}(S,t)  \},
\end{align}
where $\Ibb\{\cdot\}$ is the indicator function ($\Ibb\{B\} = 1$ if $B$ is true and $0$ otherwise). Note that if a product $p \in \Ncal = \{1,2,\ldots,N\}$ is not in assortment $S$, i.e., $p \notin S$, then $\Pb^{(F,\lambdab)}(p \mid S) = 0$; this is a consequence of Requirement 2 from Section~\ref{subsec:model_decision_trees}, that is, for any leaf, we must have $x_\ell \in  \left( \{ 0 \} \cup  \bigcup_{s \in \mathbf{LS}(\ell) } \{x_s\} \right)$. We refer to the pair $(F, \lambdab)$ as a \emph{decision forest model}.

\subsection{Comparison to ranking-based model}
\label{subsec:model_ranking}

Our decision forest model resembles the ranking-based model of \cite{farias2013nonparametric}. In the model of \cite{farias2013nonparametric}, each customer type corresponds to a \emph{ranking} over all products and the no-purchase option. When offered an assortment, a customer will choose the product in the assortment that is most preferred according to that customer's ranking. A ranking-based model can be represented as a collection $\Sigma$ of rankings and a probability distribution $\lambdab$ over rankings in $\Sigma$. The ranking-based choice model is thus given by
\begin{align*}
\Pb^{(\Sigma,{\lambdab})} (o \mid S)= \sum_{\sigma \in \Sigma} \lambda_\sigma \cdot \Ibb\{ \text{ranking } \sigma \text{ selects option }o \text{ given }S   \}.
\end{align*}
The ranking-based model and the decision forest model are structurally similar, in that they are both probability distributions over a collection of ``primitive'' choice models. However, it turns out that the decision forest model is more general than the ranking-based model, which we formalize in the proposition below.

\begin{proposition}
	\label{prop:rank_included_by_tree}
	Let $\Sigma = \{ \sigma_1,\ldots,\sigma_m \}$ be a collection of rankings and $\lambdab$ be a probability distribution over them. Then there exists a forest $F$ such that, for all $o \in \Ncal^+$ and $S \subseteq \Ncal$,
	\begin{align*}
	\Pb^{(F,{\lambdab})} (o \mid S) = \Pb^{(\Sigma,\lambdab)} (o \mid S).
	\end{align*}
\end{proposition}

Note that the class of RUM choice models is equivalent to the class of ranking-based models \citep{block1959random}. Thus, Proposition~\ref{prop:rank_included_by_tree} also implies that we can represent any RUM choice model by a decision forest model. The proof of Proposition~\ref{prop:rank_included_by_tree} is presented in Section~\ref{subsec:appendix_proof:prop:rank_included_by_tree}, where we explicitly represent each ranking in $\Sigma$ by a purchase decision tree. We illustrate the same idea in the following example.

\begin{exmp}[RUM choice model]
	\label{example:ranking_included_by_trees}
	Consider a ranking-based model $(\Sigma,\lambdab)$ that consists of two rankings $\sigma_1 = \{ 2 \succ 3 \succ 0  \}$ and $\sigma_2 = \{ 3 \succ 2 \succ 1 \succ 0  \}$, where $a \succ b$ denotes that $a$ is preferred to $b$, and distribution $\lambdab = (0.4,0.6)$. This ranking-based choice model can be represented by a decision forest model $(F,\lambdab)$ such that $F$ consists of two trees $t_1$ and $t_2$ (see Figure~\ref{fig:tree_ranking_example}). The ranking $\sigma_1$ and the decision tree $t_1$ give the same decision process: if product $2$ is in the assortment, then the customer buys it;  otherwise, if product $2$ is not in the assortment but $3$ is, then the customer buys product $3$; otherwise, if both $2$ and $3$ are not available, the customer will not buy anything. The equivalence between the ranking $\sigma_2$ and the tree $t_2$ can be argued in the same way. By using the same probability distribution, the decision forest model $(F,\lambdab)$ is equivalent to the ranking-based model $(\Sigma,\lambdab)$.
\end{exmp}

\begin{minipage}{\linewidth}
	\centering
	\begin{minipage}[b]{0.65\linewidth}
		\begin{figure}[H]
			\centering
			\includegraphics[width=8.cm]{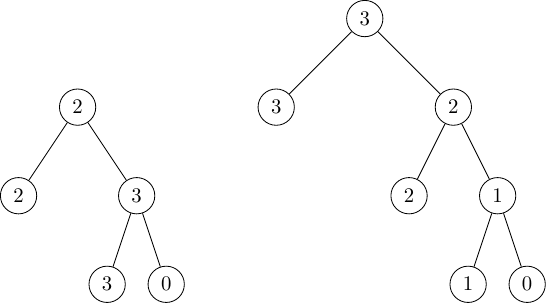}
			\caption{Decision tree $t_1$ (left) and $t_2$ (right) give the same purchase decisions as the two rankings $\sigma_1$ and $\sigma_2$ in Example~\ref{example:ranking_included_by_trees}, respectively.}
			\label{fig:tree_ranking_example}
		\end{figure}
		\vskip.2\baselineskip
	\end{minipage}
	\hspace{0.02\linewidth}
	\begin{minipage}[b]{0.3\linewidth}
		\begin{figure}[H]
			\centering
			\includegraphics[width=3cm]{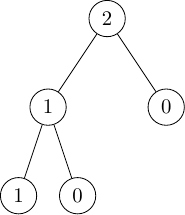}
			\caption{A decision tree that cannot be modeled by a ranking-based model.}
			\label{fig:tree_example_apx_1}
		\end{figure}
		\vskip.2\baselineskip
	\end{minipage}
\end{minipage}

We remark that the reverse statement of Proposition \ref{prop:rank_included_by_tree} is not true. That is, there exist decision forest models that cannot be represented as a ranking-based model. For instance, consider a decision forest model that consists of a single purchase decision tree $t$ as in Figure \ref{fig:tree_example_apx_1}, for which the probability $\lambda_t$ must be 1. This decision tree $t$ gives the following choice probabilities: $\Pb\left(  1 \mid \{ 1,2  \}  \right) = 1$ and $\Pb\left(  1 \mid \{ 1  \}  \right) = 0$. Since the inequality $\Pb\left(  1 \mid \{ 1,2  \}  \right) > \Pb\left(  1 \mid \{ 1  \}  \right)$ violates the regularity property (inequality~\eqref{eq:regularity}), no RUM choice model can satisfy both $\Pb\left(  1 \mid \{ 1,2  \}  \right) = 1$ and $\Pb\left(  1 \mid \{ 1  \}  \right) = 0$.

Before continuing, it is worth interpreting how choices are made by a purchase decision tree and differentiating them from those of a ranking. Example~\ref{example:ranking_included_by_trees} shows how a ranking is effectively a purchase decision tree that is constrained to always grow to the right. 
In addition, each purchase decision of a leaf corresponds to the product on its parent split (with the exception of the right-most leaf, which is always the no-purchase option). 
A customer who chooses according to a ranking behaves in the following way: they check the assortment in accordance with a sequence of products (their ranking); as soon as they reach a product that is contained in the assortment, they choose it; and if they go through their entire sequence without successfully finding a product, they choose the no-purchase option. Such a decision process is always forced to immediately choose a product when the existence of the product in the assortment has been verified. In contrast, for a purchase decision tree, the decision process can be more complicated: if the customer checks for a product and finds that it is indeed contained in the assortment, the customer is not forced to immediately choose the product; instead, the customer can continue checking for other products in the assortment before making a purchase decision. This difference is why purchase decision trees are potentially valuable: a purchase decision tree can model more complicated, assortment-dependent customer behavior than a ranking can. Indeed, in Section~\ref{subsec:model_examples}, we will see some simple examples of non-rational behavior that can be represented in the decision forest framework.

\subsection{Modeling Irrational Behavior by Decision Forest Models}
\label{subsec:model_examples}

Research in marketing, psychology, and economics has documented numerous examples of choice behavior that is inconsistent with the RUM principle. We show how two well-known examples of irrational choices, the decoy effect and the preference cycle, can be modeled by decision forests.

\begin{figure}[t!]
	\centering
	\begin{subfigure}[t]{0.4\textwidth}
		\centering
		\includegraphics[width=3cm]{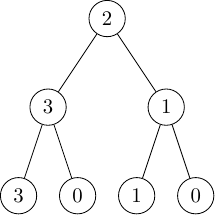}
		\caption{Tree $t_1$ with $\lambda_1 = 0.52$}
		\label{fig:tree_decoy_1}
	\end{subfigure}%
	\begin{subfigure}[t]{0.25\textwidth}
		\centering
		\includegraphics[width=1.5cm]{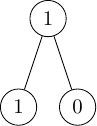}
		\caption{Tree $t_2$ with $\lambda_2 = 0.16$}
		\label{fig:tree_decoy_2}
	\end{subfigure}
	\begin{subfigure}[t]{0.25\textwidth}
		\centering
		\includegraphics[width=1.5cm]{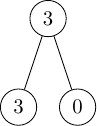}
		\caption{Tree $t_3$ with $\lambda_2 = 0.32$}
		\label{fig:tree_decoy_3}
	\end{subfigure}
	\caption{The forest-distribution pair $F = \{t_1,t_2,t_3\}$ and $\lambda = (\lambda_1,\lambda_2,\lambda_3)$ that can model the decoy effect in \emph{The Economist} subscription example in Table \ref{tb:economist}.}
	\label{fig:decoy_trees}
\end{figure}

\begin{exmp}[Decoy Effect]
	In marketing, the decoy effect is the phenomenon whereby consumers tend to change their preference between two options when a third option exists and it is asymmetrically dominated. The experiment involving the \emph{The Economist} from \cite{ariely2008predictably}, shown in Table~\ref{tb:economist}, 
	is an example of this effect. When the option ``Print-Only'' is strictly dominated by option ``Print-\&-Internet'' (same price but with additional online access), the preference between the other two options changes. 
	
	We model the example in Table~\ref{tb:economist} 
	as follows: denote the subscription options ``Internet-Only'', ``Print-Only'', and ``Print-\&-Internet'' as products $1$, $2$, and $3$, respectively. Define $F = \{t_1,t_2,t_3\}$ as in Figure \ref{fig:decoy_trees} and the corresponding distribution as $\lambdab = (0.52,0.16,0.32)$; it can be verified that this model leads to the choice probabilities in Table~\ref{tb:economist}. 
	Note that customers following $t_2$ will always choose ``Internet-Only''(option $1$), regardless of whether ``Print-Only''(option $2$) is available or not. Similarly, customers following $t_3$ will always choose ``Print-\&-Internet'' (option 3). But for customers following $t_1$, the preference between ``Internet-Only'' (option $1$) and ``Print-\&-Internet'' (option $3$) changes when``Print-Only'' (product $2$) exists. As in Figure \ref{fig:tree_decoy_1}, if product $2$ is included in the assortment, the decision process proceeds to the left subtree and chooses according to the ranking $\{3 \ispreferredto 0\}$, i.e., if product $3$ exists then we buy it; otherwise, we do not buy anything. If product $2$ is not included in the assortment, the decision process proceeds to the right subtree and chooses according to the ranking $\{1 \ispreferredto 0 \}$, i.e., if product $1$ exists then we buy it; otherwise, we do not buy anything. Thus, customers of type $t_1$ account for the decoy effect observed in Table~\ref{tb:economist}.
\end{exmp}

\begin{minipage}{\linewidth}
	\centering
	\begin{minipage}[b]{0.45\linewidth}
		\begin{table}[H]
			\centering
			\begin{tabular}{ccc}
				\toprule
				Gamble & Prob. of Winning & Payoff \\ \midrule
				$A$       & $7/24$             & 5.00   \\
				$B$       & $8/24$             & 4.75   \\
				$C$       & $9/24$             & 4.50   \\
				$D$       & $10/24$            & 4.25   \\
				$E$       & $11/24$            & 4.00  \\ \bottomrule \\
			\end{tabular}
			\caption{Gambles used by \cite{tversky1969intransitivity} to demonstrate preference cycle.}
			\label{tb:pref_cycle}
		\end{table}
		\vskip.2\baselineskip
	\end{minipage}
	\hspace{0.02\linewidth}
	\begin{minipage}[b]{0.45\linewidth}
		\begin{figure}[H]
			\centering
			\includegraphics[width=7cm]{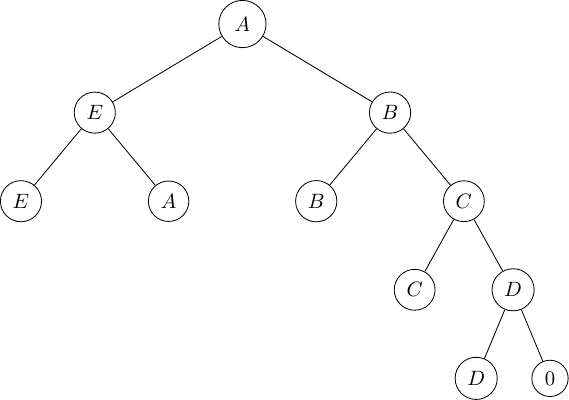}
			\caption{A decision tree representation of a preference cycle.}
			\label{fig:tree_cycle}
		\end{figure}
		\vskip.2\baselineskip
	\end{minipage}
\end{minipage}

\begin{exmp}[Preference Cycle]
	
	The preference cycle is a behavorial anomaly in which the preference relation is not transitive. A classic example is given by \cite{tversky1969intransitivity} and is re-created in Table \ref{tb:pref_cycle} (see also \citealt{rieskamp2006extending}). Participants were offered gambles varying in winning probabilities but with similar payoffs. One group of participants behaved in the following way: when offered two gambles with similar probabilities, they preferred the gamble with the larger payoff. Specifically, they preferred $A$ to $B$, $B$ to $C$, $C$ to $D$, and $D$ to $E$. However, when offered gambles where the winning probabilities were significantly different, they would prefer the gamble with the higher winning probability, e.g., preferring $E$ to $A$. 
	
	We can use a purchase decision tree to model this type of preference cycle, as in Figure \ref{fig:tree_cycle}. It is easy to see that, when assortments $\{A,B\}$, $\{B,C\}$, $\{C,D\}$, $\{D,E\}$, $\{A,E\}$ are given, participants who follow the decision tree would choose $A$, $B$, $C$, $D$, and $E$, respectively. Note that the right subtree of the root node corresponds to the ranking $\{ A \ispreferredto B  \ispreferredto C \ispreferredto D \ispreferredto E\}$ but the left subtree corresponds to the ranking $\{ E \ispreferredto A \}$, therefore leading to the cycle.
	
\end{exmp}

\subsection{Decision Forest Models are Universal}
\label{subsec:model_universal}

As we have shown that two classic examples of irrational choices can be modeled by decision forest model, a natural question to ask is: what is the class of the choice models that can be represented by a decision forest model? Stated differently, for any given general choice model $\Pb(\cdot \mid \cdot)$, does there exist a forest $F$ and a probability distribution $\lambdab$ such that $\Pb(o \mid S) = \Pb^{(F,\lambdab)}(o \mid S)$ for every assortment $S$ and purchase option $o$? The answer, given by Theorem \ref{thm:universal_choice_model}, is in the affirmative.

\begin{theorem}
	\label{thm:universal_choice_model}
	Assume a universe of $N$ products. Let $F_d$ be the collection of all purchase decision trees that satisfy Requirement 1-3 in Section~\ref{subsec:model_decision_trees} and are of depth at most $d$. For any customer choice model $\Pb(\cdot \mid \cdot )$, there exists a distribution $\lambdab$ over $F_{N+1}$ such that 
	\begin{align}
	\Pb(o \mid S) = \Pb^{(F_{N+1},\lambdab)}(o \mid S),
	\end{align}
	for any assortment $S$ and purchase option $o \in \Ncal^+$.
\end{theorem}

The proof of Theorem~\ref{thm:universal_choice_model}, given in Section~\ref{subsec:appendix_proof_thm:universal_choice_model}, 
follows by explicitly constructing a forest of balanced trees of depth $N+1$ that gives identical choice probabilities to $\Pb$, where each tree corresponds to a possible combination of purchase decisions on all $2^N$ assortments and the probability of each tree is given by the product of the choice probabilities of those purchase decisions. Theorem \ref{thm:universal_choice_model} shows that the decision forest model is \emph{universal}: \emph{any} choice model can be represented by a decision forest model. Additionally, Theorem \ref{thm:universal_choice_model} gives another way to prove Proposition \ref{prop:rank_included_by_tree}: since any choice model can be modeled by the decision forest model, ranking-based choice models are thus included as a special case.

\section{Model Complexity Guarantees}
\label{sec:simple_trees}

In this section, we theoretically analyze the problem of estimating a decision forest model $(F,\lambdab)$ from data corresponding to a set of $M$ historical assortments. While Theorem \ref{thm:universal_choice_model} implies that a forest of trees of depth at most $N+1$ is sufficient to fit any data set, this choice may not be attractive when $N$ is large, as the trees will be extremely deep and contain an exponentially large number of leaves. In this section, we ask the question of whether it is possible to fit the data using a ``simple'' decision forest model. In Section~\ref{subsec:simple_trees_motivation}, we define the estimation problem precisely and provide further motivation for considering simple decision forest models. We then show how the number of assortments relates to the depth of the trees, number of leaves of the trees, and number of trees in the forest.

\subsection{Motivation for Simple Decision Forests}
\label{subsec:simple_trees_motivation}

To motivate the value of considering simple forests, let us assume that we have access to sales rate information for a collection of historical assortments $S_1, \dots, S_M$, and let $v_{o,S_m}$ denote the probability with which customers selected option $o \in \Ncal^+$ when assortment $S_m$ was offered, for $m = 1, 2, \dots, M$. We let $\vb_S$ denote the vector of $v_{o,S}$ values for each historical assortment $S$, and we use $\Scal = \{S_1,\dots, S_M\}$ to denote the set of historical assortments. 

We will make the assumption that $v_{o,S_m}$ is known exactly, that is, $v_{o,S_m} = \Pb(o \, | \, S_m)$ for every $(o, S_m)$, where $\Pb( \cdot \, | \, \cdot)$ is the ground truth choice model. This is a reasonable assumption if the number of transaction records for each assortment is large enough that each $v_{o,S_m}$ will be close to the true choice probability $\Pb(o \, | \, S_m)$. Later, in Section~\ref{subsec:estimation_other_objectives}, we will discuss how our estimation methodology can be readily adapted to the setting where the $v_{o,S_m}$ values are derived from limited data. 

We now define the estimation problem. For now, let us assume that we have fixed a collection of candidate trees $F$. For each tree $t \in F$, let us define $A_{t,(o,S)}$ to be 1 if tree $t$ chooses option $o$ when offered assortment $S$, and 0 otherwise. Let us also define $\Ab_{t,S}$ to be the vector of $A_{t,(o,S)}$ values for $o \in \Ncal^+$ with a given assortment $S$. Let $\lambdab = (\lambda_t)_{t \in F}$ be the probability distribution over $F$. With these definitions, to find the probability distribution for the decision forest model, we must find a vector $\lambdab$ that satisfies the following system of constraints:
\begin{subequations}
	\label{prob:feasibility}%
	\begin{alignat}{2}
	& & &	\sum_{t \in F} \Ab_{t,S} \lambda_t = \vb_S, \quad \forall \ S \in \Scal, \label{prob:feasibility_predicted_eq_actual} \\
	& & &	\oneb^T \lambdab = 1, \label{prob:feasibility_unitsum}\\
	& & &	\lambdab \geq \zerob. \label{prob:feasibility_nonnegative}
	\end{alignat}%
\end{subequations}
In the above constraint system, constraints~\eqref{prob:feasibility_unitsum} and \eqref{prob:feasibility_nonnegative} model the requirement that $\lambdab$ be a probability distribution, while constraint~\eqref{prob:feasibility_predicted_eq_actual} requires that for each assortment $S$ in the data, the vector of predicted choice probabilities, $\sum_{t \in F} \Ab_{t,S} \lambda_t$, is equal to the vector of actual choice probabilities, $\vb_S$. Thus, if we could select a reasonable set of candidate trees for our decision forest, then we could, at least in theory, solve the feasibility problem~\eqref{prob:feasibility} to obtain the corresponding probability distribution $\lambdab$.

Notwithstanding any computational questions surrounding problem~\eqref{prob:feasibility}, the remaining question is how one should choose the forest $F$ of candidate trees. According to Theorem~\ref{thm:universal_choice_model}, decision forest models that are defined with $F = F_{N+1}$, where $F_{N+1}$ is the set of trees of depth at most $N+1$, are sufficient to represent any choice model, whether it belongs to the RUM class or not. Thus, an immediate choice of $F$ is $F_{N+1}$, and we would simply solve the feasibility problem~\eqref{prob:feasibility} with $F_{N+1}$ to obtain the corresponding probability distribution $\lambdab$. However, upon closer examination, this particular choice of $F$ is problematic. The flexibility of the $F_{N+1}$ decision forest model that we established in Theorem~\ref{thm:universal_choice_model} implies that, without any additional structure, it is impossible to learn this model from data. Specifically, a consequence of Theorem~\ref{thm:universal_choice_model} is that there always exists a distribution and a set of trees of depth at most $N+1$ such that (i) the model perfectly fits the training data $\{ (S,\vb_{S}) \}_{S \in \Scal}$, and (ii) the model also perfectly fits \emph{any other possible choice probabilities on the assortments outside of the training data}. For example, there exists a forest model that is consistent with the training data $\{ (S,\vb_{S}) \}_{S \in \Scal}$, but always chooses the no-purchase option for every other assortment, i.e., $\Pb(0 \mid S) = 1$ for any $S \notin \Scal$.

This challenge with estimating the decision forest model motivates the need to impose some form of structure on the set $F$ of candidate trees that may be used in the decision forest model. While there are many ways to quantify the size or complexity of a tree, we will primarily focus on two measures: (i) depth and (ii) number of leaves. Both of these measures are commonly applied in tree-based models found in machine learning. For example, the method of limiting the depth of decision trees has been widely used in machine learning algorithms, such as in CART \citep{breiman1984classification}, to avoid overfitting. Similarly, limiting the number of leaves can also prevent overfitting and has been adapted in tree boosting methods \citep{chen2016xgboost}. 
Both depth and number of leaves are closely linked to model complexity: intuitively, as the purchase decision trees in the forest become deeper or have more leaves (which is equivalent to having more splits), each tree is able to exhibit a wider range of behavior as the assortment varies.

There are three advantages to estimating decision forest models consisting of simple trees:
\begin{enumerate}
	\item \textbf{Generalization}. Given two decision forest models that perfectly fit a set of training assortments, it is reasonable to expect that the decision forest that is simpler will be more likely to yield good predictions on new assortments outside of the training set. 
	
	\item \textbf{Tractability}. It is also reasonable to expect that the estimation problem will become more tractable, as the set of possible trees will be much smaller than the set of all possible trees of depth $N+1$ as required by Theorem \ref{thm:universal_choice_model}.
	\item \textbf{Behavioral plausibility}. Lastly, forests of simple trees are more behaviorally plausible than trees of depth $N+1$. As discussed in \cite{hauser2014consideration}, customers often make purchase decisions by first forming a consideration set (a small set of products out of the whole assortment) and then choosing from among the considered products. Restricting the depth or limiting the number of leaves of each tree implies that customers only check for a small collection of products before making their purchase decision, and is congruent with empirical research on how customers choose.
\end{enumerate}

Before presenting the results, we require some additional definitions. We define the \emph{size} of a forest $F$ as $|F|$, the number of trees in the forest; the \emph{depth} of a forest $F$ as $\max_{t \in F} \Depth(t)$, the maximal depth of any tree in the forest $F$; and the \emph{leaf complexity} of a forest $F$ as $\max_{t \in F}  |  \leaves(t)| $, the maximal number of leaves of any tree in the forest $F$.

\subsection{Forests of Simple Tree are Sufficient to Fit Data}
\label{subsec:simple_trees_depth}

Previously, we motivated the estimation of forests comprised of simple trees, i.e., trees whose depth is bounded by some value $d$ or trees whose number of leaves is bounded by some value $\NumLeaves$. However, selecting the right depth $d$ and $\NumLeaves$ is not straightforward. While Theorem \ref{thm:universal_choice_model} guarantees the existence of a forest of depth $N+1$ that is consistent with the training data $\{ (S, \vb_S) \}_{S \in \Scal}$, it is not clear whether there exists a forest of depth $d \ll N+1$ that is consistent with the training data. Additionally, the trees guaranteed by Theorem \ref{thm:universal_choice_model} may have up to $2^N$ leaves.

In this section, we explore the relation between depth, leaf complexity, and size of a decision forest model to the number of historical assortments $M$. We propose two theoretical results that provide guidance on how these complexity parameters may be selected. 

	\begin{theorem}
		\label{thm:master_theorem_complexity}
		For any training data $\Scal$ with $M$ distinct historical assortments, $\Scal = \{S_1, \dots, S_M\}$, there exists a probability distribution $\lambdab$ and a forest $F$ of depth at most $\min\{ N+1, M+1\}$, of leaf complexity at most $2M$, and of size at most $M(N+1) + 1$ such that
		\begin{align*}
		\Pb^{(F,\lambdab)}(o \mid S) = v_{o,S}
		\end{align*}
		for all $S \in \Scal$ and $o \in \Ncal^+$. 
	\end{theorem}

The proof of Theorem~\ref{thm:master_theorem_complexity} (see Section~\ref{subsec:appendix_proof_thm:thm:master_theorem_complexity} of the ecompanion) involves mathematical induction and polyhedral theory. In terms of depth, while Theorem \ref{thm:universal_choice_model} guarantees that we can fit the data with a forest of depth $N+1$, Theorem \ref{thm:master_theorem_complexity} ensures that we can fit the data with a forest of depth at most $\min \{N+1,M+1\}$. This result is particularly attractive when $M < N$. For example, if a seller has only offered $5$ historical assortments over $20$ products, then instead of building a decision forest of depth $21$ as in Theorem \ref{thm:universal_choice_model}, the seller can fit the customer behavior in the data by a forest of depth $6$. In terms of leaf complexity, while Theorem \ref{thm:universal_choice_model} implicitly bounds the leaf complexity by $2^N$, Theorem~\ref{thm:master_theorem_complexity} guarantees that the complexity that scales only linearly in $M$. This result is also attractive because in practice $M$ is unlikely to scale exponentially with respect to $N$ and thus $M \ll 2^N $. Finally, in term of size, Theorem~\ref{thm:master_theorem_complexity} guarantees that number of trees in the decision forest scales as $O(NM)$. We note that \cite{farias2013nonparametric} established a similar size result for ranking-based models, showing that there exists a worst-case distribution over the set of all rankings that is consistent with the data and that has at most $K+1$ non-zero components, where $K$ is the number of item-assortment pairs (see the proof of Theorem 1 of that paper); our result here about forest size can be viewed as a generalization of that result to the decision forest model.

In the case that the number of products $N$ and the number of assortments $M$ are both large, then the forests furnished by Theorem~\ref{thm:master_theorem_complexity} will be very deep. A natural question is whether it is possible to do better than $\min\{N+1, M+1\}$ in this setting. To address model complexity when both $N$ and $M$ are sufficiently large, we propose our second theoretical result, which is formalized below as Theorem~\ref{thm:asymptotic_size_of_forest}. This theorem assumes a simple generative model of how historical assortments are chosen and establishes that, with high probability over the historical assortments, one can fit the data with a decision forest whose depth scales logarithmically in $M$.

\begin{theorem}
	\label{thm:asymptotic_size_of_forest}
	Assume the $M$ assortments $\Scal =  \{ S_1,S_2,S_3, \dots,S_M \} $ of $N$ products are drawn uniformly at random and independently from the set of all $2^N$ possible assortments. With probability at least  $1 - O \left(  M^2 \cdot 2^{- C N / \log_2 M}  \right)$, there exists a distribution $\lambdab$ and a forest $F$ of depth $ O \left( \log_{2} M \right)$ such that $\Pb^{(F,\lambdab)}( o \mid S) = v_{o,S} $ for all $S \in \Scal$ and $o \in \Ncal^+$, where $C > 0$ is a positive constant.
\end{theorem}

Theorem~\ref{thm:asymptotic_size_of_forest} provides an asymptotic lower bound on the probability of the event that there exists a forest of depth logarithmic in $M$ that can perfectly fit the training data, where the randomness is over the draw of $M$ assortments from the set of all $2^N$ assortments. Note that the inequality $N \geq \log_2 M$ always holds, since one will have at most $2^N$ assortments for $N$ products. On the other hand, in real-world data, $M$ is unlikely to scale exponentially with respect to $N$; for example, a retailer offering $1000$ products is unlikely to have offered $2^{1000} \approx 10^{300}$ subsets of those products in the past. Thus, when $N$ is large and $M$ does not scale exponentially with respect to $N$, the factor $N / \log_2 M$ makes the probability lower bound very close to 1. Stated differently, when $N$ is large and $M$ is not too large, most data sets -- that is, most collections of assortments $\Scal$ of $M$ assortments of the $N$ products -- will admit a forest representation that has depth $O(\log_2 M)$. We note that the result is completely independent of the choice probabilities: the result holds no matter what $(\vb_S)_{S \in \Scal}$ is. 

To prove Theorem~\ref{thm:asymptotic_size_of_forest}, we prove an intermediate result, Theorem~\ref{theorem:shallow_general_result} (see Section~\ref{subsec:appendix_proof_thm:asymptotic_size_of_forest}), which provides an explicit upper bound on the probability of not being able to find a forest of a specific choice of depth that is $O(\log_2 M)$ that fits the  data. 
To give a sense of the scale of the probability bound, for a retailer with $N = 10000$ products and $M = 2000$ historical assortments, the bound implies that the probability that the data set cannot be fit by a decision forest of depth at most $33$ is no greater than $6.2 \times 10^{-12}$. In contrast, Theorem \ref{thm:universal_choice_model} and Theorem~\ref{thm:master_theorem_complexity} yield decision forests of depths 10001 and 2001 respectively.

\section{Estimation Methods}
\label{sec:model_estimation_methods}

In this section, we describe two methods to estimate the decision forest model from data, based on column generation (Section~\ref{subsec:estimation_by_column_generation}) and randomized tree sampling (Section~\ref{subsec:estimation_by_sampling_approach}). In Section~\ref{subsec:estimation_overfitting}, we discuss two practical strategies for addressing overfitting. Lastly, in Section~\ref{subsec:estimation_other_objectives}, we discuss how our methods can be extended to other forms of data and other types of objectives.

\subsection{Method \#1: Column Generation}
\label{subsec:estimation_by_column_generation}

Suppose for now that we select a large collection $F$ of candidate trees. As discussed earlier, we wish to find a probability distribution $\lambdab$ over $F$ that satisfies the constraint system~\eqref{prob:feasibility} for $F$. 
If we specify the set of candidate trees $F$ according to the depth or leaf complexity given in Theorem~\ref{thm:master_theorem_complexity} then we are guaranteed the existence of a probability distribution $\lambdab$ that satisfies the constraint system~\eqref{prob:feasibility}. However, the collection of trees $F$ may still be large enough that directly solving the feasibility problem~\eqref{prob:feasibility} with $F$ is computationally unwieldy. More importantly, if we specify $F$ to consist of trees that are simpler (have a lower depth or fewer leaves) than those prescribed in Theorem~\ref{thm:master_theorem_complexity}, then it may not be possible to find a $\lambdab$ that exactly satisfies \eqref{prob:feasibility}. 

Thus, we will instead focus on finding a $\lambdab$ for which $\hat{\vb}_S = \sum_{t \in F} \Ab_{t,S} \lambda_t$, the vector of predicted choice probabilities for the assortment $S$, is close to $\vb_S$, the vector of actual choice probabilities for $S$, for all $S \in \Scal$. One approach to finding such a $\lambdab$ is to formulate an optimization problem where the objective is to minimize the average $L_1$ norm of the prediction errors in the choice probabilities over all historical assortments:
\begin{subequations}
	\begin{alignat}{2}
	& \underset{\lambdab, \hat{\vb}}{\text{minimize}} & \quad & \frac{1}{| \Scal |} \cdot \sum_{S \in \Scal} \left\|  \hat{\vb}_S  - \vb_S \right\|_1 \\
	& \text{subject to} & & \hat{\vb}_S = \sum_{t \in F} \Ab_{t,S} \lambda_t, \quad \forall \ S \in \Scal,\\
	&&& \oneb^T \lambdab = 1,\\
	& & & \lambdab \geq \zerob.
	\end{alignat}
	\label{prob:estimation_abs}%
\end{subequations}
By introducing additional variables $\epsilonb^+_S$ and $\epsilonb^-_S$ for each assortment $S \in \Scal$, we can reformulate problem~\eqref{prob:estimation_abs} as a linear optimization problem. For a given data set $\Scal$ and forest $F$, we refer to this problem as \EstLO, which we define below: 
\begin{subequations}
	\label{prob:master_primal}
	\begin{alignat}{2}
	\EstLO(\Scal, F)\ = \ & \underset{\lambdab, \epsilonb^+, \epsilonb^-}{\text{minimize}} \quad & &  \frac{1}{| \Scal |} \cdot  \left( \sum_{S \in \Scal} \oneb^T \epsilonb^+_S  + \sum_{S \in \Scal} \oneb^T \epsilonb^-_S  \right) \label{prob:master_primal_obj}\\
	& \text{subject to} & & \sum_{t\in F}\Ab_{t,S} \lambda_t + \epsilonb^-_S - \epsilonb^+_S = \vb_S, \quad \forall \ S \in \Scal,  \label{prob:master_primal_Alambdaeqb}\\
	& & & \oneb^T \lambdab = 1, \label{prob:master_primal_unitsum}\\
	& & & \epsilonb^+_S, \epsilonb^-_S \geq \zerob, \quad \forall \ S \in \Scal, \label{prob:master_primal_epsnonnegative} \\
	& & & \lambdab \geq \zerob.
	\end{alignat}%
\end{subequations}

Before presenting our algorithm for solving this problem, we pause to comment on problem~\eqref{prob:master_primal}. Problem~\eqref{prob:master_primal} is similar to the estimation problem that arises for ranking-based models. In particular, \cite{van2014market} study a maximum likelihood estimation problem, while \cite{misic2016data} studies a similar $L_1$ estimation problem, both of which are formulated in a similar way to problem~\eqref{prob:master_primal}. Both \cite{van2014market} and \cite{misic2016data} study solution methods for this general type of problem that are based on column generation, where one alternates between solving a master problem like \eqref{prob:master_primal} for a fixed set of rankings, and solving a subproblem to obtain the new ranking that should be added to the collection of rankings. In a different direction, the conditional gradient approach of \cite{jagabathula2018limit} also involves iteratively adding rankings to a ranking-based model, which also involves solving a similar subproblem.

In the same way, one can also apply a column generation strategy to solve the decision forest estimation problem~\eqref{prob:master_primal}, which we now describe at a high level. For a fixed forest $\hat{F}$, we solve the problem $\EstLO(\Scal, \hat{F})$ to obtain the primal solution $\lambdab$ and the dual solution $(\alphab, \nu)$, where $\alphab = (\alpha_{o,S})_{o \in \Ncal^+, S \in \Scal}$ is the dual variable corresponding to constraint~\eqref{prob:master_primal_Alambdaeqb} and $\nu$ is the dual variable corresponding to the unit sum constraint~\eqref{prob:master_primal_unitsum}. We then solve a subproblem to identify the tree in $F$ with the lowest reduced cost:
\begin{equation}
\min_{t \in F} \left[ -\sum_{S \in \Scal} \alphab^T_S \Ab_{t, S} - \nu \right]. \label{prob:CG_subproblem_abs}
\end{equation}
If the lowest reduced cost is nonnegative, we terminate with $\lambdab$ as the optimal solution. (Note that $\lambdab$ is an optimal solution to $\EstLO(\Scal, \hat{F})$; by setting $\lambda_t = 0$ for all $t \in F \setminus \hat{F}$, $\lambdab$ can be extended to be an optimal solution of $\EstLO(\Scal, F)$.) If the reduced cost is negative, then we add the tree to $\hat{F}$, solve the problem again, and repeat the procedure until the reduced cost becomes nonnegative. The steps of this approach are summarized in Algorithm~\ref{alg:CG}. 

\begin{algorithm}
	{\SingleSpacedXI
		\begin{algorithmic}[1]
			\Procedure{ColumnGeneration}{$F$}
			\State Initialize $\hat{F} \gets \emptyset$
			\Repeat
			\State Solve $\EstLO(\Scal, \hat{F})$ to obtain $\lambdab$, $\alphab$, $\nu$
			\State Set $t^* \gets \arg \min_{t \in F} \left[ -\sum_{S \in \Scal} \alphab^T_S \Ab_{t, S} - \nu \right]$
			\State Set $\hat{F} \gets \hat{F} \cup \{ t^* \}$
			\Until{ $ -\sum_{S \in \Scal} \alphab^T_S \Ab_{t^*, S} - \nu \geq 0$ }
			\State \Return $(\hat{F}, \hat{\lambdab})$
			\EndProcedure
		\end{algorithmic}
		\caption{Column generation method for solving problem~\eqref{prob:estimation_abs}. \label{alg:CG}}
	}	
\end{algorithm}

The key difference in the column generation approach for decision forests compared to column generation approaches for ranking-based models is the subproblem~\eqref{prob:CG_subproblem_abs}: rather than optimizing over the set of all rankings of the $N+1$ options, one must optimize over a collection of trees. This subproblem can be formulated exactly as an integer optimization problem, with a structure that is different from the integer optimization problem that arises in ranking-based models (as in \citealt{van2014market}); we provide the details of the formulation in Section~\ref{subsec:appendix_CG_integer_program} of the ecompanion. Although the resulting exact column generation approach is able to solve problem~\eqref{prob:master_primal} to provable optimality, it is unfortunately not scalable; for example, for trees of depth $d = 4$, $N = 8$ products and $M = 50$ training assortments, the approach can require over 6 hours (see Section~\ref{subsec:appendix_CG_runtime} of the ecompanion for detailed runtime results). 

Motivated by the intractability of solving the subproblem~\eqref{prob:CG_subproblem_abs} exactly, we consider an alternate strategy where we solve the subproblem heuristically. The heuristic procedure involves starting from a degenerate tree consisting of a single leaf, and then iteratively replacing each leaf with a split with two child leaf nodes. The leaf that is chosen for splitting, as well as the product that is placed on that leaf node and the purchase decisions for the two new leaves, are chosen in a greedy fashion, so as to result in the largest improvement in the reduced cost. The procedure terminates when the reduced cost can no longer be decreased. In addition, the procedure also grows each tree to a user-specified maximum depth of $d$; stated differently, a leaf cannot be considered for splitting when it reaches a depth of $d$. 

We formally define our top-down induction heuristic as Algorithm~\ref{alg:TDI}. Within Algorithm~\ref{alg:TDI}, we use $t_0$ to denote a degenerate tree that consists of a single leaf node, whose purchase decision is the no-purchase option 0. We use $\leaves(t, d)$ to denote the set of all leaves in the tree $t$ that are at a depth up to (but not including) $d$. We define $Z_{t, \ell, p, o_1, o_2}$ as the reduced cost of the tree that is obtained by replacing leaf $\ell$ of tree $t$ with a split, setting the product $x_{\ell}$ of that new split to the product $p$, and setting the left child leaf node's purchase decision to $o_1$ and the right child leaf node's purchase decision to $o_2$; we also use $\textsc{GrowTree}(t, \ell, p, o_1, o_2)$ to denote the tree that is obtained from growing tree $t$ in this way. Lastly, we use $P(\ell)$ to denote the set of products that have appeared in the ancestral splits of leaf $\ell$ (i.e., the set of products $p$ for which $x_s = p$ for some split $s$ along the path from the root node to leaf $\ell$). When choosing the product $p$ to appear on the split at leaf $\ell$, Algorithm~\ref{alg:TDI} is restricted to using only those products that have not appeared in an ancestral split, i.e., those products in $\Ncal \setminus P(\ell)$; this ensures that the trees generated by Algorithm~\ref{alg:TDI} satisfy Requirement 3 in Section~\ref{subsec:model_decision_trees}.

When we use the top-down induction heuristic (Algorithm~\ref{alg:TDI}) within the column generation method (Algorithm~\ref{alg:CG}), we refer to the overall method as the \emph{heuristic column generation} (HCG) method. 

\begin{algorithm}

	{\SingleSpacedXI	
		\begin{algorithmic}[1]
			\Procedure{TopDownInduction}{$\alphab$, $\nu$, $d$}
			\State Initialize $t \gets t_0$
			\State Initialize $Z_c \gets \left[ - \sum_{S \in \Scal} \alphab_{S}^T \Ab_{t,S} - \nu \right]$
			\While{ $|\leaves(t,d)| > 0$ }
			\State Compute $Z_{t,\ell, p, o_1, o_2}$ for all $\ell \in \leaves(t,d)$, $p \in \Ncal \setminus P(\ell)$, \par
			\hskip \algorithmicindent$o_1 \in \{p, 0\} \cup \{ x_s \mid s \in \mathbf{LS}(\ell) \}$, $o_2 \in \{0 \} \cup \{ x_s \mid s \in \mathbf{LS}(\ell) \}$
			\State Set $Z^* \gets \min_{\ell, p, o_1, o_2} Z_{t,\ell, p, o_1, o_2}$
			\State Set $(\ell^*, p^*, o^*_1, o^*_2) \gets \arg \min_{(\ell, p, o_1, o_2)} Z_{t,\ell, p, o_1, o_2}$
			\If{$Z^* < Z_c$}
			\State Set $Z_c \gets Z^*$
			\State Set $t \gets \textsc{GrowTree}(t, \ell^*, p^*, o^*_1, o^*_2)$
			\Else
			\State \textbf{break}
			\EndIf
			\EndWhile
			\State \Return $t$, $Z_c$
			\EndProcedure
		\end{algorithmic}
		\caption{Top-down induction method for heuristically solving column generation subproblem~\eqref{prob:CG_subproblem_abs}. \label{alg:TDI}}
	}
\end{algorithm}

We comment on three important aspects of our heuristic column generation method. First, our top-down induction procedure resembles greedy heuristics that are used for other tree models in the machine learning literature, such as CART \citep{breiman1984classification}, C4.5 \citep{quinlan1993c4} and ID3 \citep{quinlan1986induction}. In addition, such algorithms are also used in algorithms that build collections of trees. Within this literature, our heuristic column generation method most resembles boosting, wherein one adds trees (or other weak learners) iteratively to reduce the training error; see, for example, \cite{freund1996experiments}, \cite{chen2016xgboost} and \cite{friedman2001greedy}.

Second, since our top-down induction heuristic considers trees of maximum depth $d$, the overall column generation approach -- Algorithm~\ref{alg:CG} combined with Algorithm~\ref{alg:TDI} to solve the subproblem -- effectively solves the problem $\EstLO(\Scal, F_d)$, where $F_d$ is the set of unbalanced trees of depth at most $d$. We note that the overall approach \emph{heuristically} solves $\EstLO(\Scal, F_d)$; it does not guarantee that the resulting solution is an optimal solution of $\EstLO(\Scal, F_d)$. However, we find that the approach performs well in practice. In Section~\ref{subsec:appendix_CG_runtime} we numerically compare the heuristic column generation approach against the exact approach; we find that the heuristic approach obtains optimal or near-optimal training error in a fraction of the time required by the exact approach. 

Third, the main complexity control in Algorithm~\ref{alg:TDI} is the limit imposed on the depth of the tree. As discussed in Section~\ref{subsec:simple_trees_depth}, one could use the number of leaves instead of the depth to control the complexity of the trees. We can thus consider a variant of Algorithm~\ref{alg:TDI} wherein one terminates the induction procedure upon reaching a user-specified limit on the total number of leaves. We formally define this alternate method in Section~\ref{sec:appendix_leaf_HCG} of the ecompanion.

\subsection{Method \#2: Randomized Tree Sampling}
\label{subsec:estimation_by_sampling_approach}

In this section, we present our second estimation method, which we refer to as the \emph{randomized tree sampling} (RTS) approach. In this approach, instead of sequentially adding trees to a growing collection, we directly sample a large number of trees to serve as the forest $\hat{F}$, and then solve an optimization problem to find the corresponding probability distribution $\lambdab$.

The overall procedure requires three inputs. The first input $K$ is the number of trees to be sampled. The second input $F$ is a base collection of trees that the algorithm will sample from, while the third input $\xib$ is a probability distribution over $F$ according to which we will draw our sample of $K$ trees. We formally define the method as Algorithm \ref{alg:RTS}.

\begin{algorithm}
	{\SingleSpacedXI
		\caption{Randomized tree sampling method for solving problem~\eqref{prob:estimation_abs}. \label{alg:RTS}}
		\begin{algorithmic}[1]
			\Procedure{RandomizedTreeSampling}{$K$, $F$, $\xib$}
			\State Draw $K$ trees $t_1,t_2,\ldots,t_K$ from $F$ according to distribution $\xib$
			\State Set $\hat{F} \gets \{ t_1,t_2,\ldots,t_K  \} $
			\State Solve $\EstLO(\Scal, \hat{F})$ to obtain probability distribution $\hat{\lambdab}$
			\State \Return $(\hat{F}, \hat{\lambdab})$
			\EndProcedure
		\end{algorithmic}
	}
\end{algorithm}

We theoretically characterize how the distribution $\xib$ and the sample size $K$ affect the performance of Algorithm~\ref{alg:RTS} as follows. We first define the training error or \emph{empirical risk} of a decision forest model $\left( F,\lambdab \right)$ with respect to the data $\{ (S,\vb_S) \}_{S \in \Scal}$ as 
\begin{equation}
\mathbf{R}( F,\lambdab ) \equiv \frac{1}{ |\Scal| }  \sum_{S \in \Scal} \left\| \sum_{t \in F} \Ab_{t,S} \lambda_t  - \vb_{S} \right\|_1.
\end{equation}
Our main theoretical result (Theorem~\ref{thm:convergence_of_sampling_method}) states that with high probability, the empirical risk of the model returned by Algorithm~\ref{alg:RTS} converges to the lowest risk attainable by any forest model in a set $\Lambda(C, \xib)$, which will be defined in Theorem~\ref{thm:convergence_of_sampling_method}, with rate $1 / \sqrt{K}$.

\begin{theorem}
	\label{thm:convergence_of_sampling_method}
	Let $F$ be any collection of trees, let $\xib$ be a probability distribution over $F$ such that $\xi_t > 0$ for all $t \in F$, and let $C > 1$ be a constant. Define the set 
	\begin{align}
	\label{eq:set_definition_feasible}
	\Lambda(C, \xib) \equiv \left\lbrace  \lambdab \in \Rbb^{|F|} \mid  \lambda_t \leq C \cdot \xi_t, \,\, \forall t \in F; \ \oneb^T \lambdab = 1; \ \lambdab \geq \zerob \right\rbrace
	\end{align}
	as a collection of probability distributions over $F$. Then for any $\delta > 0$, Algorithm \ref{alg:RTS} returns a forest model $( \hat{F},\hat{\lambdab} )$ such that its empirical risk $\Rb( \hat{F}, \hat{\lambdab} )$ satisfies
		\begin{align*}
		\Rb( \hat{F}, \hat{\lambdab} ) \leq  \underset{ \lambdab  \in \Lambda(C, \xib)}{\min}  \Rb( F,\lambdab ) + \frac{C}{\sqrt{K}} \cdot  {\left( \sqrt{N+1} + 3 \sqrt{ \log (4 /\delta)}  \right)}
		\end{align*} 
	with probability at least $1 - \delta$ over the sample of trees $t_1,\ldots,t_K$ that comprise $\hat{F}$.
\end{theorem}

In words, the training error (i.e., the objective value of problem~\eqref{prob:estimation_abs}) of the decision forest model $(\hat{F}, \hat{\lambdab})$ is bounded with high probability by the sum of two terms, where the first term measures the best possible training error over decision forest models $(F, \lambdab)$ where $\lambdab$ is in $\Lambda(C, \xib)$, while the second term depends linearly on $C$. When $C$ is large, the first term will be small because the set $\Lambda(C, \xib)$ will be larger, but the second term will be large. Similarly, when $C$ is small, the second term will be reduced, but the first term will become larger because the set $\Lambda(C,\xib)$ will shrink. 

The set $\Lambda(C, \xib)$ reflects the ``coverage'' ability of the distribution $\xib$. If the choice probabilities $(\vb_S)_{S \in \Scal}$ can be generated by a decision forest model $(F, \lambdab)$ for some $\lambdab$ from $\Lambda(C, \xib)$ corresponding to a small value $C$, then the number of trees that we need to sample in order to obtain a low training error $\Rb( \hat{F}, \hat{\lambdab})$ will be small. As an example, if $\xib$ corresponds to the uniform distribution over $F$ and if the optimal $\lambdab^*$ that fits the choice probabilities $(\vb_S)_{S \in \Scal}$ is ``close'' to being uniform, then we only need to sample a small number of trees to achieve a low training error, because the implied value of $C$ (i.e., the value of $C$ needed for $\lambdab^*$ to be contained in $\Lambda(C, \xib)$) is small. As another example in contrast to the previous one, if the optimal $\lambdab^*$ is one where (for example) one tree $t'$ has a disproportionately higher probability than the other trees, then we will need to sample many trees from $\xib$ because the implied value of $C$ is large; this makes sense intuitively because one has a low likelihood of sampling $t'$ from $\xib$ when $|F|$ is large. We also note that the effect of the structure of $F$ in terms of the depth or the number of leaves of the trees is captured in the term $\min_{\lambdab \in \Lambda(C, \xib)} \Rb(F, \lambdab)$. As $F$ contains a richer collection of trees, this term will in general become smaller.

We note that Theorem~\ref{thm:convergence_of_sampling_method} is inspired by the literature on randomization in machine learning -- specifically, the idea of training weighted combinations of (nonlinear) features by randomly sampling the features \citep{moosmann2007fast,rahimi2008random,rahimi2009weighted}. Indeed, our proof of Theorem~\ref{thm:convergence_of_sampling_method} adapts the technique in \cite{rahimi2009weighted}, which considers the problem of learning arbitrary weighted sums of feature functions, to the problem of learning a probability distribution (in the setup of \citealt{rahimi2009weighted},\ the weights need not add up to one). In choice modeling, random sampling was previously used in \cite{farias2013nonparametric}. In that paper one formulates the problem of finding the worst-case probability distribution over a collection of rankings, which is a linear optimization problem of a similar form to our estimation problem~$\EstLO(\Scal, F)$. To solve this worst-case problem, one formulates the dual and randomly samples a collection of constraints (i.e., rankings). The paper of \cite{farias2013nonparametric} justifies this by appealing to the paper of \cite{calafiore2005uncertain}, which shows that with $O((1/\epsilon)(NM \ln (1/\epsilon) + \ln( 1 / \delta))$ constraints being sampled, at most an $\epsilon$ fraction of the constraints will be violated, with probability at least $1 - \delta$ over the sampling. However, as noted in \cite{farias2013nonparametric}, the theory of \cite{calafiore2005uncertain} does not govern how far the optimal objective of the sampled problem will be from the complete problem, which is the focus of our result here.

\subsection{Addressing Overfitting}
\label{subsec:estimation_overfitting}

Given the richness of the decision forest model, an important concern is overfitting. In this section, we describe two practical strategies for addressing overfitting in the decision forest model. \\

\noindent \textbf{$k$-fold cross-validation}: As in other machine learning methods, one can use $k$-fold cross validation to tune the hyperparameters for the decision forest model. In this approach, we divide the training set into a collection of $k$ smaller subsets or \emph{folds}. For a fixed value of a hyperparameter, we use the $k-1$ folds as training data to estimate the model with that hyperparameter value, and evaluate the model's performance on the remaining hold-out fold; we repeat this $k$ times, with each of the $k$ folds serving as the hold-out fold, and average over the $k$ folds. We then repeat this for each value of interest for the hyperparameter, and choose the best value. This approach can be used to set the depth limit $d$ for the top-down induction method (Algorithm~\ref{alg:TDI}) within HCG. This approach can also be used to select an appropriate collection of trees $F$ and probability distribution $\xib$ for the randomized tree sampling method; a simple implementation of this idea is to specify $F$ as the set of all balanced trees of depth $d$ that satisfy Requirements 1-3 in Section~\ref{subsec:model_decision_trees}, specify $\xib$ as the uniform distribution over $F$, and use $k$-fold cross-validation to determine the optimal depth $d$. In our numerical experiments in Section~\ref{subsec:experiment_IRI_asst}, we use $k$-fold cross-validation to tune the depth $d$ for the HCG and RTS approaches. \\

\noindent \textbf{Warm-starts}: Both the heuristic column generation method and the randomized tree sampling method build the collection of trees $\hat{F}$ from scratch, without any set of trees explicitly provided by the user. However, they can be easily modified to take an initial set of trees $F_0$ as an input: in Algorithm~\ref{alg:CG}, we can modify line 2 so that we initialize $\hat{F} \gets F_0$, while in Algorithm~\ref{alg:RTS}, we can modify line 3 to set $\hat{F} \gets \{t_1,\dots, t_K\} \cup F_0$. With regard to $F_0$, the simplest choice is the independent demand model, which corresponds to the forest shown in Figure~\ref{fig:independent_demand_model}. Another natural choice for $F_0$ is the set of trees that correspond to a ranking-based model learned by another method (such as \citealt{van2014market} or \citealt{misic2016data}). By warm-starting either Algorithm~\ref{alg:CG} or Algorithm~\ref{alg:RTS} in this way, one can bias the estimation so that the resulting decision forest model is close to the best-fitting ranking-based model, and reduce the possibility of overfitting in cases where the customer choice behavior is close to a rational model. 

\begin{figure}
	\begin{center}
		\includegraphics[width=8cm]{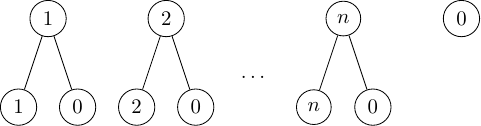}
	\end{center}
	\caption{Collection of trees $F_0$ corresponding to an independent demand model. \label{fig:independent_demand_model} }
\end{figure}

\subsection{Estimating Decision Forests with Log-Likelihood Objective}
\label{subsec:estimation_other_objectives}

So far, we have assumed that the choice probabilities $\vb = (v_{o,S})_{o \in \Ncal^+,S \in \Scal} $ for a set of historical assortments $\Scal$ is known. Our goal has thus been to minimize the error between $\vb$ and $\hat{\vb}$, the choice probabilities predicted by a decision forest model, and we have measured error using the $L_1$ norm. In practice, when the number of transactions is sufficiently large for each assortment $S$, then the frequency of each observed option $o$ given assortment $S$ can serve as an ideal value for $v_{o,S}$.

In other real-world settings, transaction records may be abundant for some assortments but scarce for others. A more common objective function for this finite sample setting is log-likelihood. Let $c(o,S)$ be the number of transactions in which $o$ was chosen given assortment $S$. The maximum likelihood problem can be represented as the following concave optimization problem:
\begin{subequations}
	\label{prob:MLE_estimation}
	\begin{alignat}{2}
	& \underset{\lambdab, \hat{\vb}}{\text{maximize}} \quad
	& &  \sum_{S \in \Scal} \sum_{o \in \Ncal^+} c(o,S) \cdot \log \hat{v}_{o,S}  \label{prob:MLE_estimation_obj} \\
	& \text{subject to} \quad
	& & \hat{\vb}_S = \sum_{t \in F} \Ab_{t,S} \lambda_t, \quad \forall \ S \in \Scal, \label{prob:MLE_estimation_Alambdaeqv}\\
	& & & \oneb^T \lambdab = 1, \label{prob:MLE_estimation_unitsum}\\
	& & & \lambdab \geq \zerob,
	\end{alignat}
\end{subequations}
where the objective is the log-likelihood of the transaction records and $\hat{v}_{o,S}$ is the choice probability of the forest model for option $o$ given assortment $S$. Problem~\eqref{prob:MLE_estimation} only differs from problem~\eqref{prob:estimation_abs} in the objective function. Note that when each column $(\Ab_{t,S})_{S \in \Scal}$ corresponds to a ranking, then problem~\eqref{prob:MLE_estimation} coincides with the maximum likelihood problem that is solved in \cite{van2014market}. The paper of \cite{van2014market} solves this problem using column generation, and shows how one can obtain the dual variables for constraints~\eqref{prob:MLE_estimation_Alambdaeqv} and \eqref{prob:MLE_estimation_unitsum} in closed form. In addition, the paper of \cite{van2017expectation} proposes a specialized expectation maximization (EM) method for solving the ranking-based maximum likelihood problem, without invoking a nonlinear optimization solver. It turns out that for the forest maximum likelihood problem~\eqref{prob:MLE_estimation}, the dual variables can be obtained in the same way as in \cite{van2014market} and the problem itself can be solved with the same EM algorithm from \cite{van2017expectation}. We thus adapt the heuristic column generation and randomized tree sampling methods as follows. For the heuristic column generation, we solve the restricted master problem at each iteration using the EM algorithm of \cite{van2017expectation} and solve the subproblem using our top-down induction method, with the dual variables obtained as in \cite{van2014market}. For the randomized tree sampling algorithm, instead of solving $\EstLO(\Scal, \hat{F})$ with a sampled collection of trees $\hat{F}$, we solve problem~\eqref{prob:MLE_estimation} with $\hat{F}$ using the EM algorithm of \cite{van2017expectation}.

\section{Numerical Experiments with Real Customer Transaction Data}
\label{sec:experiment_IRI}

In this section, we apply our decision forest model to the IRI Academic Dataset \citep{bronnenberg2008database} and evaluate its predictive performance. In addition to these experiments, Section~\ref{sec:appendix_synthetic} of the ecompanion provides results for additional  experiments involving synthetic data.

\subsection{Background}
\label{subsec:experiment_IRI_background}

The IRI dataset is comprised of real-world transaction records of store sales and consumer panels for thirty product categories, and includes sales information for products collected from 47 U.S. markets. The purpose of these experiments is to show how the decision forest model can lead to better predictions of real-world customer choices. We note that the same data set was used in \cite{jagabathula2018limit} to empirically demonstrate the loss of rationality in real customer purchase data. 

To pre-process the data, we follow the same pre-processing steps as in \cite{jagabathula2018limit}. In the dataset, each item is labeled with its respective universal product code (UPC). By aggregating the items with the same vendor code (denoted by digits four through eight of the UPC) as a product, we can identify products from the raw transactions; we note that this is a common pre-processing technique \citep[see][]{bronnenberg2004market,nijs2007retail}. By selecting the top nine purchased products and combining the remaining products as the no-purchase option, we create transaction records for the model setup. Due to the large number of transactions, we follow \cite{jagabathula2018limit} by only focusing on data from the first two weeks of calendar year 2007.

After pre-processing the data, we convert the sales transactions for each product category into assortment-choice pairs $\{(S_t,o_t)\}_{t \in \mathcal{T}}$, where $\mathcal{T}$ is a collection of transactions, as follows. Each transaction $t$ contains the following information: the week of the purchase $(w_t)$, the store ID where the purchase was recorded $(z_t)$, the UPC of the purchased product $(p_t)$. Let $\mathcal{W}$ and $\mathcal{Z}$ be the non-repeated collection of $\{ w_t \}_{t \in \mathcal{T}}$ and $\{ z_t \}_{t \in \mathcal{T}}$, respectively. With week $w \in \mathcal{W}$ and store $z \in \mathcal{Z}$, we define the offer set $S_{w,z} =  \left\lbrace  \bigcup_{t \in \mathcal{T}} \{  p_t \mid w_t = w, z_t = z \} \right\rbrace \cup \{ 0 \}$, as the collection of the products as well as the no-purchase option, purchased at least once at store $z$ in week $w$. As in Section~\ref{subsec:estimation_other_objectives}, we define $c(o,S)$ as the purchase count for option $o$ given assortment $S$, i.e., $c(o,S) = { \sum_{t \in \mathcal{T} } \Ibb \{ S_t = S, o_t = o \} }$.

To quantify the out-of-sample performance of each predictive model on testing transaction set $\Tcal_{\text{test}}$, we use Kullback-Leibler (KL) divergence per transaction, which is defined as
\begin{align*}
\text{KL}(\Tcal_{\text{test}}) = - \frac{1}{|\Tcal_{\text{test}}|}\sum_{S \in \Scal(\Tcal_{\text{test}}) } \sum_{o \in \Ncal^+} {c}(o,S, \Tcal_{\text{test}}) \log \left( p_{o,S} /{v_{o,S}(\Tcal_{\text{test}})}  \right),
\end{align*}
where $\Scal(\Tcal_{\text{test}})$ is the set of assortments found in $\Tcal_{\text{test}}$, $c(o,S, \Tcal_{\text{test}})$ is the number of purchases of option~$o$ given assortment $S$ observed in $\Tcal_{\text{test}}$, $p_{o,S}$ is the predicted choice probability for option $o$ given assortment $S$, and $v_{o,S}(\Tcal_{\text{test}})$ is the empirical choice probability for option $o$ given assortment $S$ derived from the transaction set $\Tcal_{\text{test}}$. Specifically, $v_{o,S}(\Tcal_{\text{test}}) = {c}(o,S,\Tcal_{\text{test}}) / \sum_{o' \in \Ncal^+} {c}(o', S, \Tcal_{\text{test}})$. We remark that \cite{jagabathula2018limit} also used KL divergence as a measure of goodness of fit. While their work focused on the \emph{in-sample} information loss from fitting any RUM model, our numerical experiments here emphasize \emph{out-of-sample} predictive ability.

\subsection{Experiment \#1: Assortment Splitting}
\label{subsec:experiment_IRI_asst}

In our first experiment, we test the out-of-sample predictive ability of our models using five-fold cross validation, where the splitting is done with respect to assortments. We divide the set of assortments $\Scal$ into five (approximately) equally-sized subsets $\Scal_1, \dots, \Scal_5$, and for each $i \in \{1,\dots,5\}$, we use the transaction data for assortments $\Scal_1,\dots, \Scal_{i-1}, \Scal_{i+1}, \dots, \Scal_5$ to build each predictive model and the remaining fold $\Scal_i$ is used for testing. We note that this is a more stringent test of the predictive performance of the models than the standard cross validation based on splitting the transactions, as each model is used to make predictions on assortments that are different from the assortments used to train the models.

In addition to the decision forest model, we test four other models: the ordinary (single-class) MNL model, the latent-class MNL (LC-MNL) model, the ranking-based model and the HALO-MNL model \citep{maragheh2018customer}. For both the MNL and the HALO-MNL models, we fit the parameters using maximum likelihood estimation.  

For the LC-MNL model, we implement the EM algorithm of \cite{train2009discrete}. We tune the number of classes $K$ within the set $\{2,3,5,10,15\}$ using $k$-fold cross validation with $k = 4$, using the previously-defined folds $\Scal_1, \dots, \Scal_5$. We emphasize here that this ``inner'' cross-validation, which involves four folds and is used for tuning the number of classes $K$, is distinct from the ``outer'' cross-validation, which involves five folds and is for the purpose of obtaining a reliable estimate of the out-of-sample KL divergence.

For the ranking-based model, we estimate the model using the column generation method of \cite{van2014market}, where the master problem is solved using the EM algorithm in \cite{van2017expectation}. We define the parameter $\CSsize$ for this model as the maximum allowable consideration set size; in other words, any ranking must be such that there are no more than $\CSsize$ products that are more preferred to the no-purchase option. We tune the parameter $\CSsize$ within the set $\{2,3,4,5,6,7,8,9\}$ using $k$-fold cross validation with $k = 4$, using the folds $\Scal_1,\dots, \Scal_5$. We note that ranking-based models with constrained consideration sets have been considered in previous research on the ranking-based model (see \citealt{feldman2018assortment}).

For the decision forest model, we estimate the model in two different ways. The first involves using the heuristic column generation method in Section~\ref{subsec:estimation_by_column_generation} with log-likelihood as the objective function (as in Section~\ref{subsec:estimation_other_objectives}). We solve the master problem using the same EM algorithm from \cite{van2017expectation}. 
We warm start the model by setting the initial set of trees $F_0$ to be the set of trees corresponding to the rankings estimated for the ranking-based model with $\CSsize = 9$ (note that since the number of products $N = 9$, this value corresponds to estimating ranking-based model without a constraint on the consideration set size). In the same way that we tune $K$ for the LC-MNL model, we also tune the value of $d$, the maximum depth parameter of the top-down induction method (Algorithm~\ref{alg:TDI}). We tune $d$ within the set $\{3,4,5,6,7\}$ using $k$-fold cross validation with $k = 4$, again using the folds $\Scal_1,\dots, \Scal_5$ defined earlier.

The second approach for the decision forest model that we consider is the randomized tree sampling method in Section~\ref{subsec:estimation_by_sampling_approach}, again with log-likelihood as the objective function. We find the optimal $\lambdab$ using the EM algorithm from \cite{van2017expectation}, and as with the HCG method, we warm start the model by setting the initial set of trees $F_0$ to be the set of trees corresponding to the rankings estimated for the ranking-based model with $\CSsize = 9$. We set the base collection of trees $F$ to be sampled as the set of all balanced trees of depth $d$ that satisfy Requirements 1-3, and the distribution $\xib$ as the uniform distribution over $F$. We tune $d$ within the set $\{3,4,5,6,7\}$ using $k$-fold cross-validation with $k = 4$. We fix the number of sampled trees $K$ to 2000; for simplicity, we do not tune the value of $K$.

In the electronic companion, we provide results on other estimation approaches for the decision forest model. In Section~\ref{subsec:appendix_IRI_ass_coldVsWarm}, we compare the ranking-based warm-starting approach against a simpler warm-starting approach using the independent demand model (see Figure~\ref{fig:independent_demand_model} in Section~\ref{subsec:estimation_overfitting}). In Section~\ref{subsec:appendix_IRI_assLL}, we compare the leaf-based heuristic column generation approach described in Section~\ref{sec:appendix_leaf_HCG} to the depth-based heuristic column generation approach. Lastly, in Section~\ref{subsec:appendix_IRI_assRTS}, we provide numerical results that further compare the randomized tree sampling method and the depth-based heuristic column generation method.

Table~\ref{table:IRI_ass_kfold_KL} summarizes the out-of-sample performance of each predictive model over the thirty product categories. The first three columns under ``Datasets'' show the product category, and the number of historical assortments and transactions in that category. The remaining columns report the average out-of-sample KL divergence over five folds.

Out of 30 product categories, the MNL model attains the lowest KL divergence in 1 category, the LC-MNL model attains the lowest in 4 categories, the HALO-MNL model in 8 categories, the ranking-based model in 1 category, and the decision forest model (using either HCG or RTS) in 16 categories. Comparing the decision forest using HCG to the three RUM models (the MNL, LC-MNL and ranking-based models), we find that the decision forest model leads to a lower out-of-sample KL divergence in 22 out of 30 categories. Similarly, the decision forest model using HCG also outperforms the HALO-MNL model in 22 out of 30 categories. In addition, the decision forest model (using either HCG or RTS) achieves lower average, median and maximum KL divergences over the thirty product categories than the other benchmark models. These results suggest the potential of the decision forest model to provide accurate predictions of choice probabilities on new, unseen assortments.

\begin{table}
	{\SingleSpacedXI
		\begin{center}
			\begin{tabular}{lrrrccccc} \toprule
				Product Category & $|\Scal|$ & $|\Tcal|$ & MNL & LC-MNL & HALO-MNL & RM & DF & DF \\ 
				& & & & & & & (HCG) & (RTS) \\ \midrule 
				Beer & 55 & 380,932 & 6.43 & 5.68 & \bfseries 0.79 & 5.49 & 0.88 & 1.56 \\ 
				Blades & 57 &  92,404 & 0.48 & \bfseries 0.40 & 0.52 & 1.44 & 0.41 & 1.02 \\ 
				Carbonated Beverages & 31 & 721,506 & 2.85 & 2.54 & \bfseries 0.95 & 2.65 & 1.56 & 1.56 \\ 
				Cigarettes & 68 & 249,668 & 1.91 & 1.67 & \bfseries 0.91 & 1.65 & 0.98 & 0.96 \\ 
				Coffee & 47 & 372,536 & 2.99 & 2.03 & 2.11 & 2.03 & 1.96 & \bfseries 1.64 \\ 
				Cold Cereal & 15 & 577,236 & 1.73 & 1.79 & \bfseries 0.58 & 2.10 & 0.90 & 0.69 \\ 
				Deodorant & 45 & 271,286 & 0.61 & 0.73 & 0.82 & 0.83 & \bfseries 0.42 & 0.68 \\ 
				Diapers & 18 & 143,055 & 3.34 & 1.54 & 58.56 & 7.13 & \bfseries 1.07 & 1.51 \\ 
				Facial Tissue & 43 &  73,806 & 1.39 & 1.09 & 1.47 & 1.21 & \bfseries 0.77 & 1.32 \\ 
				Frozen Dinners & 30 & 979,936 & 1.44 & 0.95 & 3.84 & \bfseries 0.94 & 2.40 & 1.98 \\ 
				Frozen Pizza & 61 & 292,878 & 2.76 & 2.13 & \bfseries 1.04 & 2.10 & 1.10 & 1.13 \\ 
				Hotdogs & 100 & 101,624 & 3.52 & 3.22 & \bfseries 2.81 & 3.17 & 2.97 & 2.92 \\ 
				Household Cleaners & 19 & 282,981 & 0.94 & 0.93 & 1.61 & 0.96 & 0.68 & \bfseries 0.51 \\ 
				Laundry Detergent & 56 & 238,163 & 2.37 & 2.30 & 2.29 & 2.39 & \bfseries 2.13 & 2.33 \\ 
				Margarine/Butter & 18 & 140,969 & 2.21 & 2.06 & 1.68 & 2.04 &  1.19 & \bfseries 0.74 \\ 
				Mayonnaise & 48 &  97,282 & 1.33 & 0.94 & 0.93 & 0.90 & \bfseries 0.84 & 0.89 \\ 
				Milk & 49 & 240,691 & 4.22 & 3.63 & 1.59 & 2.78 & \bfseries 1.29 & 1.45 \\ 
				Mustard/Ketchup & 44 & 134,800 & 1.32 & 1.06 & 0.78 & 1.10 & \bfseries 0.74 & 0.80 \\ 
				Paper Towels & 40 &  82,636 & 1.21 & \bfseries 1.09 & 1.42 & 1.17 & 1.09 & 1.10 \\ 
				Peanut Butter & 51 & 108,770 & 2.05 & 1.52 & 1.86 & 1.66 & \bfseries 1.49 & 1.51 \\ 
				Photo & 80 &  17,047 & 0.84 & \bfseries 0.76 & 4.66 & 3.33 & 1.31 & 1.28 \\ 
				Salty Snacks & 39 & 736,148 & 1.87 &  1.74 & 2.09 & 1.79 & 1.77 & \bfseries 1.70 \\ 
				Shampoo & 66 & 290,429 & 1.17 & 1.37 & \bfseries 0.86 & 1.34 & 0.95 & 1.21 \\ 
				Soup & 24 & 905,541 & 1.19 & 1.17 & 2.86 & 1.04 & \bfseries 0.95 & 1.63 \\ 
				Spaghetti/Italian Sauce & 38 & 276,860 & 3.38 & 3.26 & 4.44 & 2.89 & 3.37 & \bfseries 2.88 \\ 
				Sugar Substitutes & 64 &  53,834 & 0.83 & \bfseries 0.76 & 0.79 & 0.93 & 0.77 & 0.88 \\ 
				Toilet Tissue & 27 & 112,788 & \bfseries 1.42 & 1.49 & 2.09 & 1.79 & 1.47 & 1.86 \\ 
				Toothbrush & 114 & 197,676 & 1.53 & 1.28 & \bfseries 0.60 & 1.23 & 0.99 & 1.19 \\ 
				Toothpaste & 42 & 238,271 & 0.53 & 0.55 & 0.37 & 0.64 & \bfseries 0.35 & 0.39 \\ 
				Yogurt & 43 & 499,203 & 4.71 & 4.38 & 4.07 & 3.16 & 2.80 & \bfseries 1.58 \\  \midrule
				(Mean) & -- &      -- & 2.09 & 1.80 & 3.65 & 2.06 & \bfseries 1.32 & 1.36 \\ 
				(Median) & -- &      -- & 1.63 & 1.50 & 1.53 & 1.72 & \bfseries 1.08 & 1.30 \\ 
				(Maximum) & -- &      -- & 6.43 & 5.68 & 58.56 & 7.13 & 3.37 & \bfseries 2.92 \\ \bottomrule
			\end{tabular}
		\end{center}
		
		\caption{Out-of-sample KL divergence (in units of $10^{-2}$) for each model over the thirty product categories in the IRI data set, under assortment-based splitting (Section~\ref{subsec:experiment_IRI_asst}). The best performing method in each category is indicated in bold. \label{table:IRI_ass_kfold_KL}}
	}
\end{table}

In addition to out-of-sample performance, it is also interesting to compare the models in terms of runtime. Due to space considerations, these results are relegated to Section~\ref{subsec:appendix_IRI_ass_runtime_modelsize} of the ecompanion. In terms of runtime, we find that the decision forest model requires on average about 3 minutes for the HCG approach, and about 5 minutes for the RTS approach, which includes the time to estimate the ranking-based model and the time for the 4-fold cross-validation to determine the depth limit $d$ (in the case of the HCG approach) or the depth $d$ of the base forest (in the case of the RTS approach). This compares favorably to LC-MNL and the ranking-based model, which require about 14 minutes and 50 minutes on average, respectively (note that this time includes the time to perform cross-validation for the number of classes $K$ and the consideration set size $\tau$).

\subsection{Experiment \#2: Temporal Splitting}
\label{subsec:experiment_IRI_temporal}

In addition to the assortment-based splitting schemes, we consider an additional splitting scheme that we term \emph{temporal splitting}. In this experimental approach, we use the first two weeks of transactions in 2007 in the IRI data set as training data, and then use the following four weeks as test data; this approach emulates how one would use the predictive models to make predictions prospectively (i.e., for transactions occurring in the future). We find that the decision forest model continues to deliver excellent performance in this experimental regime. Due to space considerations, the results are described in greater detail in Section~\ref{sec:appendix_IRI_temporal} of the ecompanion.

\subsection{Extracting Substitution and Complementarity Behavior}
\label{subsec:experiment_IRI_interpretation}

In addition to obtaining predictions, choice modeling is useful for obtaining insights on the relationship between products, i.e., how the presence of one product will affect the choice probability of another product. For parametric choice models, such as the LC-MNL model, such insights can be easily obtained by examining the estimated utility parameters. In contrast, for nonparametric models such as the ranking-based model or the decision forest model, it is less straightforward to obtain a simple picture of the relationship between products.

In this section, we propose a simple method for extracting substitution and complementarity effects between products for a given choice model, and use it to analyze the decision forest model for a single product category in the IRI dataset. We note that our procedure is not specific to the decision forest model and can be used for other choice models (such as the ranking-based model, which we also analyze), and thus may be of independent interest.

We first define a function $\Delta(j,k,S)$ of a product $j$, a product $k \neq j$, and an assortment $S$ that does not include product $j$ and $k$, as
\begin{align}
\label{eq:substitute_and_complement}
\Delta(j,k,S) = \frac{ \mathbf{P} ( j \mid S \cup \{j,k \}) - \mathbf{P}( j \mid S \cup \{j \})    }{\mathbf{P}( j \mid S \cup \{ j  \}) },
\end{align}
which measures the relative change in the choice probability of product $j$ when product $k$ is introduced to assortment $S$. For convenience, we define $\Delta(j,k,S) \equiv 0$ if $j=k$. 
We say that product $k$ complements product $j$ under assortment $S$ when $\Delta(j,k,S) > 0$. Similarly, we say product $k$ substitutes product $j$ under $S$ when $\Delta(j,k,S) < 0$.

The substitution and complementarity relation depends on the existence of other products, i.e., on the assortment $S$. To quantify the overall impact of product $k$ toward product $j$, we consider the averaged version of $\Delta(j,k,S)$. That is, we consider $\Delta^{\text{avg}}_{j,k} = (1 / | \Scal^{\backslash j k}  |) \cdot \sum_{S \in  \Scal^{\backslash j k}  } \Delta(j,k,S) $, where $ \Scal^{\backslash j k}$ is the set of all assortments that do not include product $j$ and $k$. Similarly, we can also define $\Delta^{\text{avg}}_{0,k}$ to measure the average impact of the addition of product $k$ on the no-purchase option. We use $\Deltab^{\text{avg}} = [\Delta^{\text{avg}}_{j,k}]_{j \in \Ncal^+, k \in \Ncal}$ to denote the matrix of all such values.

Figure~\ref{fig:demand_pattern_coffee} illustrates two $\Deltab^{\text{avg}}$ matrices, one corresponding to the decision forest model with depth $d = 3$ (left matrix) and the other corresponding to the ranking-based model (right matrix), for the coffee product category. 
For each matrix, starting from the top-left corner that corresponds to $\Delta^{\text{avg}}_{1,1}$, the first 9-by-9 submatrix corresponds to $[\Delta^{\text{avg}}_{j,k}]_{j,k = 1,\ldots,9}$ and the tenth row represents to $[\Delta^{\text{avg}}_{0,k}]_{k = 1,\ldots,9}$, which captures the effects of the presence of each brand on the choice probability of the no-purchase option. Each cell corresponds to the effect of adding the brand on the corresponding column towards the brand on the corresponding row. The color level of each cell in each matrix represents the numeric value of $\Deltab^{\text{avg}}$ in accordance with the color bar on the right hand side of the figure: green corresponds to positive values and shows complementarity behavior, while red corresponds to negative values and shows substitution behavior.

\begin{figure}
	\centering
	\begin{tabular}{cc}
		\includegraphics[width = 0.5\textwidth]{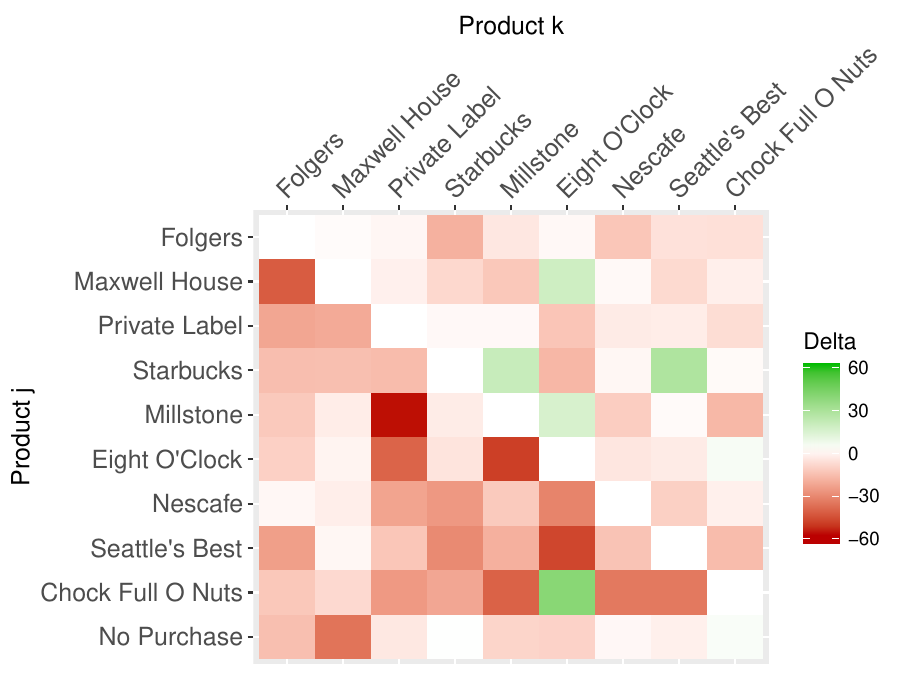} & \includegraphics[width = 0.5\textwidth]{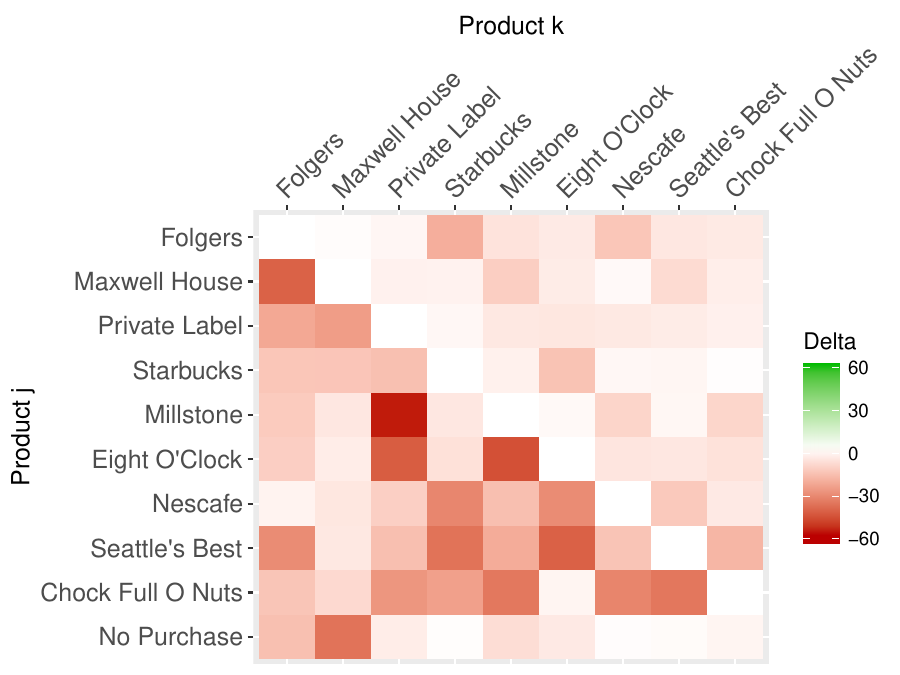}
	\end{tabular}
	
	\caption{Illustration of the substitution/complementarity matrices $\Deltab^{\text{avg}}$ on coffee brands, corresponding to the decision forest model (left) and the ranking-based model (right).}
	\label{fig:demand_pattern_coffee}
\end{figure}

Figure~\ref{fig:demand_pattern_coffee} shows that the decision forest model and the ranking-based model capture similar substitution patterns. For example, both models show that, on average, the choice probabilities of {\it Millstone} and {\it Eight O'Clock} decrease by about 60\% and 40\%, respectively, when {\it Private Label} is added, as shown in elements (5,3) and (6,3) from the top-left corner. However, since the ranking-based model satisfies the regularity property, all elements in $\Deltab^{\text{avg}}$ are forced to be non-positive. Thus, with the ranking-based model we are restricted to understanding only the substitution behavior between products, and we cannot use it to identify any complementarity behavior. 

In contrast, the decision forest model is not constrained by the regularity property, and thus we can use it to identify interesting complementarities between certain brands. For example, Figure~\ref{fig:demand_pattern_coffee} shows that adding {\it Seattle's Best} to the assortment increases the choice probability of {\it Starbucks} by about 25\% on average; the addition of {\it Eight O'Clock} provides a similar boost of about 16\% to the choice probability of {\it Maxwell House}. 

Another way to identify substitution and complementarity effects between products is to directly inspect the decision forest model. Figure~\ref{fig:coffee_decoy_tree} visualizes the top three trees by $\lambda_t$ value of the decision forest model used in the left hand matrix of Figure~\ref{fig:demand_pattern_coffee}. The second tree exhibits the decoy effect that is described in Section~\ref{subsec:model_examples}. This customer type behaves in the following way: when {\it Eight O'Clock} exists in the assortment, the customer will purchase {\it Maxwell House} if it is available; otherwise, if {\it Eight O'Clock} does not exist in the assortment, the customer will purchase {\it Private Label} if it is available. This matches the complementarity effect shown in Figure~\ref{fig:demand_pattern_coffee} (element (2,6) in the left-hand matrix).  The decision forest model is also capable of capturing effects that do not fit into well-studied customer behaviors in the marketing literature. For example, the first tree in Figure~\ref{fig:coffee_decoy_tree} corresponds to customers who purchase {\it Millstone} only if {\it Eight O'Clock} is observed in the assortment; otherwise, they do not make a purchase. Similarly, the third tree represents a decoy-like effect, where a customer checks for the existence of {\it Starbucks}: if it exists, the customer will purchase {\it Starbucks} if {\it Seattle's Best} is available; if not, the customer will purchase {\it Maxwell House} if it is available. This highlights another benefit of our nonparametric approach: since we do not impose any assumptions on how the data is generated, we are able to discover interesting customer behaviors that fall outside of well-studied irrational behaviors.

\begin{figure}[H]
	
	\centering
	\begin{tabular}{ccc}
		\includegraphics[width=0.24\textwidth]{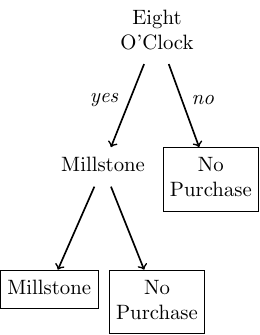}
		& \includegraphics[width=0.37\textwidth]{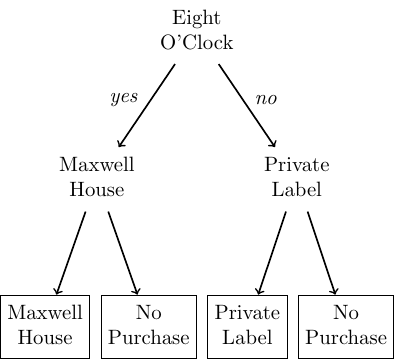}
		& \includegraphics[width=0.39\textwidth]{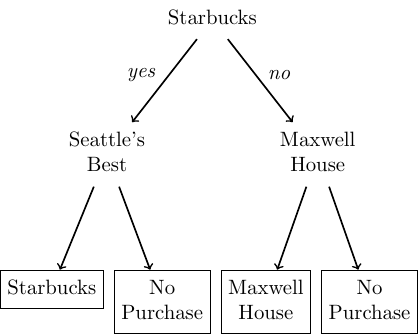}
	\end{tabular}	
	\caption{Top three decision trees on coffee brands with highest probability weights in the decision forest model learned from data. From left to right, the probability weights ($\lambda_t$ values) are $4.1 \%$, $2.5\%$, and $2.3\%$, respectively. \label{fig:coffee_decoy_tree}}
\end{figure}

\section{Conclusions}
\label{sec:conclusions}
In this paper, we proposed the decision forest model, which can model any discrete choice behavior, regardless of whether it belongs to the RUM class or not. Given data in the form of a collection of historical assortments, we proved that simple trees, whose depth scales logarithmically and leaf complexity scales linearly with the number of assortments, are sufficient to fit the data. We further proposed two practical estimation methods for learning the decision forest model from historical assortments. Through experiments with real data, we showed that the decision forest model generally outperforms other rational and non-rational models in out-of-sample prediction in the presence of non-rational customer behavior, and can be used to generate insights on the complementarity/substitution behaviors between products. We hope that this work will encourage the further exploration of data-driven methodologies in the non-rational choice modeling space.

\ACKNOWLEDGMENT{The authors sincerely thank the department editor Chung-Piaw Teo, the associate editor and two anonymous reviewers for their thoughtful comments and feedback that have helped to significantly improve the paper. The authors also thank Huseyin Topaloglu for helpful feedback and being the discussant of this paper at the 2019 INFORMS Revenue Management and Pricing Conference. The authors gratefully acknowledge Information Resources Inc. (IRI) and the authors of \cite{bronnenberg2008database} for the IRI Academic Data Set that was used in Section~\ref{sec:experiment_IRI}. Any findings expressed in this paper are those of the authors and do not necessarily reflect the views of IRI.}

\bibliographystyle{plainnat} %
\bibliography{Reference_Decision_Forest}

\ECSwitch

\ECHead{Electronic companion for ``Decision Forest: A Nonparametric Approach to Modeling Irrational Choice''}

\section{Proofs}
\label{sec:appendix_proofs}

\subsection{Proof of Proposition \ref{prop:rank_included_by_tree}}
\label{subsec:appendix_proof:prop:rank_included_by_tree}
For each ranking $\sigma_j$, $j=1,2,\ldots,m$, we can write down its preference order explicitly as $\sigma_j = \{p^{(j)}_1 \ispreferredto p^{(j)}_2 \ispreferredto \ldots \ispreferredto p^{(j)}_{K_j} \ispreferredto 0\}$, where $a \ispreferredto b$ denotes that $a$ is preferred to $b$; for this ranking, $K_j$ products are preferred over the no-purchase option, and product $p^{(j)}_1$ is the most preferred. Assume each ranking $\sigma_j$ has probability weight $\lambda_j$. Now we construct the forest $F$ as follows: for $j = 1 ,\ldots, m$, we build a decision tree $t_j$ with the structure shown in Figure \ref{fig:tree_ranking}. Additionally, we associate tree $t_j$ with probability $\lambda_j$. Note that ranking $\sigma_j = \{ p^{(j)}_1 \ispreferredto p^{(j)}_2 \ispreferredto \ldots \ispreferredto p^{(j)}_{K_j} \ispreferredto 0\}$ and the decision tree in Figure \ref{fig:tree_ranking} give the same decision process: if product $p^{(j)}_1$ is in the assortment, we buy it;  otherwise, if product $p^{(j)}_1$ is not in the assortment but $p^{(j)}_2$ is, we buy $p^{(j)}_2$; otherwise, if both $p^{(j)}_1$ and $p^{(j)}_1$ are not in assortment but $p^{(j)}_3$ is, we buy $p^{(j)}_3$; and so on. Therefore, for any option $o$ and any assortment $S$, we have
\begin{align*}
\Pb^{(F,{\lambda})}(o \mid S) & = \sum_{j =1}^m \lambda_j \cdot \Ibb\{ o = \hat{A}(S,t_j)  \} \\  & = \sum_{j =1}^m \lambda_j \cdot \Ibb\{ \text{ranking } \sigma_j \text{ selects option }o \text{ given }S   \} \\
& =  \Pb^{(\Sigma,\lambdab)}(o \mid S),
\end{align*}
where the second equality comes from the fact that ranking $\sigma_i$ and tree $t_i$ makes same decision given $S$, and the third equality is from the definition of the ranking-based model. \hfill $\Halmos$

\begin{figure}[H]
	\centering
	\includegraphics[width=6cm]{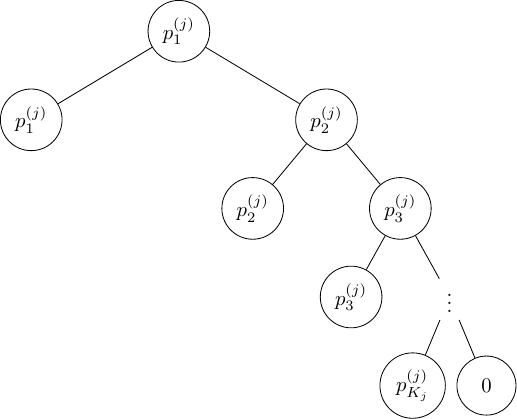}
	\caption{Ranking $\sigma_j = \{ p^{(j)}_1 \ispreferredto p^{(j)}_2 \ispreferredto \ldots \ispreferredto p^{(j)}_{K_j} \ispreferredto 0\}$ can be represented as a purchase decision tree.}
	\label{fig:tree_ranking}
\end{figure}

\subsection{Proof of Theorem \ref{thm:universal_choice_model}}
\label{subsec:appendix_proof_thm:universal_choice_model}

We prove the theorem by constructing a forest $F$ consisting of $(N+1)^ {2^N}$ balanced trees. We first define the common structure of each tree in the forest. Each tree $t$ in the forest has depth $N+1$ and shares the following structure for first the $N$ levels:
\begin{align}
& x_s = 1, \quad \forall \ s \in \splits(t)\ \text{such that}\ \textrm{dist}(r(t),s) = 0, \label{eq:universal_tree_level1} \\
& x_s = 2, \quad \forall \ s \in \splits(t)\ \text{such that}\ \textrm{dist}(r(t),s) = 1, \\
& x_s = 3, \quad \forall \ s \in \splits(t)\ \text{such that}\ \textrm{dist}(r(t),s) = 2, \\
& \vdots \nonumber \\
& x_s = N, \quad \forall \ s \in \splits(t)\ \text{such that}\ \textrm{dist}(r(t),s) = N-1, \label{eq:universal_tree_levelN} 
\end{align}
In words, the root split node $r(t)$ checks for the existence of product $1$ in the assortment; the split nodes in the second level of the tree (those with $\textrm{dist}(r(t),s) = 1$) check for product 2; the split nodes in the third level check for product $3$; and so on, all the way to the $N$th level, at which all splits node check for product $N$. Figure~\ref{fig:thm1_proof} provides an example of this tree structure for $N=3$. In this tree, the left-most leaf node corresponds to assortment $S = \{1,2,3\}$, the second leaf node from the left corresponds to assortment $S = \{1,2\}$, and so on, until the right-most leaf which corresponds to the empty assortment ($S = \emptyset$). 

\begin{figure}[h!]
	\centering
	\includegraphics[width=8cm]{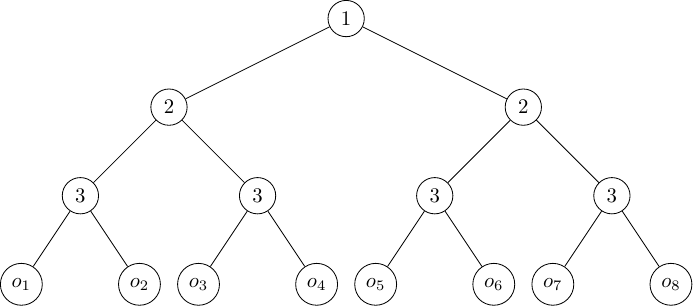}
	\caption{An example of the structure of the trees in the forest needed for the proof of Theorem~\ref{thm:universal_choice_model} for $N=3$.}
	\label{fig:thm1_proof}
\end{figure}

In this tree structure, there are exactly $2^{N}$ leaf nodes, which we will index from left to right as $l_1,\ldots,l_{2^N}$. Note that each leaf node of a tree in $F$ has a one-to-one correspondence with one of the $2^N$ possible assortments of the products. 

To specify the leaves, we require some additional definitions. Let us denote the $2^N$ possible assortments of the $N$ products by $S_1,S_2,\dots,S_{2^N}$, in correspondence with the leaf nodes $l_1,\ldots,l_{2^N}$, respectively. For the leaves $\ell_1, \dots, \ell_{2^N}$, we use $o_1, o_2, \dots, o_{2^N}$ to denote the purchase decisions associated with those leaves, and we use $\ob = (o_1, \dots, o_{2^N})$ to denote the $2^N$-tuple of purchase decisions. Let $\ob^1,\dots, \ob^{ (N+1)^{2^N} }$ be a complete enumeration of the $(N+1)^{2^N}$ possible $2^N$-tuples of the purchase decisions from $\prod_{i=1}^{2^N} \Ncal^+$. 

With these definitions, we define our forest $F$ as consisting of $(N+1)^{2^N}$ trees, where each tree follows the structure in equations~\eqref{eq:universal_tree_level1}--\eqref{eq:universal_tree_levelN}, each tree is indexed from $t = 1$ to $t = (N+1)^{2^N}$ and the purchase decisions of the leaves in tree $t$ are given by the tuple $\ob^t$ as defined above. We define the probability distribution $\lambdab = (\lambda_1,\dots, \lambda_{(N+1)^{2^N}})$ by defining the probability $\lambda_t$ of each tree $t$ as:
\begin{equation}
\lambda_t = \prod_{j=1}^{2^N} \Pb(o^t_j \mid S_j). \label{eq:universal_lambda_def}
\end{equation}
It is straightforward to verify that $\lambdab$ is nonnegative and sums to one. Before continuing, we note that not all of the trees defined in this way will satisfy Requirement 2 from Section~\ref{subsec:model_decision_trees}, which requires that the purchase decision in each leaf must be consistent with the products along the path of splits from the root node to the leaf; in other words, for some trees $o^t_j$ will not be contained in $S_j$. However, for any tree where this is the case, by the definition in equation~\eqref{eq:universal_lambda_def}, the corresponding $\lambda_t$ will be zero because $\Pb(o^t_j \mid S_j)$ is zero whenever $o^t_j \notin S_j$. Thus, those trees can be safely removed from $F$ without changing the overall model.

We now show that the decision forest model $(F, \lambdab)$ outputs the same choice probabilities as the true model $\Pb(\cdot \mid \cdot)$. For any assortment $S_i$ and option $o$, we have
\begin{align*}
& \Pb^{(F,\lambda)}(o \mid S_i) \\
& = \sum_{t=1}^{(N+1)^{2^N}} \lambda_t \cdot \Ibb\{ o = \hat{A}(S_i, t) \} \\
& = \sum_{t\, :\, o^t_i = o} \lambda_t \\
& = \sum_{t\, :\, o^t_i = o} \ \prod_{i'=1}^{2^N} \Pb(o^t_{i'} \mid S_{i'} ) \\
& = \sum_{o_1 =0}^N \cdots \sum_{o_{i-1} =0}^N  \sum_{o_{i+1} =0}^N \cdots \sum_{o_{2^N} =0}^N \Pb(o_1 \mid S_1) \cdots \Pb(o_{i-1} \mid S_{i-1}) \cdot \Pb(o \mid S_i) \cdot \Pb(o_{i+1} \mid S_{i+1}) \cdots \Pb( o_{2^N} \mid S_{2^N}) \\
& = \Pb(o \mid S_i) \left[ \sum_{o_1 = 0}^{ N } \Pb(o_1 \mid S_1) \right] \cdots \left[ \sum_{o_{i-1} = 0}^{ N } \Pb(o_{i-1} \mid S_{i-1}) \right]
\left[\sum_{o_{i+1} = 0}^{ N } \Pb(o_{i+1} \mid S_{i+1}) \right] \cdots \left[ \sum_{o_{2^N} = 0}^{ N }   \Pb(o_{2^N} \mid S_{2^N}) \right] \\
& = \Pb(o \mid S_i),
\end{align*}
where the first equality follows from how choice probabilities under the decision forest model are defined in equation~\eqref{eq:choice_prob_forest_weight}; the second follows from how our forest is constructed and how the $\ob^t$ tuples are defined; the third equality follows from the definition of $\lambdab$; the fourth from the definition of the $\ob^t$ tuples; the fifth by algebra; and the last by recognizing that the choice probabilities for a given assortment must sum to one.
\hfill \Halmos

\subsection{Proof of Theorem~\ref{thm:master_theorem_complexity}}
\label{subsec:appendix_proof_thm:thm:master_theorem_complexity}
Before we prove Theorem~\ref{thm:master_theorem_complexity}, we establish three useful lemmas. Some of the results will be also used in the proof of Theorem~\ref{thm:asymptotic_size_of_forest} in Section~\ref{subsec:appendix_proof_thm:asymptotic_size_of_forest}.

	\begin{lemma}
		\label{lemma:equivalence_no_repeated_products}
		For any decision tree $t$ that satisfies Requirements 1 and 2 in Section \ref{subsec:model_decision_trees}, there exists a purchase decision tree $t'$ such that:
		\begin{enumerate}
			\item[(1)] Tree $t'$ satisfies Requirement 1, 2 and 3;
			\item[(2)] Tree $t$ and $t'$ have same purchase decision under any given assortment;  and 
			\item[(3)] $\Depth(t') \leq \Depth(t)$ and $| \leaves(t')| \leq |\leaves(t)| $.
		\end{enumerate}
	\end{lemma}

\proof{Proof of Lemma~\ref{lemma:equivalence_no_repeated_products}:} Let $t$ be a tree that does not satisfy Requirement 3, and consider the following procedure:

\begin{enumerate}
	\item Select any leaf $\ell \in \leaves(t)$ for which Requirement 3 is violated. Let $path(\ell) \equiv \{ s_1 ,s_2,\ldots,s_{d-1} \}$ be the sequence of splits from root $r(t) \equiv s_1$ to $s_{d-1}$, which is the parent node of $\ell$. Let $i \in \Ncal$ be a product encountered at least two times as the decision process traverses $path(\ell)$. Let $s_{u_{1}}, s_{u_{2}},\ldots,s_{u_{e_i}}$, where $u_1 < u_2 < \dots < u_{e_i}$, be the subsequence of split nodes associated with product $i$ in $path(\ell)$. (Note that $e_i \geq 2$.)
	\item If $s_{u_1} \in \textbf{LS}(\ell)$, then remove all right subtrees branched at $s_{u_2},\ldots,s_{u_{e_i}}$ from $t$. Otherwise, if $s_{u_1} \in \textbf{RS}(\ell)$, then remove all left subtrees branched at $s_{u_2},\ldots,s_{u_{e_i}}$ from $t$.
	\item Delete split nodes $s_{u_2},\ldots,s_{u_{e_i}}$ from tree $t$, and ``glue'' the remaining pieces by setting $s_{u_j - 1}$ as the parent node of the remaining child node of $s_{u_j}$ for $j = 2,\ldots,e_i$.
\end{enumerate}

Consider now applying steps 1 - 3 repeatedly to tree $t$, until all of the leaves in the tree satisfy Requirement 3; let the resulting tree be denoted by $t'$. Note that we are guaranteed to terminate with such a tree, because each time we apply steps 1 - 3, we delete at least one subtree from the tree (and thus at least one leaf), and the tree contains finitely many leaves. In addition to Requirement 3, steps 1 - 3 also preserve Requirements 1 and 2. Thus, tree $t'$ satisfies condition (1) of the lemma.

Note also that for a given tree, the tree we obtain after applying steps 1 - 3 is equivalent to that initial tree, in that any assortment is mapped by the two trees to the same purchase decision. This is true because the leaves within the subtrees that are removed are leaves that are unreachable (i.e., there does not exist an assortment that can reach them). For example, if $\s_{u_1} \in \textbf{LS}(\ell)$, then any leaf in the right subtree branched at $s_{u_2}, \ldots, s_{u_{e_i}}$ is such that product $i$ must be in the assortment and not in the assortment in order to reach the leaf, which is impossible. Thus, tree $t'$ satisfies condition (2) of the lemma.

Lastly, since steps 1 - 3 involve deleting subtrees and splits and re-attaching the disconnected pieces, tree $t'$ is no deeper than tree $t$ and has no more leaves than tree $t$, thus verifying condition (3) of the lemma. \hfill \Halmos

Lemma~\ref{lemma:equivalence_no_repeated_products} plays an important role in the proofs of Theorem~\ref{thm:master_theorem_complexity} and \ref{thm:asymptotic_size_of_forest}. In fact, we prove both theorems by directly constructing decision forest models. However, the trees in the forest may violate Requirement 3 in Section~\ref{subsec:model_decision_trees}. If such violation happens on a tree $t$, we will use Lemma~\ref{lemma:equivalence_no_repeated_products} to find an equivalent tree $t'$ that satisfies Requirement 3 without increasing either the depth or the number of leaves, and replace $t$ by $t'$ in the forest. For convenience, we summarize the procedure in Lemma~\ref{lemma:equivalence_no_repeated_products} as the following algorithm.
	
	\begin{algorithm}
		{\SingleSpacedXI
			\caption{Modifying a tree of Requirements 1 and 2 to satisfy Requirement 3 in Section~\ref{subsec:model_decision_trees}}. \label{alg:check_requirement_3}
			\begin{algorithmic}[1]
				\Procedure{TreeModification}{a tree $t$ that satisfies Requirements 1 and 2}
				\While{there exists a leaf $\ell \in \leaves(t)$ violates Requirement 3}
				\State Set $path(\ell) \gets \{ s_1,s_2,\ldots,s_{d-1}  \}$ (see the definition in the proof of Lemma~\ref{lemma:equivalence_no_repeated_products}).
				\State Set $i \gets \arg\max_{ k \in \Ncal} \sum_{j=1}^{d-1} \mathbb{I}\{ k = x_{s_j}  \}.$
				\State Set $\{ u_1 < u_2 < \ldots < u_{e_i} \} \gets \{  j   \mid x_{s_j} = i  , j \in \{ 1,2,\ldots,d-1 \}  \}$.
				\If{$s_{u_1} \in \textbf{LS}(\ell)$}
				\State Remove all right subtrees branched at $s_{u_2},\ldots,s_{u_{e_i}}$ from tree $t$.
				\Else
				\State Remove all left subtrees branched at $s_{u_2},\ldots,s_{u_{e_i}}$ from tree $t$.
				\EndIf
				\For{$w = e_i,e_i-1,\ldots,2$}
				\State Delete split node $s_w$.
				\State Set the remaining child node of $s_w$ as the child node of $s_{w-1}$.
				\EndFor
				\EndWhile
				\State \Return $t$
				\EndProcedure
			\end{algorithmic}
		}
	\end{algorithm}

Our next result, Lemma~\ref{lemma:perfect_fitting_m1_d2} states that a data set of a single assortment can be perfectly fit by a depth 2 decision forest.

\begin{lemma}
	\label{lemma:perfect_fitting_m1_d2} 
	For any dataset $\Scal$ consisting of only one assortment $S$, there exists a forest $F$ of depth $2$ and of leaf complexity $2$ and a probability distribution $\lambdab$ over $F$ such that $\Pb^{(F,\lambdab)}(o \mid S) = v_{o,S}$ for every $o \in \Ncal^+$. 
\end{lemma}

\proof{Proof of Lemma~\ref{lemma:perfect_fitting_m1_d2}:} Consider a forest $F$ of depth $2$ such that $F = \{ t_1,t_2, \dots , t_N, t_0  \}$, where each tree is as shown in Figure~\ref{fig:randomized_base_case}, and define the probability distribution $\lambdab$ so that $\lambda_{t_o} = v_{o,S}$ for each $o \in \Ncal^+$; by construction, $\sum_{t \in F} \lambda_t = 1$ and $\lambda_t \geq 0$ for each $t \in F$. For this forest, each option $o \in \Ncal^+$ is chosen by exactly one tree, $t_o$, and the probability mass of that tree is $v_{o,S}$, which establishes that $\Pb^{(F, \lambdab)}(o \mid S) = v_{o,S}$ for all $o \in \Ncal^+$.  \hfill \Halmos

\begin{figure}[h!]
	\centering
	\includegraphics[width=8cm]{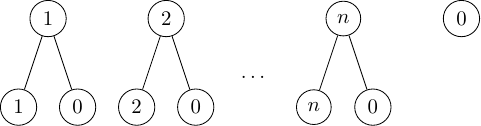}
	\caption{Trees $t_1,t_2, \ldots, t_{N}$ and $t_0$ (shown from left to right) for the forest $F$ described in Lemma \ref{lemma:perfect_fitting_m1_d2}}
	\label{fig:randomized_base_case}
\end{figure}

\begin{lemma}
	\label{lemma:perfect_fitting_add_depth_add_leaves}
	Consider two sets of assortments $\Scal_1$ and $\Scal_2$ satisfying the following two conditions: 
	\begin{enumerate}
		\item[(1)] For $i = 1,2$, there exists a forest $F_i$ of depth at most $d_i$ and of leaf complexity at most $\NumLeaves_i$, and a probability distribution $\lambdab^{(i)}$ such that $\Pb^{(F_i,\lambdab^{(i)})}(o \mid S) = v_{o,S}$ for all $S \in \Scal_i$; and 
		\item[(2)] There exists a product $p$ such that $p \in S$ for all $S \in \Scal_1$ and $p \notin S$ for all $S \in \Scal_2$.
	\end{enumerate}
	
	Then there exists a probability distribution $\lambdab$ and a forest $F$ of depth at most $1 + \max\{d_1,d_2\}$ and of leaf complexity at most $\NumLeaves_1 + \NumLeaves_2$ such that $\Pb^{(F,\lambdab)}(o \mid S) = v_{o,S}$ for all $S \in \Scal_1 \cup \Scal_2$.
\end{lemma}

\proof{Proof of Lemma~\ref{lemma:perfect_fitting_add_depth_add_leaves}:} We prove the statement by constructing an appropriate forest $F$ and a probability distribution $\lambdab$ such that $\Pb^{(F,\lambdab)}(o \mid S) = v_{o,S}$ for all $S \in \Scal_1 \cup \Scal_2$. For $i = 1,2$, let us denote the trees in forest $F_i$ by $t^{(i)}_1,t^{(i)}_2,\ldots,t^{(i)}_{n_i}$ and the corresponding probability distribution by $\lambdab^{(i)} = (\lambda^{(i)}_1, \dots \lambda^{(i)}_{n_i})$. Let us construct the forest $F$ and probability distribution $\lambdab$ for $\Scal_1 \cup \Scal_2$ as follows. We define the forest $F$ as
\begin{equation*}
F = \{ t_{\alpha,\beta}  \mid \alpha \in \{1,2,\ldots,n_1\}, \beta \in \{1,2,\ldots,n_2\} \},
\end{equation*}
where each tree $t_{\alpha,\beta} $ is formed by placing product $p$ at the root node, placing $t^{(1)}_\alpha$ as the left subtree of the root node, and $t^{(2)}_\beta$ as the right subtree of the root node. 
For the probability distribution $\lambdab$ over $F$, we set the probability of each tree $\lambda_{\alpha,\beta} = \lambda^{(1)}_\alpha \cdot \lambda^{(2)}_\beta$. By construction, $\lambdab$ is nonnegative, and adds up to 1, since 
\begin{equation*}
\sum_{\alpha = 1}^{n_1} \sum_{\beta = 1}^{n_2} \lambda_{\alpha,\beta} = \sum_{\alpha = 1}^{n_1} \sum_{\beta = 1}^{n_2} \lambda^{(1)}_{\alpha} \cdot \lambda^{(2)}_\beta = \left( \sum_{\alpha = 1}^{n_1} \lambda^{(1)}_\alpha \right) \left( \sum_{\beta = 1}^{n_2} \lambda^{(2)}_\beta \right) =  1.
\end{equation*}

We now show that $(F, \lambdab)$ ensures that $\Pb^{(F,\lambdab)}(o \mid S) = v_{o,S}$ for all $S \in \Scal_1 \cup \Scal_2$. For any $S \in \Scal_1$, we know that $p \in S$, and thus each purchase decision tree $t_{\alpha,\beta} \in F$ will immediately take the left branch at the root node. This implies that the purchase decision of tree $t_{\alpha,\beta}$ will be exactly the same as its left subtree $t^{(1)}_{\alpha}$ when any $S \in \Scal_1$ is given. Thus, for any $S \in \Scal_1$ and $o \in \Ncal^+$:
\begin{align*}
\Pb^{(F,\lambdab)}(o \mid S) & = \sum_{\alpha =1}^{n_1} \sum_{\beta =1}^{n_2} \lambda_{\alpha,\beta} \cdot \Ibb\{ o = \hat{A}(S, t_{\alpha,\beta}) \} \\
& = \sum_{\alpha =1}^{n_1} \sum_{\beta =1}^{n_2} \lambda^{(1)}_\alpha \cdot \lambda^{(2)}_\beta \cdot \Ibb\{ o = \hat{A}(S, t^{(1)}_\alpha) \} \\
& = \sum_{\alpha =1}^{n_1} \lambda^{(1)}_\alpha \cdot \Ibb\{ o = \hat{A}(S, t^{(1)}_\alpha) \}  = \Pb^{(F_1, \lambdab^{(1)})}(o \mid S) = v_{o,S},
\end{align*}
where we recall that $\hat{A}(S, t)$ is the option chosen by tree $t$ when given assortment $S$. 

Similarly, we can also establish that for any $S \in \Scal_2$ and $o \in \Ncal^+$, each tree $t_{\alpha,\beta}$ will make the same purchase decision as $t^{(2)}_{\beta}$, and so $\Pb^{(F,\lambdab)}(o \mid S) = \Pb^{(F_2, \lambdab^{(2)})}(o \mid S) = v_{o,S}$. This establishes that $(F,\lambdab)$ satisfies $\Pb^{(F,\lambdab)}(o \mid S) = v_{o,S}$ for all $S \in \Scal_1 \cup \Scal_2$.

With regard to the depth of $F$, we observe that each tree in $F$ is built by adding one level to trees from $F_1$ and $F_2$, and so the forest $F$ will be of depth at most $1 + \max\{d_1, d_2\}$. With regard to the leaf complexity of $F$, each tree in $F$ is built by combining two subtrees, where one has at most $\NumLeaves_1$ leaves and the other has at most $\NumLeaves_2$ leaves, so the forest $F$ will have at most $\NumLeaves_1 + \NumLeaves_2$ leaves. Finally, trees in $F$ may violate Requirement 3 in Section~\ref{subsec:model_decision_trees}. In that case, we apply the procedure in Lemma \ref{lemma:equivalence_no_repeated_products} (Algorithm~\ref{alg:check_requirement_3}) to find equivalent trees without increasing either the depth or leaf complexity of the forest.\hfill \Halmos

\proof{Proof of Theorem~\ref{thm:master_theorem_complexity}:}  With regard to depth and leaf complexity, we prove the statement by induction on the number of assortments $M$. The base case is established by Lemma~\ref{lemma:perfect_fitting_m1_d2}. Assume the statement holds for all integers $k < M$. With $M$ historical assortments $S_1,S_2,\ldots,S_M$, let $p$ be a product included in at least one assortment and meanwhile not included in all assortments. Such a $p$ must exist, otherwise $S_1 = S_2 = \ldots = S_M$, which violates the requirement of distinct assortments. Denote $\Scal_p = \{ S \mid S \in \Scal, p \in S  \}$ as the collection of historical assortments that include product $p$, and $\Scal^c_p = \{ S \mid S \in \Scal, p \notin S  \}$ as the collection of historical assortments that do not include product $p$. We further denote their cardinalities as $M_p = |\Scal_p|$ and $M^c_p = |\Scal^c_p|$.

Note that $1 \leq M_p\leq M-1$ and $1 \leq M^c_p\leq M-1$ by the definition of product $p$. To prove the inductive step, we assume that there exists a forest $F_p$ of depth at most $M_p + 1$ and of leaf complexity at most $2 M_p$, and a distribution $\lambdab_p$ such that $\Pb^{(F_p,\lambdab_p)}(o \mid S) = v_{o,S}$ for all $o \in \Ncal^+$ and $S \in \Scal^p$. Similarly, we also assume there exists a forest $F^c_p$ of depth at most $M^c_p + 1$ and of leaf complexity at most $2 M^c_p$, and a distribution $\lambdab^c_p$ such that $\Pb^{(F^c_p,\lambdab^c_p)}(o \mid S) = v_{o,S}$ for all $o \in \Ncal^+$ and $S \in \Scal^c_p$. By Lemma \ref{lemma:perfect_fitting_add_depth_add_leaves}, there exists a distribution $\lambdab$ and a forest $F$ of depth at most $1 + \max \{M_p + 1, M^c_p + 1 \} \leq 1 + M$ and leaf complexity at most $2 M_p + 2M^c_p = 2M$ such that $\mathbf{P}^{(F,\lambdab)} = v_{o,S}$ for $S \in \Scal$.

Let $(F,\lambdab)$ be the corresponding forest model. We can further use the procedure described in Lemma \ref{lemma:equivalence_no_repeated_products} (Algorithm~\ref{alg:check_requirement_3}) to ensure that for any tree $t \in F$, any path from root to a leaf will not encounter the same product twice on split nodes. Since applying the procedure described in Lemma \ref{lemma:equivalence_no_repeated_products} (Algorithm~\ref{alg:check_requirement_3}) would not increase depth and leaf complexity, the resulting trees will again have depth at most $\min \{ M+1, N+1 \}$ and leaf complexity again at most $2M$.

Let $D^* = \min \{ M+1, N+1 \}$ and let $F_{D^*,2M}$ be the collection of all decision trees of depth at most $D^*$ and of at most $2M$ leaves. Obviously, $F_{D^*,2M}$ is a finite set. By the above induction proof, we know that the following constraint system has a solution:
\begin{subequations}
	\begin{alignat}{2}
	& & &	\sum_{t \in F_{D^*,2M}} \Ab_{t,S} \lambda_t = \vb_S, \quad \forall \ S \in \Scal,\\
	& & &	\oneb^T \lambdab = 1,\\
	& & &	\lambdab \geq \zerob. 
	\end{alignat}
	\label{prob:feasibility_2}
\end{subequations}
The set defined by \eqref{prob:feasibility_2} is a polyhedron in standard form and is non-empty. Therefore, by standard linear optimization results (e.g., Corollary 2.2 of Bertsimas and Tsitsiklis 1997) there exists a basic feasible solution $\lambdab^*$ to \eqref{prob:feasibility_2}, which possesses the property that $\lambda^*_t \geq 0$ for $M(N+1) + 1$ trees $t \in F_{D^*,2M}$ and $\lambda^*_t = 0$ for all other trees, where $M(N+1) + 1$ is the number of equality constraints in \eqref{prob:feasibility_2}. Defining $F$ as $F = \{ t \in F_{D^*,2M} \mid \lambdab^*_t > 0\}$ and $\lambdab = ( \lambda^*_t )_{t \in F_{D^*,2M}}$, we obtain the required decision forest model. \hfill \Halmos

\subsection{Proof of Theorem~\ref{thm:asymptotic_size_of_forest}}
\label{subsec:appendix_proof_thm:asymptotic_size_of_forest}

\subsubsection*{Proof Strategy}

Before diving into the proof, we first demonstrate the basic idea of the proof with a simple example, and then provide an informal overview of the strategy of the proof.

\begin{exmp}
	Suppose $\Scal$ is a collection of $M = 128$ assortments with sufficiently large $N$. By Theorem~\ref{thm:master_theorem_complexity}, there exists a decision forest model of depth at most 129 that perfectly fits $\Scal$. Now, consider a product $p$ and the two subsets of $\Scal$ consisting of assortments that contain and do not contain $p$:
	\begin{align*}
	\Scal_p = \{ S \in \Scal \mid p \in S \}, \\
	\Scal^c_p = \{ S \in \Scal \mid p \notin S \}.
	\end{align*}
	Suppose that the product $p$ is such that $|\Scal_p| = 64$ and $|\Scal^c_p| = 64$. By invoking Theorem~\ref{thm:master_theorem_complexity}, we obtain separate decision forest models $(F_1, \lambdab_1)$ and $(F_2, \lambdab_2)$ that respectively fit $\Scal_p$ and $\Scal^c_p$ that are of depth 65. By invoking Lemma~\ref{thm:master_theorem_complexity}, we can combine the two models into a single decision forest model of depth at most $1+\max\{65,65\} = 66$ that perfectly fits the assortment collection $\Scal$. \hfill \Halmos 
\end{exmp}

What the above example illustrates is that when we can find a product $p$ that perfectly divides the assortment collection $\Scal$, we can actually fit a model where the depth is at most roughly $M/2$, instead of roughly $M$. In this example, we stopped after finding one product $p$ that perfectly splits $\Scal$. However, there is nothing preventing us from repeating the process again with $\Scal_p$ and $\Scal^c_p$. If we can repeat the same process again with each of $\Scal_p$ and $\Scal^c_p$ -- i.e., we find a product $p'$ that perfectly splits $\Scal_p$, and a product $p''$ (possibly different from $p'$) that perfectly splits $\Scal^c_p$ -- then we would be able to obtain a forest of depth at most $1 + \max\{ 1 + \max\{33, 33\}, 1 + \max\{33, 33\} \} = 35$ (roughly $M / 4$). 

We can keep repeating the process to obtain a smaller and smaller decision forest model; each time we can find such a splitting product, we reduce the size of the collections by half and we obtain roughly a factor of two reduction in the depth of the forest. This procedure naturally gives rise to a forest of depth $O(\log_2 M )$. 

The above procedure assumes that we are always able to find a product that perfectly splits a given subcollection of assortments, that is obtained after some rounds of splitting. If the collection of assortments $\Scal$ is drawn randomly, then this will not always be possible. In addition, this will also not be possible if a subcollection that we encounter contains an odd number of assortments. 

Instead of aiming to divide the collection of assortments $\Scal$ exactly in half by finding a ``perfect'' splitting product $p$, what we can instead hope to do is to split $\Scal$ almost evenly by finding a ``good'' splitting product $p$. To do this, we fix an $\epsilon \in (0,1)$, and consider the factor $1 / (2 - \epsilon)$, which is a number in the interval $(1/2, 1)$. The factor $1/(2 - \epsilon)$ defines how big the two subcollections, $\Scal_p$ and $\Scal^c_p$, should be relative to $\Scal$. In other words, we now look for a product $p$ such that $|\Scal_p| \leq M/(2 - \epsilon)$ and $|\Scal^c_p| \leq M / (2 - \epsilon)$. If we succeed in doing this, then we will reduce the size of the collection of assortments by a factor of $1 / (2 - \epsilon)$ with each splitting product we find, giving rise to a forest of depth $O( \log_{2 - \epsilon} M)$. 

The parameter $\epsilon$ controls a trade-off between the depth of the ultimate forest and the probability of being able to create that forest. When $\epsilon$ is small, the factor $1 / (2 - \epsilon)$ will be closer to 1/2, leading to a large reduction in the size of the subcollections and a small depth. However, the probability of finding a product that results in this split will be small. When we enlarge $\epsilon$, then probability of existence of finding a ``good'' splitting product $p$ increases, but this comes with a price, because the depth scales like $O(\log_{2-\epsilon} M)$, which is increasing in $\epsilon$. %

In the proof of Theorem~\ref{thm:asymptotic_size_of_forest}, we essentially use this idea to obtain a bound on the probability of finding a forest of depth at most $\log_{2-\epsilon}(M / M_0) + M_0 + 1$, where $M_0$ is an integer constant that defines a limit on how small the size of a subcollection of assortments can be before applying Theorem~\ref{thm:master_theorem_complexity}. Due to the recursive nature of how the splitting procedure is applied to repeatedly divide the collection of assortments, the probability bound satisfies a recursive inequality: for a given collection of assortments, the bound on the probability of finding a forest of depth at most $d$ that fits $\tilde{\Scal}$ is bounded by a quantity that involves the same probability bound but corresponds to a forest of depth at most $d - 1$ that fits a smaller subcollection of assortments.

\subsubsection*{Notation} Before we are able to prove the theorem, we require some additional definitions. First, as discussed above, we let $\epsilon \in (0,1)$ be an arbitrary constant and let $M_0 > 1$ be an arbitrary fixed positive integer. As alluded to above in \textbf{Proof Strategy}, the integer $M_0$ will later serve as a stopping point for the partitioning process. That is, when the current collection of assortments is of size $M_0$ or lower, we stop partitioning assortments and apply the depth bound provided by Theorem~\ref{thm:master_theorem_complexity}.

For convenience, we will also use the constant $k$ to denote $k = \epsilon^2 / (2 (2 - \epsilon)^2)$, and the constant $\beta$ to denote $\beta = 1 / (2 - \epsilon)$. Note that for all $\epsilon \in (0,1)$, $k > 0$ and $\beta \in (1/2, 1)$. The constant $k$ is a quantity that will appear later in our application of Hoeffding's inequality to bound the probability of finding a good splitting product. The constant $\beta$ is simply the reduction factor of $1/(2-\epsilon)$ that we desire for the splitting product (described above under \textbf{Proof Strategy}); in other words, when we split a collection of assortments, the cardinality of each subcollection should be at most a factor of $\beta$ of the parent collection. Both $k$ and $\beta$ are introduced to make the mathematical expressions we encounter later less cumbersome.

Given a number of assortments $M > M_0$, we define the integer $\bar{d}$ as 
\begin{equation}
\bar{d}(M, M_0, \epsilon) = \left\lceil \log_{2 - \epsilon} \left( \frac{M}{M_0} \right) \right\rceil.
\end{equation}
To understand the meaning of $\bar{d}$, observe that $\beta < 1$. The proof that we will present shortly relies on repeatedly dividing a collection of assortments by selecting a product $p$ such that the subcollection of assortments with $p$ and the subcollection that does not contain $p$ both have cardinality that is at most $\beta$ of the parent collection. The minimum number of such divisions needed to reach a collection of assortments of size $M_0$ or lower, starting with a collection of $M$ assortments, is exactly $\bar{d}$. For ease of exposition, we will suppress the arguments of $\bar{d}$, but it should be regarded as a function of the number of assortments in the data set $M$, as well as the constants $M_0$ and $\epsilon$.

We define the integer function $q(N, M, M_0)$ as 
\begin{equation}
q(N, M, M_0) = \left \lfloor N / \bar{d} \right\rfloor.
\end{equation}
The integer $q(N, M, M_0)$ is interpreted as the number of candidate splitting products that are considered at each level of splitting. The rationale for $q(N, M, M_0)$ is as follows. Since we assume that the $M$ assortments are drawn independently and uniformly from the collection of all $2^N$ assortments, then for each product $p' \in \Ncal$ and each assortment $m$, the presence of product $p$ in assortment $m$, $\Ibb\{p \in S_m\}$, is an independent Bernoulli(1/2) random variable. As described above in \textbf{Proof Strategy}, for a given collection of assortments, we need to find a product $p$ that splits that collection into two subcollections that are a factor of $1/(2-\epsilon)$ ($= \beta$) of the size of the parent collection. For a collection of $M'$ independent random assortments, where each product is included in each assortment independently with probability $1/2$, the probability that a single, \emph{fixed} product $p$ can split the collection in this way can be written as a binomial probability (i.e., the number $X_p$ of assortments in the collection of $M'$ assortments that contain product $p$ is a Binomial($M'$, $1/2$) random variable), and readily bounded using Hoeffding's inequality. If instead of considering a fixed product, we consider all $N$ products, then the probability of finding a product $p$ that can split the collection in this way increases. However, if we condition on the event that there exists a good splitting product among all $N$ products, then we can no longer guarantee that the random variables $\Ibb\{ p \in S \}$ are independent Bernoulli(1/2) random variables for any product $p$ and any assortment $S$ within either subcollection that is generated. This is problematic, because if the random variables $\Ibb\{p \in S\}$ are no longer independent Bernoulli(1/2) variables within the subcollections, then we cannot bound the probability of finding a product that splits each subcollection, and we cannot succeed in constructing our bound.

Thus, when we search for a good splitting product, we do not search over all $N$ products. Instead, we search over only $q(N, M, M_0)$ products. In that sense, when we condition on the existence of a good splitting product out of those $q(N, M, M_0)$ products, then we only ``contaminate'' the $q(N, M, M_0)$ products that we searched over, and we can protect the independent Bernoulli(1/2) nature of the remaining products. Since there are $\bar{d}$ levels of splitting, the size of the candidate splitting set should be such that we do not run out of products, i.e., we need $\bar{d} \cdot q(N, M, M_0) \leq N$. The choice of $q(N, M, M_0)$ as $\left \lfloor N / \bar{d} \right\rfloor$ gives us the largest possible size for the set of candidate splitting products without running out of products. As with $\bar{d}$, for ease of exposition we will suppress the arguments of $q$, but again, it is a quantity that depends on the number of assortments $M$, as well as $M_0$ and $\epsilon$.

Given $\epsilon$ and $M_0$, we define the function $g(d)$ as
\begin{equation}
g(d) = (2^{d-M_0 - 1} - 1)\cdot 2^{-q(kM_0 - 1)}. \label{eq:g_defn}
\end{equation}
The function $g(d)$ will later serve as a upper bound of probability.

\subsubsection*{Main Proof} The event in the statement of the theorem is that there exists a forest of a particular depth that fits a set of assortments that are sampled uniformly from the set of all assortments over a fixed number of products. For an arbitrary number of assortments $M' > 0$ and an arbitrary depth $d'$, let us define $R(M', d')$ to be the event that there exists a forest of depth at most $d'$ over the universe of $N$ products that fits a set of $M'$ assortments. We now need to carefully define the probability distribution with which we will measure the probability of this event and its complement. We will use $F$ to denote a distribution according to which a collection of $M'$ assortments are drawn. The probability $\Pbb_F( R(M',d')^c )$ is the probability that we do not succeed in finding a decision forest of depth at most $d'$ that fits $M'$ assortments sampled from $F$. Define $\Fcal(M', N')$ as the set of distributions over collections of $M'$ assortments of the $N$ products, such that at least $N'$ products are sampled independently with probability 1/2. (To ``sample a product $p$ independently with probability 1/2'' means to draw an independent $\text{Bernoulli}(1/2)$ random variable that is 1 if the product $p$ is to be included, and 0 if it is not to be included.) We then define the maximum failure probability as 
\begin{equation}
Q(M', N', d') = \sup_{F \in \Fcal(M', N') }  \Pbb_F( R(M', d')^c ).
\end{equation}

Note that in the statement of Theorem~\ref{thm:asymptotic_size_of_forest}, $M$ assortments are sampled uniformly at random from the set of all assortments, which is exactly the same as independently sampling each of the $N$ products with probability 1/2, i.e., each assortment is generated by drawing, for each $p \in \{1,\dots,N\}$, a $\text{Bernoulli}(1/2)$ variable that is 1 if product $p$ is to be included, and 0 if it is not included. The set of distributions $\Fcal(M, N)$ is thus a singleton consisting of exactly this distribution over $M$ assortments.

To prove Theorem~\ref{thm:asymptotic_size_of_forest}, we will prove a more general result concerning the maximum failure probability $Q$. Once we prove this general result, we will show that for specific choices of $\epsilon$ and $M_0$, the forest depth and the corresponding probability of the forest fitting the data will exhibit the asymptotic behavior stated in Theorem~\ref{thm:asymptotic_size_of_forest}. 

The general result we will prove is stated as follows:
\begin{theorem}
	Let $M > 0$, $N > 0$. Let $\epsilon \in (0,1)$ be a fixed constant and $M_0$ be a positive integer such that $M_0 < M$, and let $k$, $q$, $\bar{d}$ and $g(d)$ be defined as above. We then have, for any $M' \leq M$,
	\begin{equation}
	Q(M', N,\ \bar{d} + M_0 + 1) \leq g( \bar{d} + M_0 + 1) = (2^{\bar{d}} - 1) \cdot 2^{-q (k M_0 - 1) }.  \label{eq:Qbound_theorem}
	\end{equation}
	\label{theorem:shallow_general_result}
\end{theorem}

We will prove this result by induction. To set up the induction proof, we need to set up two auxiliary results. The first result, Lemma~\ref{lemma:shallow_base_case}, will serve to establish the base case for the induction proof.

\begin{lemma}
	For any positive integer $M' \leq M_0$, $N' \geq 0$, we have:
	\begin{equation}
	Q(M', N', M_0+1) \leq g(M_0 + 1).
	\end{equation}
	\label{lemma:shallow_base_case}
\end{lemma}
\proof{Proof of Lemma~\ref{lemma:shallow_base_case}:}
Let  $M'' \leq M' \leq M_0$ be the number of distinct assortments in the collection $\Scal'$ of $M'$ assortments. Theorem~\ref{thm:master_theorem_complexity} guarantees almost surely that there exists a forest of depth at most $M''+1$ such that $\Pbb^{(F,\lambdab)}(o \mid S) = v_{o,S}$ for every $o$ and $S \in \Scal'$. Since $M'' \leq M_0$, it immediately follows that the forest is of depth at most $M_0 + 1$. Thus, the maximum failure probability $Q(M', N', M_0 +1)$ will be equal to zero for any $M' \leq M_0$. Since the upper bound $g(M_0 + 1)$ is exactly zero (by the definition of $g$ in equation~\eqref{eq:g_defn}), the lemma follows.  \hfill \Halmos \\

The second auxiliary result, Lemma~\ref{lemma:shallow_inductive_step}, will serve to establish the induction hypothesis.

\begin{lemma}
	Let {$N' \geq q$}, $d \geq M_0 + 2$ and $M' > 0$. If the collection of inequalities
	\begin{equation}
	Q(M'', N' - q, d-1) \leq g(d-1), \quad \forall \ M'' \leq \lfloor M' \beta \rfloor \label{eq:induction_hypothesis}
	\end{equation}
	holds, then we have
	\begin{equation}
	Q(M', N', d) \leq g(d).
	\end{equation}
	\label{lemma:shallow_inductive_step}
\end{lemma}
\proof{Proof of Lemma~\ref{lemma:shallow_inductive_step}:}
First, let us handle the case when $M' \leq M_0$. Note that in this case, by exactly the same reasoning as in Lemma~\ref{lemma:shallow_base_case}, we automatically have $Q(M', N', d) \leq g(d)$: we apply Theorem~\ref{thm:master_theorem_complexity} to obtain a forest of depth at most $M'+1$, and since $d \geq M_0+2 \geq M'+1$, this forest is automatically of depth at most $d$ as well. This establishes that $Q(M', N',d) = 0$, and since $g(d) \geq 0$, the statement follows, without any use of the hypothesis~\eqref{eq:induction_hypothesis}. Thus, in what follows, we will focus on the case when $M' > M_0$.

Let $F$ be any distribution from $\Fcal(M', N')$. Let $\Scal$ be the set of $M'$ assortments drawn from $F$. Fix a set of products $\splitterset$ of size $q$ from the set of products of size at least $N'$ that are known to be independent. Define the set $\splitterset^*$ as 
\begin{equation}
\splitterset^* = \left\{ p \in \splitterset \   \vline \  \begin{array}{l} M'(1 - \beta) \leq | \Scal_p | \leq M' \beta, \\  M'(1 - \beta) \leq |\Scal^c_p| \leq M' \beta \end{array} \right\},
\end{equation}
where the collections $\Scal_p$ and $\Scal^c_p$ are defined as in the proof of Theorem~\ref{thm:master_theorem_complexity}:
\begin{equation}
\Scal_p = \{ S \in \Scal \ \vline \ p \in S\}, \label{eq:left_assortment_set}
\end{equation}
\begin{equation}
\Scal^c_p = \{ S \in \Scal \ \vline \ p \notin S \}.
\end{equation}
In words, $\Scal_p$ is the collection of assortments that include product $p$, while $\Scal^c_p$ is the collection of assortments that do not include product $p$. The set $\splitterset^*$ is the set of all products $p$ that essentially divide the collection of assortments $\Scal$ in a ``balanced'' way, such that the resulting collections of assortments $\Scal_p$ and $\Scal^c_p$ contain at least $(1-\beta)$ fraction of the assortments and at most $\beta$ fraction of the assortments. That is to say, such $p \in \splitterset^*$ will be a ``good'' splitting product, as described in {\bf Proof Strategy}.

With $\splitterset^*$ defined, let us define the product $p^*$ as 
\begin{equation*}
p^* = \left\{ \begin{array}{ll} \min_{p \in \splitterset} p & \text{if}\ \splitterset^* = \emptyset, \\
\min_{p \in \splitterset^*} p & \text{if}\ \splitterset^* \neq \emptyset. \end{array} \right.
\end{equation*}
In words, the product $p^*$ is the lowest index product from $\splitterset^*$ if the latter turns out to not be empty, and otherwise it is the lowest index product from $\splitterset$. Both $p^*$ and $\splitterset^*$ are random. Note that the definition of $p^*$ when $\splitterset^*$ is empty is not important for the proof; it is just needed to ensure that some events we will construct shortly are well-defined. 

Having defined $p^*$, let us now define the following events:
\begin{itemize}
	\item $A$: the event that $\splitterset^* \neq \emptyset$. 
	\item $B_1$: the event that there exists a decision forest model $(F_1, \lambdab_1)$ of depth at most $d-1$ such that $\Pb^{(F_1,\lambdab_1)}(o \mid S) = v_{o,S}$ for all $o$ and $S \in \Scal_{p^*}$. 
	\item $B_2$: the event that there exists a decision forest model $(F_2, \lambdab_2)$ of depth at most $d-1$ such that $\Pb^{(F_2, \lambdab_2)}(o \mid S) = v_{o,S}$ for all $o$ and $S \in \Scal^c_{p^*}$. 
\end{itemize}

Observe that if all three events hold, then Lemma~\ref{lemma:perfect_fitting_add_depth_add_leaves} guarantees that there exists a decision forest model $(F, \lambdab)$ of depth at most
\begin{equation*}
1 + \max\{ d-1, d-1\}  = d
\end{equation*}
such that $\Pb^{(F,\lambdab)}(o \mid S) = v_{o,S}$ for all $o$ and $S \in \Scal$. In other words, if the events $A$, $B_1$ and $B_2$ occur, then the event $R(M', d)$ occurs. 

By taking the contrapositive of the above statement, we can bound the probability $\Pbb_F( R(M', d)^c )$ as
\begin{align}
\Pbb_F( R(M', d)^c ) & \leq \Pbb_F( \ (A \cap B_1 \cap B_2)^c \ ) \nonumber \\
& = \Pbb_F( A^c \cup B^c_1 \cup B^c_2 ) \nonumber \\
& = \Pbb_F( A^c) + \Pbb_F( A \cap (B^c_1 \cup B^c_2) ) \nonumber \\ 
& \leq \Pbb_F(A^c) + \Pbb_F( B^c_1 \cup B^c_2 \mid A) \nonumber \\
& \leq \Pbb_F(A^c) + \Pbb_F(B^c_1 \mid A) + \Pbb_F(B^c_2 \mid A),  \label{eq:QdecompositionAB1B2}
\end{align}
where the fourth step follows by the definition of conditional probability and the last step follows by the union bound. 

At this stage, we pause to comment on the utility of bounding $\Pbb_F( R(M', d')^c )$ using the events $A$, $B_1$ and $B_2$, as in inequality~\eqref{eq:QdecompositionAB1B2}. The event $A$ is useful because it is defined in terms of the product set $\splitterset$. Each product $p$ in $\splitterset$ is such that $p$ exists in an assortment in $\Scal$ with probability 1/2, independently across the $M'$ assortments and independently of any other product; we will see shortly that this allows us to conveniently bound the probability of $A^c$. Second, the conditional events $B^c_1 \mid A$ and $B^c_2 \mid A$ are events that bear a resemblance to $R(M', d)^c$: both $B^c_1 \mid A $ and $B^c_2 \mid A$ are events in which we fail to find a decision forest model that fits the assortment sets $\Scal_{p^*}$ and $\Scal^c_{p^*}$, respectively. The difference is that while the distribution $F$ is such that at least $N'$ products are sampled independently with probability 1/2 to generate $M'$ assortments, the distribution according to which $\Scal_{p^*}$ and $\Scal^c_{p^*}$ are sampled conditional on $A$ is such that at least $N' - q$ products are sampled independently with probability 1/2 to generate a random number of assortments that is at most $\beta M'$. We will later leverage this similarity to invoke the induction hypothesis and obtain a bound on $\Pbb_F(B^c_1 \mid A)$ and $\Pbb_F(B^c_2 \mid A)$.

Our goal now will be to bound each of the three terms in the inequality~\eqref{eq:QdecompositionAB1B2}. For $\Pbb(A^c)$, observe that we can write this event as
\begin{equation*}
\Pbb_F( A^ c) = \Pbb_F \left( \bigcap_{p \in \splitterset} \left\{  |\Scal_p| > M'\beta \ \text{or} \ |\Scal_p| < M' (1-\beta) \right\} \right).
\end{equation*}
By the assumption that $F \in \Fcal(M', N')$ and that $\splitterset$ is a subset of those products which are sampled independently, the random variables $\Ibb\{ p \in S\}$ for each $p \in \splitterset$ and $S \in \Scal$ are independent Bernoulli(1/2) random variables. The size of the subcollection $\Scal_p$ can be written as $|\Scal_p| = \sum_{S \in \Scal} \Ibb\{ p \in S\}$; thus, the random variables $\{ |\Scal_p| \}_{p \in \splitterset}$ are distributed as independent $\text{Binomial}(M', 1/2)$ random variables. Letting $X_p$ denote each such binomial random variable, we can bound the probability $\Pbb_F(A^c)$ as 
\begin{align*}
\Pbb_F(A^c) & = \Pbb_F \left( \bigcap_{p \in \splitterset} \left\{ X_p > M' \beta \ \text{or}\ X_p < M' (1-\beta) \right\} \right) \\
& = \prod_{p \in \splitterset} \Pbb \left(  X_p > M' \beta \ \text{or}\ X_p < M' (1-\beta) \right) \\
& \leq \prod_{p \in \splitterset}  (2e^{- M' k} )\\
& = 2^q e^{-M' k q} \\
& \leq 2^{-q (M' k - 1)} \\
& \leq 2^{-q (M_0 k - 1)}
\end{align*}
where the second step follows by the independence of the $X_p$ random variables; the third step follows by Hoeffding's inequality for a sum of $M'$ Bernoulli random variables and recognizing that $M' \beta - M'/2 = M'/2 - M'(1-\beta) = M'\epsilon / (2 (2-\epsilon))$; the fourth step follows by the fact that $|\splitterset| = q$; the fifth step follows by the fact that $2 < e$ and algebra; and the last step by the fact that $q > 0$ and $M_0 < M'$. Before continuing, we draw the reader's attention to the dependence of this bound on $q$: the bound becomes exponentially smaller with $q$. In words, when we search over a larger set of candidate splitting products, it becomes easier to find a product that splits $\Scal$ in a balanced way. Ideally, we would search over all $N'$ products, instead of only $q$ products; however, as we discussed earlier under {\bf Notation} when defining the constant $q$, we have to limit the size of $\splitterset$ to ensure that we preserve independent Bernoulli(1/2) products for later stages of splitting. %

To bound $\Pbb_F(B^c_1 \mid A)$ and $\Pbb_F(B^c_2 \mid A)$, let us first define two conditional random variables, $\Gamma$ and $\Gamma^c$, as 
\begin{align}
& \condsetsize  = |\Scal_{p^*}| \phantom{\prod} \hspace{-0.75em}  \vline \ A, \\
& \condsetsize^c  = |\Scal^c_{p^*}| \phantom{\prod} \hspace{-0.75em} \vline \ A, 
\end{align}
i.e., $\condsetsize$ is the number of assortments containing $p^*$ given that $A$ occurs, while $\condsetsize^c$ is the number of assortments not containing $p^*$ given that $A$ occurs. We can now write $\Pbb_F(B^c_1 \mid A)$ and $\Pbb_F(B^c_2 \mid A)$ by conditioning and de-conditioning on $\condsetsize$ and $\condsetsize^c$ respectively:
\begin{align}
\Pbb_F( B^c_1 \mid A) & = \Exp_{\condsetsize} \left[ \Pbb_F(B^c_1 \mid A, \condsetsize ) \right], \\
\Pbb_F( B^c_2 \mid A) & = \Exp_{\condsetsize^c } \left[ \Pbb_F(B^c_2 \mid A, \condsetsize^c ) \right],
\end{align}
where the expectations are taken with respect to the conditional random variables $\condsetsize$ and $\condsetsize^c$ respectively. 

To understand how we will next proceed, let us focus on $\Pbb_F(B^c_1 \mid A)$. Let $\tilde{M}$ be a given realization of $\condsetsize$ and consider the conditional probability $\Pbb_F( B^c_1 \mid A, \condsetsize = \tilde{M} )$. We now claim that:
\begin{align}
\Pbb_F( B^c_1 \mid A, \condsetsize = \tilde{M} ) & = \Pbb_{\tilde{F}}( R( \tilde{M}, d-1)^c ) \\
& \leq Q( \tilde{M} , N' - q, d-1),
\end{align}
where $\tilde{F}$ is the distribution over collections of $\tilde{M}$ assortments that is induced by conditioning on the event $A$ and the event $\condsetsize = \tilde{M}$. (To actually sample from such a distribution, one can repeatedly sample $M'$ assortments according to $F$, discard those draws of the $M'$ assortments that do not satisfy both $A$ and $|\Scal_{p^*}| = \tilde{M}$, and return each collection $\Scal_{p^*}$ for the remaining draws.) The first step in the above follows from the definition of $\tilde{F}$. The second step follows by the fact that, by definition, $\tilde{F}$ is a member of $\Fcal(\tilde{M}, N' - q)$. To understand why, observe that after conditioning on $A$, which is an event that involves a set of products of size $q$ from the set of at least $N'$ products that are known to be independent with respect to $F$, the distribution $\tilde{F}$ may be such that the independence of the $q$ products is no longer guaranteed. However, after conditioning in this way, we know that there still remain at least $N' - q$ products that are independent, because they have not yet been used in any way (we have not conditioned on any event that involves these products). Thus, we obtain the upper bound of $Q( \tilde{M}, N' - q, d- 1)$. This observation is critical, because it is what ultimately allows us to link $Q(M', N', d)$ to $Q(M'', N' - q, d-1)$, and bound $Q(M', N', d)$ via the induction hypothesis.

With this insight in hand, let us continue with bounding $\Pbb_F( B^c_1 \mid A)$. We have
\begin{align}
\Pbb_F( B^c_1 \mid A) & = \Exp_{ \condsetsize } \left[ \Pbb( B^c_1 \mid A, \condsetsize ) \right] \\
& \leq \Exp_{ \condsetsize } \left[  Q( \condsetsize, N' - q, d-1) \right] \\
& \leq g(d-1) \\
& = (2^{d - M_0 - 2} - 1) 2^{-q (kM_0 - 1) },
\end{align}
where the first inequality follows by our reasoning above; the second inequality follows by our induction hypothesis~\eqref{eq:induction_hypothesis}, and the fact that the integer random variable $\condsetsize$ is almost surely bounded by $\lfloor M' \beta \rfloor$; and the last step follows by the definition of $g$. 

Applying the same steps for $\Pbb_F( B^c_2 \mid A)$, allows us to also conclude that
\begin{equation}
\Pbb_F( B^c_2 \mid A) \leq  (2^{d - M_0 - 2} - 1) 2^{-q (kM_0 - 1) }.
\end{equation}

Now that we have constructed a bound for $\Pbb_F(A)$, $\Pbb_F( B^c_1 \mid A)$ and $\Pbb_F( B^c_2 \mid A)$, we can return to completing the bound in \eqref{eq:QdecompositionAB1B2}. We have:
\begin{align}
\Pbb_F( R(M', d)^c ) & \leq \Pbb_F( A^c) + \Pbb_F( B^c_1 \mid A) + \Pbb_F( B^c_2 \mid A) \\
& \leq 2^{-q (kM_0 -1)} + 2( 2^{d - M_0 - 2} - 1) 2^{-q(kM_0 - 1) } \\
& =  \left( 1 + 2^{d-M_0 -1} - 2 \right) \cdot 2^{-q(kM_0 -1)} \\
& = (2^{d-M_0 - 1} - 1) \cdot 2^{-q(kM_0 -1)} \\
& = g(d).
\end{align}
Since our choice of the starting distribution $F$ was arbitrary, we have that $\Pbb_F( R(M', d)^c)$ is upper bounded by $g(d)$ for any $F$ in $\Fcal(M', N')$; since $Q(M', N', d)$ is defined as the supremum of this probability over all distributions $F$ in $\Fcal(M', N')$, we thus have
\begin{equation}
Q(M', N', d) \leq g(d),
\end{equation}
as required. \hfill \Halmos \\

Having established Lemmas~\ref{lemma:shallow_base_case} and \ref{lemma:shallow_inductive_step}, we put these two results together to establish Theorem~\ref{theorem:shallow_general_result}. 

\proof{Proof of Theorem~\ref{theorem:shallow_general_result}:}
For $i = 0, 1, \dots, \bar{d}$, we define the quantity $\mu_i$ as follows:
\begin{align*}
\mu_0 & = M, \\
\mu_i & = \lfloor \beta \cdot \mu_{i-1}  \rfloor, \ i = 1, \dots, \bar{d}.
\end{align*}
It is straightforward to see that $\mu_i \leq M \beta^i$ for $i = 1, \dots, \bar{d}$.

We now show that the following collection of inequalities holds:
\begin{align}
& Q(M', N - \bar{d}q, M_0 + 1) && \leq g(M_0 + 1), \quad \forall M' \leq \mu_{\bar{d}},  \tag{Q-0} \label{ineq:Qbound_0}\\
& Q(M', N - (\bar{d} - 1)q, M_0 + 2) && \leq g(M_0 + 2), \quad \forall M' \leq \mu_{\bar{d} - 1}, \tag{Q-1} \label{ineq:Qbound_1}\\
& Q(M', N - (\bar{d} - 2)q, M_0 + 3)  && \leq g(M_0 + 3), \quad \forall M' \leq \mu_{\bar{d} - 2}, \tag{Q-2} \label{ineq:Qbound_2}\\ 
& \vdots &  \nonumber \\
& Q(M', N - (\bar{d} - i+1)q, M_0 + i)  && \leq g(M_0 + i ), \quad \forall M' \leq \mu_{\bar{d} - i + 1}, \tag{Q-($i-1$)} \label{ineq:Qbound_iminus1}\\
& Q(M', N - (\bar{d} - i)q, M_0 + i + 1)  && \leq g(M_0 + i + 1), \quad \forall M' \leq \mu_{\bar{d} - i},  \tag{Q-$i$} \label{ineq:Qbound_i}\\
& \vdots \nonumber \\ 
& Q(M', N - q, M_0 + \bar{d} )  && \leq g(M_0 + \bar{d}), \quad \forall M' \leq \mu_1, \tag{Q-($\bar{d}-1$)} \label{ineq:Qbound_bardm1}\\ 
& Q(M', N, M_0 + \bar{d} + 1 )  && \leq g(M_0 + \bar{d} + 1), \quad \forall M' \leq \mu_0. \tag{Q-$\bar{d}$} \label{ineq:Qbound_bardm}
\end{align}

The first inequality \eqref{ineq:Qbound_0} holds because $N - \bar{d}q \geq 0$ by the definition of $q$, so we can invoke Lemma~\ref{lemma:shallow_base_case} to guarantee that $Q(M', N - \bar{d} q, M_0 + 1) \leq g(M_0 + 1)$ holds for any $M' \leq M_0$. Note that the inequality holds for the range $M' \leq \mu_{\bar{d}}$, because $\mu_{\bar{d}} \leq M_0$ (this is a consequence of $\mu_{\bar{d}} \leq M\beta^{\bar{d}}$ and the definition of $\bar{d}$ as $\lceil \log_{2-\epsilon}(M / M_0) \rceil$).

For $i = 1, \dots, \bar{d}$, suppose that \eqref{ineq:Qbound_iminus1} holds. For any $M' \leq \mu_{\bar{d} - i}$, observe that the inequality of  \eqref{ineq:Qbound_i} will hold for $M'$ because $\lfloor \beta M' \rfloor \leq \lfloor \beta \mu_{\bar{d} - i} \rfloor = \mu_{\bar{d} - i + 1}$ and because of Lemma~\ref{lemma:shallow_inductive_step}. Thus,  \eqref{ineq:Qbound_iminus1} implies  \eqref{ineq:Qbound_i}.

Since  \eqref{ineq:Qbound_0} holds and  \eqref{ineq:Qbound_iminus1} implies  \eqref{ineq:Qbound_i} for $i = 1,\dots, \bar{d}$, it follows by induction that \eqref{ineq:Qbound_bardm} holds. Theorem~\ref{theorem:shallow_general_result} follows because the desired inequality~\eqref{eq:Qbound_theorem} is contained in the inequalities of \eqref{ineq:Qbound_bardm}.  \hfill \Halmos \\

Having proved Theorem~\ref{theorem:shallow_general_result}, we are now in a position to prove Theorem~\ref{thm:asymptotic_size_of_forest}. 

\proof{Proof of Theorem~\ref{thm:asymptotic_size_of_forest}:}
The statement of Theorem~\ref{theorem:shallow_general_result} allows us to use any $M_0$ and $\epsilon$; let us fix $M_0 = 20$ and $\epsilon = 0.5$. 

With these choices for $M_0$ and $\epsilon$, we will now simplify the probability bound in \eqref{eq:Qbound_theorem}. We define the constant $\xi$ as $\xi = \log_{2 - \epsilon} 2 = \log_{1.5} 2$. We observe that 
\begin{align}
\bar{d} & = \lceil \log_{2-\epsilon}( M / M_0 ) \rceil \nonumber \\
&  = \lceil \log_{1.5} M - \log_{1.5} M_0 \rceil \nonumber \\
& \leq \log_{1.5} M \nonumber \\
& = \xi \log_2 M, \label{eq:final_bound_a}
\end{align}
where the inequality follows because $\log_{1.5} 20 > 1$. Thus, the forest depth, which is at most $M_0 + 1 + \bar{d}$, is of order $O( \log_2 M)$. In addition, we have that
\begin{align}
(2^{\bar{d}} - 1) & \leq 2^{\xi \log_2 M} = M^\xi  \leq M^2,  \label{eq:final_bound_b}
\end{align}
which follows because $\xi \leq 2$. Lastly, we have
\begin{align}
2^{-q(kM_0 - 1)} & = 2^{-  \lfloor N / \bar{d} \rfloor \cdot (0.111)} \nonumber \\
& \leq 2^{- (N / \bar{d} - 1) (0.111) } \nonumber\\
& \leq 2^{-  (N/ (\xi \log_2 M) - 1) ( 0.111) } \nonumber \\
& = 2^{0.111} \cdot 2^{ - (0.111 / 1.710) (N / \log_2 M) } \nonumber \\
& = 2^{0.111} \cdot 2^{ - 0.065 N / \log_2 M } \label{eq:final_bound_c} 
\end{align}
which is of order $O(2^{-CN / \log_2 M})$, where $C$ is a positive constant. Putting together \eqref{eq:final_bound_b} and \eqref{eq:final_bound_c}, we thus obtain a bound on $Q(M, N, \bar{d} + M_0 + 1)$, which is of order $O(M^2 \cdot 2^{-CN / \log_2 M})$. The statement of the theorem thus follows. Finally, the forest may contain trees that violate Requirement 3 in Section~\ref{subsec:model_decision_trees}. For any such tree, we apply the procedure in Lemma \ref{lemma:equivalence_no_repeated_products} (Algorithm~\ref{alg:check_requirement_3}) to obtain an equivalent tree without increasing the depth and leaf complexity of the overall forest. \hfill \Halmos \\

We note that the probability bound of Theorem~\ref{theorem:shallow_general_result} may not always be less than or equal to 1; a necessary but not sufficient condition for the bound to be less than or equal to 1 is that $\epsilon$ and $M_0$ are chosen so that $kM_0 > 1$, ensuring that the coefficient of $q$ in $2^{-q (kM_0 - 1)}$ is negative.

To obtain the probability bound of Section~\ref{subsec:simple_trees_depth} with $N = 10000$ and $M = 2000$, we set $M_0 = 20$ and $\epsilon = 0.5$. We then obtain $\bar{d} = \lceil \log_{2-0.5}(2000/20) \rceil = 12$ and $Q(10000, 2000, 40) \leq (2000)^2 \cdot 2^{0.111} \cdot 2^{-0.065 \times 10000 / \log_2 2000} \leq 6.2 \times 10^{-12}$. The forest depth is at most $\bar{d} + M_0 + 1 = 12 + 20 + 1 = 33$, as required. Using Theorem~\ref{theorem:shallow_general_result}, we can actually obtain a tighter bound. To set up the bound, note that $q = \lfloor N / \bar{d} \rfloor = 833$ and $kM_0 - 1 = 1/9$, and so we obtain $Q(10000, 2000, 40) \leq (2^{\bar{d}} - 1) \cdot 2^{ -q(kM_0 - 1)} \leq 5.63 \times 10^{-25}$. In Table~\ref{table:probability_bound_values} below, we report values of both the bound in the proof of Theorem~\ref{thm:asymptotic_size_of_forest} as well as the tighter bound of Theorem~\ref{theorem:shallow_general_result} for a collection of values of $M$ and $N$.

\begin{table}[ht]
	{\SingleSpacedXI
		
		\centering
		\begin{tabular}{lllll}
			\toprule
			$N$ & $M$ & Depth & Failure Prob. Bound & Failure Prob. Bound \\ 
			& & & (Theorem~\ref{thm:asymptotic_size_of_forest}) & (Theorem.~\ref{theorem:shallow_general_result}) \\
			\midrule
			2000 & 100 & 25 & 0.0139 & $2.83 \times 10^{-16}$ \\ 
			2000 & 200 & 27 & 0.328 & $4.58 \times 10^{-10}$ \\ 
			2000 & 500 & 29 & 11.7 & $1.11 \times 10^{-6}$ \\ 
			2000 & 1000 & 31 &  128 & 0.000209 \\ 
			2000 & 2000 & 33 & $1.17 \times 10^{3}$ & 0.0115 \\ 
			2000 & 5000 & 35 & $1.77 \times 10^{4}$ & 0.292 \\ 
			2000 & 10000 & 37 & $1.23 \times 10^{5}$ & 4.32 \\ 
			2000 & 20000 & 39 & $7.88 \times 10^{5}$ & 50.8 \\ \midrule
			5000 & 100 & 25 & $2.04 \times 10^{-11}$ & $2.32 \times 10^{-41}$ \\ 
			5000 & 200 & 27 & $6.87 \times 10^{-9}$ & $8.66 \times 10^{-27}$ \\ 
			5000 & 500 & 29 & $3.31 \times 10^{-6}$ & $3.17 \times 10^{-19}$ \\ 
			5000 & 1000 & 31 & 0.000165 & $1.93 \times 10^{-14}$ \\ 
			5000 & 2000 & 33 & 0.00518 & $4.99 \times 10^{-11}$ \\ 
			5000 & 5000 & 35 & 0.295 & $1.88 \times 10^{-8}$ \\ 
			5000 & 10000 & 37 & 4.69 & $2.4 \times 10^{-6}$ \\ 
			5000 & 20000 & 39 & 61.4 & 0.000142 \\ \midrule
			10000 & 100 & 25 & $3.84 \times 10^{-26}$ & $3.6 \times 10^{-83}$ \\ 
			10000 & 200 & 27 & $1.09 \times 10^{-21}$ & $1.19 \times 10^{-54}$ \\ 
			10000 & 500 & 29 & $4.06 \times 10^{-17}$ & $3.95 \times 10^{-40}$ \\ 
			10000 & 1000 & 31 & $2.52 \times 10^{-14}$ & $3.65 \times 10^{-31}$ \\ 
			10000 & 2000 & 33 & $6.21 \times 10^{-12}$ & $5.63 \times 10^{-25}$ \\ 
			10000 & 5000 & 35 & $3.22 \times 10^{-9}$ & $2.15 \times 10^{-20}$ \\ 
			10000 & 10000 & 37 & $2.04 \times 10^{-7}$ & $8.16 \times 10^{-17}$ \\ 
			10000 & 20000 & 39 & $8.74 \times 10^{-6}$ & $7.16 \times 10^{-14}$ \\ 
			\bottomrule
		\end{tabular}
		
		\caption{Probability bound values and corresponding depths for different values of $N$ and $M$. ``Failure Prob. Bound (Theorem~\ref{thm:asymptotic_size_of_forest})'' corresponds to the loose bound from the proof of Theorem~\ref{thm:asymptotic_size_of_forest} (the bound $M^2 \cdot 2^{0.111} \cdot 2^{-0.065 N / \log_2 M}$). ``Failure Prob. Bound (Theorem~\ref{theorem:shallow_general_result})'' corresponds to the tighter bound of Theorem~\ref{theorem:shallow_general_result} (the bound $(2^{\bar{d}}-1) \cdot 2^{-q (kM_0 - 1)}$, with $M_0 = 20$ and $\epsilon = 0.5$). ``Depth'' is the bound on the depth of the forest that corresponds to the two probability bounds, which in both cases is $M_0 + \bar{d} + 1$. \label{table:probability_bound_values}}
	}
\end{table}

\clearpage

\subsection{Proof of Theorem~\ref{thm:convergence_of_sampling_method}}
Before diving into the main elements of the proof, we fix some additional notation. We use $\Ab_t = (\Ab_{t,S} )_{S \in \Scal}$ to denote the vector obtained by concatenating the vectors $\Ab_{t,S}$ over all the training assortments $S \in \Scal$. Note that each vector $\Ab_{t,S}$ is a $N+1$ dimensional ``one-hot'' vector (i.e., exactly one entry of $\Ab_{t,S}$ is one, and the remaining entries are zero); therefore, the vector $\Ab_t$, being the concatenation of $M$ one-hot vectors, will have $L_1$ norm $\| \Ab_t \|_1 = M$ and $L_2$ norm $\| \Ab_t \|_2 = \sqrt{M}$. We additionally define $\vb = (\vb_S)_{S \in \Scal}$ to be the concatenation of the $\vb_S$ vectors. 

For a collection of trees $F$ and a nonnegative weight vector $\mub = (\mu_t)_{t \in F}$ corresponding to $F$, we define $\psib(F, \mub)$ as
\begin{equation}
\psib(F, \mub) = \sum_{t \in F} \Ab_t \mu_t.
\end{equation}
When $\mub$ sums up to 1, $\mub$ corresponds to a probability distribution over $F$, and $(F, \mub)$ is a bona fide decision forest model. In that case, recall that $\sum_{t \in F} \Ab_{t,S} \mu_t$ is the vector of predicted choice probabilities for assortment $S$ given the decision forest model $(F, \lambdab)$; thus, when $\mub$ is a probability distribution over the forest $F$, then $\psib(F, \mub)$ is the concatenation of all such vectors of predicted choice probabilities, over all of the assortments in $\Scal$. We refer to a tuple $(F, \mub)$ where $\mub$ does not necessarily sum to one as an \emph{extended decision forest model}. Our definition of $\psib$ is intentionally general as we will use it in conjunction with both ordinary (non-extended) and extended decision forest models.

With these additional definitions, we now prove Theorem~\ref{thm:convergence_of_sampling_method}. Our proof relies on two auxiliary results (Lemma~\ref{lemma:risk_is_bounded_by_psib_distance} and Lemma~\ref{lemma:psib_concentrates}), which we will establish after stating the proof of Theorem~\ref{thm:convergence_of_sampling_method}. \\

\proof{Proof of Theorem~\ref{thm:convergence_of_sampling_method}:}
Recall that the randomized tree sampling method (Algorithm~\ref{alg:RTS}) returns a decision forest model $(\hat{F}, \hat{\lambdab})$, where $\hat{F} = \{t_1, \dots, t_K\}$ is a random sample of $K$ trees drawn i.i.d. from $F$ according to the distribution $\xib$, and $\hat{\lambdab}$ is obtained by solving the problem $\EstLO(\Scal, \hat{F})$. Let $\lambdab^*$ be a probability distribution over $\Lambda(C, \xib)$ that minimizes the empirical risk, that is,
\begin{equation}
\lambdab^* \in \arg \min_{\lambdab \in \Lambda(C, \xib)} R(F, \lambdab).
\end{equation}

By Lemma~\ref{lemma:psib_concentrates}, with probability $1 - \delta$ over the collection of trees $t_1,\dots, t_K$, there exists a forest model $(\hat{F}, \lambdab')$ such that 
\begin{equation}
\| \psib(\hat{F}, \lambdab') - \psib(F, \lambdab^*) \|_1 \leq \frac{MC}{\sqrt{K}} \cdot \left( \sqrt{N+1} + 3 \sqrt{\log(4 / \delta)} \right).
\end{equation}
In words, Lemma~\ref{lemma:psib_concentrates} allows us to assert that with high probability there exists a distribution $\lambdab'$ over $\hat{F}$ such that the predicted choice probabilities under $(\hat{F}, \lambdab')$ are close to those under $(F, \lambdab^*)$. 

Thus, with probability at least $1 - \delta$, we have the following bound:
\begin{align*}
\Rb(\hat{F}, \hat{\lambdab}) - \min_{\lambdab \in \Lambda(C, \xib)} \Rb(F, \lambdab) & = \Rb(\hat{F}, \hat{\lambdab}) - \Rb(F, \lambdab^*) \\
& \leq \Rb(\hat{F}, \lambdab') - \Rb(F, \lambdab^*) \\ 
& \leq \frac{1}{M} \cdot \| \psib(\hat{F}, \lambdab') - \psib(F, \lambdab^*) \|_1 \\
& \leq \frac{1}{M} \cdot \frac{MC}{\sqrt{K}} \cdot \left( \sqrt{N+1} + 3 \sqrt{\log(4 / \delta)} \right) \\
& = \frac{C}{\sqrt{K}} \cdot \left( \sqrt{N+1} + 3 \sqrt{\log(4 / \delta)} \right), %
\end{align*}
where the first equality follows by the definition of $\lambdab^*$; the first inequality follows since $\hat{\lambdab}$ minimizes $\Rb( \hat{F}, \lambdab)$ over all probability distributions $\lambdab$; the second inequality follows by Lemma~\ref{lemma:risk_is_bounded_by_psib_distance}; and the final inequality by Lemma~\ref{lemma:psib_concentrates}. Re-arranging the inequality to place $\min_{\lambdab \in \Lambda(C, \xib)} \Rb(F, \lambdab)$ to the right-hand side, we obtain the desired result. \hfill \Halmos \\
\endproof

We now establish the auxiliary results used in the proof of Theorem~\ref{thm:convergence_of_sampling_method}. Our first auxiliary result, Lemma~\ref{lemma:risk_is_bounded_by_psib_distance}, states that the difference in training error/empirical risk between two decision forests $(F, \lambdab)$ and $(F', \lambdab')$ can be bounded simply by the $L_1$ distance between their predicted choice probabilities. %
\begin{lemma}
	\label{lemma:risk_is_bounded_by_psib_distance}
	For any two forest models $(F, \lambdab)$ and $(F', \lambdab')$,
	\begin{equation}
	\Rb(F, \lambdab) - \Rb(F', \lambdab') \leq \frac{ \| \psib(F, \lambdab) - \psib(F', \lambdab') \|_1 } {M}.
	\end{equation}
\end{lemma}

\proof{Proof of Lemma~\ref{lemma:risk_is_bounded_by_psib_distance}:}
By the triangle inequality, we have
\begin{equation*}
\|\psib(F, \lambdab) - \vb\|_1 \leq \| \psib(F, \lambdab) - \psib(F', \lambdab') \|_1 + \| \psib(F', \lambdab') - \vb \|_1,
\end{equation*}
which we can re-arrange to obtain
\begin{equation*}
\Rb(F, \lambdab) - \Rb(F', \lambdab') = \frac{ \| \psib(F, \lambdab) - \vb \|_1 }{M} - \frac{ \| \psib(F', \lambdab') - \vb \|_1 }{M} \leq \frac{ \| \psib(F, \lambdab) - \psib(F', \lambdab') \|_1 }{M},
\end{equation*}
as required. \hfill \Halmos \\
\endproof

We next turn our attention to Lemma~\ref{lemma:psib_concentrates}. Before we can establish Lemma~\ref{lemma:psib_concentrates}, it is helpful to establish the following general-purpose lemma. For a collection of i.i.d. random vectors, Lemma~\ref{lemma:bounded_L1_deviation_abstract} provides a high probability bound on the $L_1$ norm of the distance between the average vector and the expectation of the average vector. 

\begin{lemma}
	Let $\zb_1, \dots, \zb_K$ be i.i.d. random vectors of size $(N+1)M$ such that $\zb_k \geq \zerob$, $\| \zb_k \|_1 \leq A$ and $\| \zb_k \|_2 \leq B$ for $k = 1, \dots, K$, for some positive constants $A$ and $B$. Let $\bar{\zb} = (1/K) \sum_{k=1}^K \zb_k$ denote their average. Then, for any $\delta > 0$, with probability at least $1- \delta$ over the draw of $\zb_1,\dots, \zb_K$, we have that
	\begin{equation}
	\| \bar{\zb} - \Exp[ \bar{\zb} ] \|_1 \leq \sqrt{ \frac{(N+1)M B^2}{K} } + \sqrt{ \frac{2 A^2}{K} \log\left(\frac{1}{\delta} \right) }\label{eq:lemma_bounded_L1_deviation}
	\end{equation}
	
	\label{lemma:bounded_L1_deviation_abstract}

\end{lemma}

\proof{Proof of Lemma~\ref{lemma:bounded_L1_deviation_abstract}:}
First, let us define the set $\Ycal$ from which $\zb_1, \dots, \zb_K$ are drawn as
\begin{equation*}
\Ycal = \left\{ \yb \in \Rbb^{(N+1)M} \mid \yb \geq \zerob, \| \yb \|_1 \leq A, \| \yb \|_2 \leq B \right\}.
\end{equation*}
Let us also define the scalar function $f: \Ycal^K \to \Rbb$ as 
\begin{equation*}
f(\yb_1, \dots, \yb_K) = \| \frac{1}{K} \sum_{k=1}^K \yb_k - \Exp[ \bar{\zb} ] \|_1.
\end{equation*}
Observe that the random variable $f(\zb_1,\dots, \zb_K)$ is equivalent to the random variable on the left-hand side of inequality~\eqref{eq:lemma_bounded_L1_deviation}; our goal will be to show that $f(\zb_1,\dots, \zb_K)$ satisfies this bound. We will show this by combining a bound on how much $f(\zb_1,\dots, \zb_K)$ deviates from its expected value (which we will obtain using McDiarmid's inequality) and a bound on the expected value of $f(\zb_1,\dots, \zb_K)$. 

To eventually use McDiarmid's inequality, we first show that $f$ possesses the bounded differences property. For any $\yb_1, \dots, \yb_K \in \Ycal$, let $\tilde{\yb}_1, \dots, \tilde{\yb}_K$ be a collection of vectors in $\Ycal$ such that $\yb_k = \tilde{\yb}_k$ for all $k \neq m$, where the index $m \in \{1,\dots, K\}$ is arbitrary. We then have:
\begin{align*}
| f(\yb_1,\dots, \yb_K) - f(\tilde{\yb}_1, \dots, \tilde{\yb}_K) | & = \left| \| \frac{1}{K} \sum_{k=1}^K \yb_k - \Exp[ \bar{\zb} ] \|_1 - \| \frac{1}{K} \sum_{k=1}^K \tilde{\yb}_k - \Exp[ \bar{\zb} ] \|_1 \right| \\
& \leq \| \frac{1}{K} \sum_{k=1}^K \yb_k - \frac{1}{K} \sum_{k=1}^K \tilde{\yb}_k \|_1 \\
& = \frac{1}{K} \| \yb_m - \tilde{\yb}_m \|_1 \\
& \leq \frac{2A}{K},
\end{align*}
where both inequalities follow by an application of the triangle inequality. 

We next bound $\Exp[ f(\zb_1,\dots, \zb_K) ]$. To do so, we first derive an auxiliary bound on $\Exp[ \| \bar{\zb} - \Exp[\bar{\zb}] \|^2_2 ]$:
\begin{align}
\Exp[ \| \bar{\zb} - \Exp[\bar{\zb}] \|^2_2 ] & = \frac{\Exp[ \| \bar{\zb} \|^2_2 ] -  \|  \Exp[\bar{\zb}] \|^2_2 }{K} \nonumber \\ 
& \leq \frac{ \Exp[ \| \bar{\zb} \|^2_2 ] }{K} \nonumber \\
& \leq \frac{ B^2 }{K}, \label{eq:bound_barzb_variance}
\end{align}
where the first inequality follows since $\| \Exp[ \bar{\zb} ] \|^2_2 \geq 0$ and the second follows because $\Ycal$ is convex, so that $\| \bar{\zb} \|_2$ is bounded by $B$ almost surely. 

Using this bound, we now derive a bound on $\Exp[ f(\zb_1,\dots, \zb_K) ]$:
\begin{align}
\Exp[ f(\zb_1,\dots, \zb_K) ] & = \Exp[ \| \bar{\zb} - \Exp[ \bar{\zb} ] \|_1 ] \nonumber \\
& \leq \sqrt{(N+1)M} \Exp[ \| \bar{\zb} - \Exp[ \bar{\zb} ] \|_2 ] \nonumber \\
& \leq \sqrt{(N+1)M} \sqrt{ \Exp[ \| \bar{\zb} - \Exp[ \bar{\zb} ] \|^2_2 ]  } \nonumber \\
& \leq \sqrt{ \frac{ (N+1)M B^2}{K} }, \label{eq:bound_barzbdeviation_expected_value}
\end{align}
where the first inequality follows by the basic properties of $L_1$ and $L_2$ norms, the second inequality follows by using Jensen's inequality and the concavity of the function $h(x) = \sqrt{x}$, and the final inequality by using inequality~\eqref{eq:bound_barzb_variance}. 

We now have all the pieces necessary to establish the bound~\eqref{eq:lemma_bounded_L1_deviation}. For any $\epsilon > 0$, we have:
\begin{align*}
\Pbb \left( f(\zb_1, \dots, \zb_K) - \sqrt{ \frac{(N+1)MB^2}{K} } \geq \epsilon \right) & \leq \Pbb \left( f(\zb_1,\dots, \zb_K) - \Exp[ f(\zb_1,\dots, \zb_K)] \geq \epsilon \right) \\
& \leq \exp \left( \frac{ -2 \epsilon^2 }{ K \cdot \frac{4A^2}{K^2} }  \right) \\
& = \exp \left( - \frac{ K \epsilon^2}{2 A^2 } \right),
\end{align*}
where the first inequality follows by our bound on the expected value (inequality~\eqref{eq:bound_barzbdeviation_expected_value}) and the second inequality follows by an application of McDiarmid's inequality. Letting $\epsilon = \sqrt{ (2A^2 / K) \log(1 / \delta) }$, we obtain that 
\begin{equation*}
f(\zb_1,\dots, \zb_K) \leq \sqrt{ \frac{(N+1)MB^2}{K} } + \sqrt{ \frac{2A^2}{K} \log \left( \frac{1}{\delta} \right) }
\end{equation*}
with probability at least $1 - \delta$, which completes the proof. \hfill \Halmos \\
\endproof 

Equipped with Lemma~\ref{lemma:bounded_L1_deviation_abstract}, we can now turn to proving Lemma~\ref{lemma:psib_concentrates}, which is at the heart of the proof of Theorem~\ref{thm:convergence_of_sampling_method}. Lemma~\ref{lemma:psib_concentrates} states that for any decision forest model $(F,\lambdab)$ where $\lambdab \in \Lambda(C, \xib)$, we can find a new distribution $\lambdab'$ over the forest $F'$, which is an i.i.d. sample of $K$ trees from $\xib$, such that the $L_1$ distance between the predicted choice probabilities of $(F,\lambdab)$ and $(F', \lambdab')$ is bounded with high probability. The distribution $\lambdab'$ is constructed by first constructing an appropriate extended forest model $(F', \mub')$, and then normalizing $\mub'$ to sum to 1. The proof then involves two steps: (1) showing that $(F,\lambdab)$ and $(F', \mub')$ are close in their choice probabilities using Lemma~\ref{lemma:bounded_L1_deviation_abstract}; and (2) showing that $(F', \mub')$ and $(F', \lambdab')$ are also close in their choice probabilities (using the fact that the normalization constant of $\lambdab'$ concentrates to 1). We note that our proof of step (1) resembles a technique used in the machine learning literature on building classifiers as weighted sums of random feature functions \citep{rahimi2009weighted}. Specifically, we use a similar procedure as in \cite{rahimi2009weighted} to construct our extended forest model $(F',\mub')$. However, unlike the weighted sum models in \cite{rahimi2009weighted}, a decision forest model requires that the weight vector $\lambdab$ sum up to 1 as it corresponds to a probability distribution; for this reason, we must also establish step (2).

\begin{lemma}
	
	Consider a decision forest model $(F, \lambdab)$ where $\lambdab \in \Lambda(C, \xib)$. Suppose that $\xi_t > 0$ for all $t \in F$, and that $t_1, \dots, t_K$ are drawn i.i.d. from $F$ according to the distribution $\xib$. For any $\delta > 0$, with probability at least $1 - \delta$ over the draw of $t_1,\dots, t_K$, there exists a decision forest model $(F', \lambdab')$ such that $F' = \{t_1,\dots, t_K\}$ and
	\begin{equation}
	\| \psib(F', \lambdab') - \psib(F, \lambdab) \|_1 \leq \frac{MC}{\sqrt{K}} \left( \sqrt{N+1} + 3 \sqrt{\log \left( \frac{4}{\delta} \right) } \right).
	\end{equation}
	
	\label{lemma:psib_concentrates}
\end{lemma}

\proof{Proof of Lemma~\ref{lemma:psib_concentrates}:}
Let $F' = \{t_1,\dots, t_K\}$. We first consider the extended forest model $(F', \mub')$, where $\mub'$ is defined as 
\begin{equation*}
\mu'_{t_k} = \frac{1}{K} \left( \frac{\lambda_{t_k}}{\xi_{t_k}} \right).
\end{equation*}
Note that $\mub'$ is not necessarily a probability distribution because it need not add up to 1. We thus define the distribution $\lambdab'$ over $F'$ by normalizing $\mub'$:
\begin{equation}
\lambdab' = \left( \frac{1}{\sum_{k=1}^K \mu'_{t_k} } \right) \mub' \label{eq:definition_lambdab_prime}
\end{equation}
We claim that $(F', \lambdab')$ satisfies the statement of the theorem. To see how, observe that we can bound the quantity $\| \psib(F', \lambdab') - \psib(F, \lambdab) \|_1$ as
\begin{equation}
\| \psib(F', \lambdab') - \psib(F, \lambdab) \|_1 \leq \underbrace{\| \psib(F, \lambdab) - \psib(F', \mub')\|_1}_{\text{Term (a)}} + \underbrace{\| \psib(F', \mub') - \psib(F', \lambdab') \|_1}_{\text{Term (b)}}, \label{eq:bound_F_Fprime_diff}
\end{equation}
which follows by applying the triangle inequality. We now show that each of the two terms on the right hand side can be bounded with high probability. \\

\noindent \textbf{Bounding term (a):}  To bound $\| \psib(F, \lambdab) - \psib(F', \mub')\|_1$, let us define the random vector $\zb_k$ as 
\begin{equation}
\zb_k = \left( \frac{\lambda_{t_k}}{\xi_{t_k}} \right) \Ab_{t_k},
\end{equation}
where we recall that $\Ab_t$ is the concatenation of $M$ one-hot vectors of size $N+1$ corresponding to the choices that tree $t$ makes on the $M$ assortments in the data. Let us also define the random vector $\bar{\zb} = (1/K) \sum_{k=1}^K \zb_k$ as the average of the $K$ random vectors.

The vectors $\zb_1,\dots, \zb_K$ have a couple of desirable properties. First, observe that
\begin{align*}
\psib(F', \mub') &= \sum_{k=1}^K \mu'_{t_k} \Ab_{t_k}, \\
& = \sum_{k=1}^K \left(\frac{1}{K} \right) \left( \frac{\lambda_{t_k} }{\xi_{t_k}} \right) \Ab_{t_k} \\
& = \frac{1}{K} \sum_{k=1}^K \zb_k,
\end{align*}
i.e., the (random) vector of choice probabilities of the extended forest model $(F', \mub')$ is equal to the average of the random vectors $\zb_1,\dots, \zb_K$. 

Second, observe that for any $k \in \{1,\dots,K\}$, we have
\begin{align*}
\Exp[\bar{\zb}] & = \Exp[ \zb_k ] \\
& = \sum_{t \in F} \xi_t \cdot \frac{\lambda_t}{\xi_t} \Ab_t \\
& = \sum_{t \in F} \lambda_t \Ab_t \\
& = \psib(F, \lambdab),
\end{align*}
i.e., the expected value of $\bar{\zb}$ is exactly equal to the vector of choice probabilities of the decision forest model $(F, \lambdab)$.

We can thus re-write the term $\| \psib(F, \lambdab) - \psib(F', \mub') \|_1$ as 
\begin{equation}
\| \psib(F, \lambdab) - \psib(F', \mub') \|_1 = \| \bar{\zb} - \Exp[ \bar{\zb} ] \|_1,
\end{equation}
which is exactly in the form of the bound in Lemma~\ref{lemma:bounded_L1_deviation_abstract}. In order to apply the bound, we only need to obtain bounds on the $L_1$ and $L_2$ norms of the random vectors $\zb_1,\dots, \zb_K$. Since $\zb_k = (\lambda_{t_k} / \xi_{t_k} ) \Ab_{t_k}$, we can use the fact that $\lambda_t \leq C \xi_t$ for every $t \in F$ (recall that $\lambdab \in \Lambda(C, \xib)$) and the fact that $\Ab_t$ consists of $M$ one-hot vectors concatenated together, to obtain the following bounds:
\begin{align*}
\| \zb_k \|_1 = \frac{\lambda_{t_k}}{\xi_{t_k}} \cdot \| \Ab_{t_k} \|_1  \leq C \| \Ab_{t_k} \|_1 = CM,
\end{align*}
\begin{align*}
\| \zb_k \|_2 = \frac{\lambda_{t_k}}{\xi_{t_k}} \cdot \| \Ab_{t_k} \|_2 \leq C \| \Ab_{t_k} \|_2 = C \sqrt{M}.
\end{align*}
With these bounds in hand, we can invoke Lemma~\ref{lemma:bounded_L1_deviation_abstract} with $\delta/2$ to obtain that 
\begin{align}
\| \psib(F, \lambdab) - \psib(F', \mub') \|_1 & = \| \bar{\zb} - \Exp[\bar{\zb}] \|_1 \nonumber \\
& \leq \sqrt{ \frac{(N+1)M \cdot C^2 M}{K} } + \sqrt{ \frac{2 C^2 M^2}{K} \log \left( \frac{1}{\delta/2} \right) } \nonumber \\
& = \frac{CM}{ \sqrt{K} } \left( \sqrt{N+1} + \sqrt{ 2 \log \left( \frac{2}{\delta} \right) } \right), \label{eq:term_a_final_bound}
\end{align}
with probability at least $1 - \delta/2$. \\

\noindent \textbf{Bounding term (b):} To bound $\| \psib(F', \mub') - \psib(F', \lambdab') \|_1$, let us define the random variable $s$ as $s = \sum_{k=1}^K \mu'_{t_k}$, which is simply the normalization constant used to define $\lambdab'$. Let us also define the random variables $\beta_1,\dots, \beta_K$ as $\beta_k = \lambda_{t_k} / \xi_{t_k}$. Observe that $s$ can then be written as 
\begin{equation}
s = \frac{1}{K} \sum_{k=1}^K \beta_k,
\end{equation} 
i.e., $s$ is the average of $K$ i.i.d. random variables, $\beta_1,\dots, \beta_K$. Moreover, each random variable $\beta_k$ is bounded between $0$ (since $\lambda_t$ is nonnegative and $\xi_t$ is positive for every $t \in F$) and $C$ (since $\lambda_t \leq C \xi_t$, by definition of $\Lambda(C, \xib)$). Lastly, observe that for each $k$,
\begin{align*}
\Exp[ \beta_k] & = \Exp[ \lambda_{t_k} / \xi_{t_k} ] \\
& = \sum_{t \in F} \xi_t \cdot \frac{\lambda_t}{\xi_t} \\
& = \sum_{t \in F} \lambda_t \\
& = 1,
\end{align*}
i.e., the expected value of each $\beta_k$ is 1, and thus, $\Exp[s]$ will also be 1. We can therefore apply Hoeffding's inequality to bound the deviation of $s$ from 1. For any $\epsilon > 0$, we have
\begin{align*}
\Pbb \left( |s - 1| \geq \epsilon \right) \leq 2 \exp \left( - \frac{2K \epsilon^2}{C^2} \right).
\end{align*}
Set $\epsilon = C \sqrt{ \log(4 / \delta) / 2K}$. We then have, with probability at least $1 - \delta/2$, that
\begin{equation*}
|s - 1| \leq C \sqrt{\frac{1}{2K} \log \left( \frac{4}{\delta} \right) }.
\end{equation*}
We can now use this to bound term (b); we have
\begin{align}
\| \psib(F', \lambdab') - \psib(F', \mub') \|_1 & = \| \sum_{k=1}^K \lambda'_{t_k} \Ab_{t_k} - \sum_{k=1}^K \mu'_{t_k} \Ab_{t_k} \|_1 \nonumber \\
& = \| \sum_{k=1}^K \lambda'_{t_k} \Ab_{t_k} - s \sum_{k=1}^K \lambda'_{t_k} \Ab_{t_k} \|_1 \nonumber \\
& = |s - 1| \cdot \| \sum_{k=1}^K \lambda'_{t_k} \Ab_{t_k} \|_1 \nonumber \\
& = |s - 1| \cdot M  \nonumber \\
& \leq M \cdot C \sqrt{ \frac{1}{2K} \log \left( \frac{4}{\delta} \right) }, \label{eq:term_b_final_bound}
\end{align}
with probability at least $1 - \delta/2$. (In the above, note that the second step follows by the definition of $\lambdab'$ in \eqref{eq:definition_lambdab_prime}.) \\

\noindent \textbf{Bounding $\| \psib(F, \lambdab) - \psib(F', \lambdab')\|_1$}: We now complete the proof. Using the bounds~\eqref{eq:term_a_final_bound} and \eqref{eq:term_b_final_bound} together with the inequality~\eqref{eq:bound_F_Fprime_diff} and the union bound, we get:
\begin{align*}
\| \psib(F, \lambdab) - \psib(F', \lambdab')\|_1 & \leq \| \psib(F, \lambdab) - \psib(F', \mub')\|_1 + \| \psib(F', \lambdab') - \psib(F', \mub') \|_1 \\
& \leq \frac{CM}{\sqrt{K}} \left( \sqrt{N+1} + \sqrt{2 \log\left( \frac{2}{\delta} \right) } \right) + \frac{CM}{\sqrt{K}}  \sqrt{ \frac{1}{2} \log \left( \frac{4}{\delta} \right) }  \\
& \leq \frac{CM}{\sqrt{K}} \left( \sqrt{N+1} + 3 \sqrt{ \log \left(\frac{4}{\delta} \right) } \right),
\end{align*}
with probability at least $1 - \delta$, as required. \hfill \Halmos
\endproof

\clearpage

\newpage

\section{Exact Column Generation}
\label{sec:appendix_CG}

\subsection{An exact formulation of the column generation subproblem}
\label{subsec:appendix_CG_integer_program}

We follow the notation in Section~\ref{sec:model_estimation_methods}. The subproblem for the column generation approach is to find a decision tree $t$ and the corresponding 0-1 vectors $\Ab_{t,S}$ for $S \in \Scal$ such that the reduced cost of the corresponding $\lambda_t$ variable, given by $- \sum_{S \in \Scal} \alphab_S^T \Ab_{t,S} - \nu$, is minimized. To formulate the subproblem, we introduce a binary decision variable $y^\ell_o$ for each leaf node $\ell \in \leaves(t)$ and option $o \in \Ncal^+$ that is 1 if the purchase decision of leaf $\ell$ is option $o$, and 0 otherwise. Similarly, for each split node $s \in \splits(t)$ and product $p \in \Ncal$, we define the binary decision variable $y^s_p$ to be 1 if product $p$ participates in split node $s$, and 0 otherwise. For each $\ell \in \leaves(t)$, we define the binary decision variable $w^\ell_S$ which is 1 if assortment $S$ is mapped to leaf node $\ell$ under the current purchase decision tree.  For each leaf $\ell \in \leaves(t)$, option $o \in \Ncal^+$ and assortment $S$, we define the binary decision variable $u^\ell_{o,S}$ to be 1 if assortment $S$ is mapped to leaf node $\ell$ and option $o$ is the resulting purchase decision. Finally, we define the binary decision variable $A_{o,S}$ to indicate whether the tree chooses option $o$ when given assortment $S$, for each historical assortment $S \in \Scal$; the vector of these decision variables for a given $S$ is denoted by $\Ab_S$. With these definitions, we can formulate the subproblem as the following mixed-integer optimization problem:
\begin{subequations}
	\label{prob:CG_subproblem}
	\begin{alignat}{2}
	& {\underset{ \Ab,\ub, \wb,\yb   }{ \text{minimize} } } & \quad  &  - \sum_{S \in \Scal} \alphab_S^T \Ab_{S} - \nu \\
	& \text{subject to} & & \sum_{p \in \Ncal} y^s_p = 1, \quad \forall\ s \in \splits(t),\label{prob:CG_subproblem_oneproductpersplit}\\
	& & & \sum_{o \in \Ncal^+} y^\ell_o = 1, \quad \forall\ \ell \in \leaves(t),\label{prob:CG_subproblem_oneoptionperleaf}\\
	& & & \sum_{\ell \in \leaves(t)} w^\ell_S = 1, \quad \forall \ m \in \{1,\dots,M\}, \label{prob:CG_subproblem_eachsettooneleaf}\\
	& & & \sum_{\ell \in \mathbf{LL}(s)} w^\ell_S \leq  \sum_{p \in S} y^{s}_p, \quad \forall\ s \in \splits(t),\ S \in \Scal, \label{prob:CG_subproblem_leftsplit}\\
	& & & \sum_{\ell \in \mathbf{RL}(s)} w^\ell_S \leq 1 - \sum_{p \in S} y^{s}_p, \quad \forall\ s \in \splits(t),\ S \in \Scal,\label{prob:CG_subproblem_rightsplit}\\
	& & & w^\ell_S = \sum_{o \in \Ncal^+} u^\ell_{o,S}, \quad \forall \ \ell \in \leaves(t),\ S \in \Scal, \label{prob:CG_subproblem_uom} \\
	& & & u^\ell_{o,S} \leq y^\ell_o ,  \quad \forall\ \ell \in \leaves(t),\ o \in \Ncal^+,\ S \in \Scal,\label{prob:CG_subproblem_uyforcing}\\
	& & & y^\ell_o \leq \sum_{s \in \mathbf{LS}(\ell)} y^s_o, \quad \forall\ \ell \in \leaves(t),\ o \in \Ncal, \label{prob:CG_subproblem_requirement2} \\ 
	& & & A_{o,S} = \sum_{\ell \in \leaves(t)} u^\ell_{o,S}, \quad \forall\ o  \in \Ncal^+,\ S \in \Scal, \label{prob:CG_subproblem_adefinition}\\
	& & & w^\ell_S \in \{0,1\}, \quad \forall\ \ell \in \leaves(t),\ S \in \Scal, \label{prob:CG_subproblem_w_binary}\\
	& & & A_{o,S} \in \{ 0,1 \}, \quad \forall\ o \in \Ncal^+,\ S \in \Scal,\label{prob:CG_subproblem_a_binary}\\
	& & & y^\ell_o \in \{0,1  \}, \quad \forall\ \ell \in \leaves(t),\ o \in \Ncal^+, \label{prob:CG_subproblem_y_ell_binary}\\
	& & & y^s_p \in \{0,1  \},  \quad \forall\ s \in \splits(t),\ p \in \Ncal, \label{prob:CG_subproblem_y_s_binary}\\
	& & & u^\ell_{o,S} \in \{0,1\}, \quad \forall\ \ell \in \leaves(t),\ o \in \Ncal^+,\ S \in \Scal.\label{prob:CG_subproblem_u_binary}
	\end{alignat}%
\end{subequations}
In order of appearance, the constraints have the following meaning. Constraint~\eqref{prob:CG_subproblem_oneproductpersplit} requires that exactly one product is chosen for each split in the tree. Constraint~\eqref{prob:CG_subproblem_oneoptionperleaf} similarly requires exactly one option to be selected to serve as the purchase decision for each leaf. Constraint~\eqref{prob:CG_subproblem_eachsettooneleaf} ensures that each assortment is mapped to exactly one leaf. Constraints~\eqref{prob:CG_subproblem_leftsplit} and \eqref{prob:CG_subproblem_rightsplit} model how the tree maps each of the assortments to a leaf. To understand these constraints, observe that the expression $\sum_{p \in S} y^s_p$ is 1 if any of the products in $S$ is chosen for split $s$, in other words, it is 1 if the purchase decision process proceeds to the left child of split $s$, and 0 if it proceeds to the right child. If the expression evaluates to 1, the decision process proceed to the left, and constraint~\eqref{prob:CG_subproblem_rightsplit} forces all the $w^\ell_S$ variables of all leaves $\ell$ that are to the right of split $s$ to zero. Similarly, if the expression evaluates to 0, the decision process proceeds to the right, and constraint~\eqref{prob:CG_subproblem_leftsplit} forcess all $w^\ell_S$ variables for leaves to the left of split $s$ to zero. Constraints~\eqref{prob:CG_subproblem_uom} and \eqref{prob:CG_subproblem_uyforcing} ensure that $u^\ell_{o,S}$ is consistent with $w^\ell_S$ and $y^\ell_o$. Constraint~\eqref{prob:CG_subproblem_requirement2} ensures that the tree conforms to Requirement 2 in Section~\ref{subsec:model_decision_trees}, i.e., a product $o$ may only be set as the purchase decision of a leaf $\ell$ if it participates in at least one split $s$ for which $\ell$ is to the left of. Constraint~\eqref{prob:CG_subproblem_adefinition} ensures that each value $A_{o,S}$ of the tree is properly defined given $u^\ell_{o,S}$. Lastly, constraints~\eqref{prob:CG_subproblem_w_binary} to \eqref{prob:CG_subproblem_u_binary} ensure that all of the decision variables are binary. In case that the resulting tree violates Requirement 3 in Section~\ref{subsec:model_decision_trees}, one applies the procedure in Lemma \ref{lemma:equivalence_no_repeated_products} (Algorithm~\ref{alg:check_requirement_3}) to modify the tree.

\subsection{Numerical comparison}
\label{subsec:appendix_CG_runtime}

We now provide a simple numerical comparison of the heuristic column generation approach (Algorithm~\ref{alg:CG} using Algorithm~\ref{alg:TDI} to solve the subproblem) and the randomized tree sampling approach (Algorithm~\ref{alg:RTS}) with the exact column generation approach (Algorithm~\ref{alg:CG} where the subproblem is solved via the MIO formulation~\eqref{prob:CG_subproblem}). In this experiment, we consider $N \in \{4, 6, 8\}$. For each value of $N$, we consider an MNL model where the product utilities $u_i$ are defined as $u_i = 0.2i$ for $i \in \{1,\dots,N\}$. For $N = 4$, we randomly generate $M = 10$ distinct assortments; for $N = 6$, we randomly generate $M = 10, 20, 50$ distinct assortments; and for $N = 8$, we randomly generate $M = 10, 20, 50, 100, 200$ distinct assortments. For each assortment $S$, we compute the exact choice probability vector $\vb_S$ and consider the $L_1$ estimation problem from Section~\ref{subsec:estimation_by_column_generation}. We consider values of the forest depth $d$ in the set $\{3,4,5\}$. 

For each combination of $N$, $M$ and $d$, we execute both the heuristic column generation approach and the exact column generation approach, with the average $L_1$ error as the objective. For the exact column generation approach, we impose a time limit of 6 hours on the whole procedure and a time limit of 30 minutes on each solve of the subproblem; the latter time limit was necessary as in some larger cases, solving just a single instance of the MIO problem~\eqref{prob:CG_subproblem} can exhaust the 6 hour time limit on the overall procedure. For the heuristic column generation approach, we use $d$ as the depth limit for the top-down induction procedure (Algorithm~\ref{alg:TDI}). For the randomized tree sampling approach, we sample $K = 5000$ trees randomly from the uniform distribution over all balanced trees of depth $d$. For all three approaches, we do not use any warm-starting, and initialize each with an empty collection of trees. 

Table~\ref{table:synthetic_ECG} compares the average $L_1$ training error and the overall runtime for the heuristic column generation, the exact column generation and randomized tree sampling approaches; additionally, the table also reports the number of iterations for the two column generation approaches. In terms of runtime, we can see that while the exact column generation approach is manageable for smaller instances, it quickly becomes unmanageable when $N$, $M$ or $d$ become large; for example, even with $N = 6$, $M = 50$ and $d = 5$, the exact approach does not terminate within the 6 hour time limit. In contrast, the heuristic column generation approach requires no more than 3 minutes to run even in the largest case, while the randomized tree sampling approach requires no more than 6 seconds. 

\begin{table}
	{\SingleSpacedXI
		\begin{center}
			\begin{tabular}{rrrrrrrrrrr} \toprule
				$N$ & $M$ & $d$ & \multicolumn{3}{c}{Training error ($10^{-2}$)} & \multicolumn{3}{c}{Runtime (s)} & \multicolumn{2}{c}{Iterations} \\ 
				& & & ECG & HCG & RTS & ECG & HCG & RTS & ECG & HCG \\ \midrule
				4 & 10 & 3 & 5.289 & 15.577 & 5.289 & 8.7 & 2.2 & 0.2 & 16 & 11 \\ 
				4 & 10 & 4 & 0.000 & 0.000 & 0.000 & 12.7 & 2.2 & 0.3 & 21 & 22 \\ 
				4 & 10 & 5 & 0.000 & 0.000 & 0.000 & 12.2 & 2.2 & 0.6 & 21 & 22 \\ \midrule
				6 & 10 & 3 & 2.109 & 6.489 & 2.109 & 21.5 & 2.2 & 0.3 & 37 & 24 \\ 
				6 & 10 & 4 & 0.000 & 0.266 & 0.000 & 24.8 & 2.2 & 0.4 & 32 & 29 \\ 
				6 & 10 & 5 & 0.000 & 0.000 & 0.000 & 30.4 & 2.3 & 0.7 & 33 & 33 \\[0.5em]
				6 & 20 & 3 & 13.814 & 14.111 & 13.814 & 38.0 & 2.2 & 0.4 & 31 & 27 \\ 
				6 & 20 & 4 & 0.000 & 0.000 & 0.000 & 224.2 & 2.5 & 0.5 & 63 & 84 \\ 
				6 & 20 & 5 & 0.000 & 0.000 & 0.000 & 529.7 & 2.6 & 0.8 & 63 & 63 \\[0.5em] 
				6 & 50 & 3 & 20.395 & 20.690 & 20.395 & 125.5 & 2.3 & 0.5 & 37 & 39 \\ 
				6 & 50 & 4 & 4.717 & 4.826 & 4.717 & 12816.8 & 2.8 & 0.9 & 122 & 131 \\ 
				6 & 50 & 5 & 1.793 & 0.355 & 1.029 & 21948.6 & 4.4 & 1.2 & 41 & 209 \\ \midrule 
				8 & 10 & 3 & 5.018 & 11.574 & 5.634 & 49.2 & 2.3 & 0.2 & 56 & 35 \\ 
				8 & 10 & 4 & 0.000 & 0.000 & 0.000 & 69.2 & 2.4 & 1.2 & 46 & 50 \\ 
				8 & 10 & 5 & 0.000 & 0.000 & 0.000 & 29.7 & 2.5 & 0.8 & 46 & 45 \\[0.5em] 
				8 & 20 & 3 & 11.089 & 13.521 & 11.089 & 101.1 & 2.3 & 0.4 & 53 & 46 \\ 
				8 & 20 & 4 & 0.000 & 0.000 & 0.340 & 1063.1 & 2.8 & 1.4 & 100 & 109 \\ 
				8 & 20 & 5 & 0.000 & 0.000 & 0.000 & 4476.3 & 3.1 & 2.2 & 97 & 95 \\[0.5em] 
				8 & 50 & 3 & 21.680 & 22.826 & 21.680 & 367.1 & 2.3 & 0.5 & 64 & 41 \\ 
				8 & 50 & 4 & 5.211 & 5.659 & 8.105 & 21648.4 & 4.2 & 1.9 & 163 & 252 \\ 
				8 & 50 & 5 & 5.397 & 0.000 & 8.472 & 22302.4 & 7.5 & 1.3 & 21 & 249 \\[0.5em] 
				8 & 100 & 3 & 29.439 & 30.161 & 29.439 & 973.5 & 2.4 & 1.0 & 58 & 44 \\ 
				8 & 100 & 4 & 9.799 & 10.538 & 12.163 & 21871.8 & 3.8 & 4.7 & 124 & 164 \\ 
				8 & 100 & 5 & 12.115 & 1.647 & 14.962 & 22250.8 & 109.9 & 3.1 & 16 & 1031 \\[0.5em] 
				8 & 200 & 3 & 30.489 & 30.801 & 30.489 & 3026.6 & 2.5 & 1.9 & 59 & 41 \\ 
				8 & 200 & 4 & 13.365 & 11.180 & 12.620 & 22058.3 & 5.8 & 6.1 & 39 & 204 \\ 
				8 & 200 & 5 & 20.451 & 3.512 & 14.646 & 21747.6 & 160.2 & 5.1 & 14 & 971 \\ \bottomrule
			\end{tabular}
		\end{center}
		
		\caption{Comparison of exact and heuristic column generation approaches. \label{table:synthetic_ECG}}
	}
\end{table}

With regard to the training error, we can see that in some small cases where $d = 3$ and the number of assortments $M$ is small, the exact approach does deliver better performance (for example, $(N,M,d) = (4, 10, 3)$). However, in cases involving more products and assortments and/or deeper trees, the heuristic column generation approach is very close to the exact approach (see for example $(N,M,d) = (6, 50, 3)$). In those cases where the exact approach exhausts its time limit, the final forest produced by the exact approach typically achieves a significantly higher training error than the heuristic column generation approach. Comparing the heuristic column generation and the randomized tree sampling approach, we can see that when $d$ is 3 or 4, the randomized tree sampling approach tends to achieve a lower training error, while the heuristic column generation method tends to do better when $d = 5$. Overall, these results suggest that exact approaches to solving the estimation problem~\eqref{prob:master_primal} are difficult to deploy in practice, and that it is necessary to consider heuristic approaches. 

\clearpage
\newpage

\section{Leaf-Based Heuristic Column Generation Method}
\label{sec:appendix_leaf_HCG}

As mentioned in Section~\ref{subsec:estimation_by_column_generation}, one can consider an alternate form of the top-down induction heuristic in which the complexity control is formulated in terms of the number of leaves. In this version of the top-down heuristic, the main termination criterion (beside the reduced cost being locally optimal) is whether or not we reach a user-defined limit $L$ on the number of leaves. We formally define this version of the heuristic as Algorithm~\ref{alg:TDI_leaves}.

\begin{algorithm}
	{\SingleSpacedXI
		\caption{Leaf-based top-down induction method for heuristically solving column generation subproblem~\eqref{prob:CG_subproblem_abs}. \label{alg:TDI_leaves}}

		\begin{algorithmic}[1]
			\Procedure{TopDownInduction-Leaf}{$\alphab$, $\nu$, $L$}
			\State Initialize $t \gets t_0$
			\State Initialize $Z_c \gets \left[ - \sum_{S \in \Scal} \alphab_{S}^T \Ab_{t,S} - \nu \right]$
			\While{ $|\leaves(t)| < L$ }
			\State Compute $Z_{t, \ell, p, o_1, o_2}$ for all $\ell \in \leaves(t)$, $p \in \Ncal \, \setminus P(\ell) $, \par
			\hskip \algorithmicindent$o_1 \in \{p, 0\} \cup \{ x_s \mid s \in \mathbf{LS}(\ell) \}$, $o_2 \in \{0 \} \cup \{ x_s \mid s \in \mathbf{LS}(\ell) \}$
			\State Set $Z^* \gets \min_{\ell, p, o_1, o_2} Z_{t,\ell, p, o_1, o_2}$
			\State Set $(\ell^*, p^*, o^*_1, o^*_2) \gets \arg \min_{(\ell, p, o_1, o_2)} Z_{t,\ell, p, o_1, o_2}$
			\If{$Z^* < Z_c$}
			\State Set $Z_c \gets Z^*$
			\State Set $t \gets \textsc{GrowTree}(t, \ell^*, p^*, o^*_1, o^*_2)$
			\Else
			\State \textbf{break}
			\EndIf
			\EndWhile
			\State \Return $t$, $Z_c$
			\EndProcedure
		\end{algorithmic}
	}
\end{algorithm}

\clearpage
\newpage

\section{Additional Numerical Results with Real Customer Transaction Data}
\label{sec:appendix_IRI}

In this section, we aim to understand the value of warm-starting (Section~\ref{subsec:appendix_IRI_ass_coldVsWarm}),  performance of leaf-based heuristic column generation (Section~\ref{subsec:appendix_IRI_assLL}), performance of the randomized tree sampling with respect to different depths and forest sizes (Section~\ref{subsec:appendix_IRI_assRTS}), followed by the runtime and model sizes results for the decision forest model and benchmark models (Section~\ref{subsec:appendix_IRI_ass_runtime_modelsize}). For simplicity, all experiments follows the cross validation based on splitting the assortments in the IRI data set as in Section~\ref{subsec:experiment_IRI_asst}.

\subsection{Results for cold-started heuristic column generation}
\label{subsec:appendix_IRI_ass_coldVsWarm}

In this subsection, we aim to understand the value of warm-starting. We estimate the decision forest model using heuristic column generation in two ways: the first approach involves warm-starting the model using the independent demand model (see Figure~\ref{fig:independent_demand_model}) and the second approach involves warm-starting the model using the ranking-based model, estimated using the method of \cite{van2014market}. 

Table~\ref{table:IRI_ass_coldVsWarm} reports the performance of these two models, for values of the HCG depth limit $d$ in $\{3,4,5,6,7\}$. In most cases, warm-starting using the ranking-based model does lead to lower KL divergences than the independent demand model. Across all combinations of product category and $d$, warm-starting via the ranking-based model leads to an average improvement in the KL divergence of 1.47, suggesting that a good initial model can lead to appreciable improvements in the predictive performance of the decision forest model. 

\begin{table}
	{\SingleSpacedXI
		\begin{center}
			\begin{tabular}{lrrrrrrrrrr} \toprule
				& \multicolumn{2}{c}{$d = 3$} & \multicolumn{2}{c}{$d = 4$} & \multicolumn{2}{c}{$d = 5$} & \multicolumn{2}{c}{$d = 6$} & \multicolumn{2}{c}{$d = 7$} \\
				Product Category & ID & RM & ID & RM & ID & RM & ID & RM & ID & RM \\ \midrule
				Beer & 1.10 & 1.36 & 4.40 & 0.85 & 7.42 & 1.98 & 6.84 & 2.88 & 35.77 & 3.05 \\ 
				Blades & 0.80 & 0.43 & 0.57 & 0.36 & 0.89 & 0.49 & 1.08 & 0.83 & 1.17 & 1.18 \\ 
				Carbonated Beverages & 1.83 & 1.60 & 1.31 & 0.86 & 1.73 & 1.30 & 1.52 & 12.53 & 2.94 & 1.31 \\ 
				Cigarettes & 1.37 & 1.37 & 0.85 & 0.78 & 1.13 & 1.06 & 2.65 & 0.86 & 1.76 & 2.15 \\ 
				Coffee & 2.62 & 1.95 & 2.35 & 1.80 & 3.20 & 2.34 & 3.41 & 2.68 & 6.69 & 3.74 \\ 
				Cold Cereal & 0.90 & 0.93 & 1.33 & 0.75 & 2.20 & 0.82 & 6.62 & 1.20 & 1.77 & 1.12 \\ 
				Deodorant & 0.62 & 0.44 & 0.58 & 0.64 & 1.33 & 0.47 & 1.07 & 1.22 & 1.74 & 1.88 \\ 
				Diapers & 9.38 & 0.82 & 12.38 & 1.37 & 8.62 & 1.22 & 4.41 & 1.40 & 5.22 & 1.63 \\ 
				Facial Tissue & 1.67 & 0.78 & 1.31 & 0.75 & 1.28 & 0.76 & 4.21 & 1.05 & 1.89 & 0.96 \\ 
				Frozen Dinners & 1.46 & 1.76 & 2.77 & 2.37 & 5.99 & 4.12 & 16.86 & 3.34 & 18.25 & 2.34 \\ 
				Frozen Pizza & 1.56 & 1.15 & 1.38 & 0.93 & 1.08 & 1.05 & 1.68 & 1.47 & 3.52 & 1.55 \\ 
				Hotdogs & 2.98 & 3.06 & 2.83 & 2.81 & 2.67 & 2.55 & 2.61 & 2.88 & 3.08 & 4.35 \\ 
				Household Cleaners & 0.58 & 0.68 & 0.87 & 0.55 & 2.82 & 1.24 & 2.34 & 2.71 & 5.82 & 4.10 \\ 
				Laundry Detergent & 2.67 & 2.13 & 2.35 & 2.27 & 2.70 & 2.54 & 3.89 & 3.00 & 2.94 & 3.21 \\ 
				Margarine/Butter & 1.32 & 1.37 & 2.23 & 0.85 & 3.65 & 1.22 & 2.78 & 1.54 & 2.62 & 1.46 \\ 
				Mayonnaise & 1.06 & 0.84 & 0.99 & 0.83 & 1.16 & 0.93 & 1.46 & 0.88 & 2.40 & 1.00 \\ 
				Milk & 1.24 & 1.29 & 1.72 & 1.77 & 3.79 & 2.37 & 3.11 & 2.53 & 5.04 & 3.03 \\ 
				Mustard/Ketchup & 0.93 & 0.72 & 0.84 & 0.68 & 2.91 & 0.62 & 3.72 & 0.88 & 6.05 & 1.05 \\ 
				Paper Towels & 1.38 & 1.03 & 1.73 & 1.45 & 2.79 & 1.77 & 2.80 & 1.52 & 3.09 & 1.82 \\ 
				Peanut Butter & 1.92 & 1.49 & 1.73 & 3.09 & 1.52 & 4.85 & 2.87 & 1.72 & 2.55 & 1.82 \\ 
				Photo & 11.06 & 1.29 & 5.27 & 1.43 & 4.25 & 1.45 & 5.81 & 1.41 & 12.94 & 1.44 \\ 
				Salty Snacks & 1.82 & 1.72 & 1.88 & 1.76 & 2.85 & 1.66 & 2.69 & 1.89 & 5.94 & 2.34 \\ 
				Shampoo & 0.63 & 0.93 & 1.55 & 0.66 & 1.47 & 0.65 & 1.08 & 1.04 & 1.87 & 1.29 \\ 
				Soup & 1.40 & 0.96 & 1.51 & 1.61 & 1.55 & 1.10 & 2.67 & 1.39 & 2.86 & 1.59 \\ 
				Spaghetti/Italian Sauce & 2.86 & 2.69 & 4.12 & 3.93 & 3.52 & 2.40 & 8.33 & 3.44 & 4.41 & 3.88 \\ 
				Sugar Substitutes & 1.01 & 0.77 & 0.64 & 0.78 & 0.90 & 0.72 & 1.16 & 0.66 & 1.09 & 0.95 \\ 
				Toilet Tissue & 3.21 & 1.37 & 1.90 & 1.63 & 2.57 & 1.91 & 3.32 & 2.01 & 1.72 & 1.90 \\ 
				Toothbrush & 0.94 & 1.00 & 1.21 & 0.68 & 1.67 & 1.19 & 3.74 & 1.66 & 3.54 & 1.97 \\ 
				Toothpaste & 0.44 & 0.35 & 0.69 & 0.37 & 0.75 & 0.40 & 1.54 & 0.61 & 1.93 & 0.47 \\ 
				Yogurt & 2.44 & 2.78 & 3.36 & 2.87 & 4.23 & 6.42 & 5.92 & 4.85 & 6.42 & 3.80 \\ \bottomrule
			\end{tabular}
		\end{center}
		\caption{Out-of-sample KL divergence (in units of $10^{-2}$) for the decision forest model warm-started with the independent demand model (ID) and the ranking-based model (RM). \label{table:IRI_ass_coldVsWarm} }
	}
\end{table}

\subsection{Results for leaf-based heuristic column generation}
\label{subsec:appendix_IRI_assLL}

In this subsection, we evaluate the performance of our heuristic column generation approach using the leaf-based top-down induction method (Algorithm~\ref{alg:TDI_leaves} described in Section~\ref{sec:appendix_leaf_HCG}), as opposed to the depth-based top-down induction method (Algorithm~\ref{alg:TDI} described in Section~\ref{subsec:estimation_by_column_generation}). We run the heuristic column generation procedure with the leaf-based top-down induction method, with values for the leaf limit $L$ in $\{4,8,16,32,64\}$. We additionally run the heuristic column generation procedure with the depth-based top-down induction method, with values for the depth limit $d$ in $\{3,4,5,6,7\}$. Note that the values for $L$ are chosen to match the maximum number of leaves for each depth limit $d$; for example, when $d = 4$, the maximum number of leaves that a purchase decision tree may have is 8. We warm start both the depth-based and the leaf-based procedures with the ranking-based model found using the method of \cite{van2014market}.

Table~\ref{table:IRI_assLL} shows the KL divergence of each decision forest model for each product category, averaged over the five folds of each product category. (Each column labeled with $d = ...$ corresponds to the depth-based heuristic column generation method, while each column labeled with $L = ...$ corresponds to the leaf-based heuristic column generation method.)  In general, for a fixed value of $L$, the leaf-based heuristic column generation method attains roughly the same or slightly lower KL divergence than the depth $d$ that corresponds to that value of $L$. The reason for this difference is because the leaf-based procedure can select from a larger set of trees: specifically, a tree of maximum depth $d$ will have at most $2^{d-1}$ leaves, but a tree with at most $2^{d-1}$ leaves could have maximum depth greater than $d$. (For example, a ranking with a consideration set of size 7 corresponds to a tree with 8 leaves and a depth of 8, which is deeper than a balanced tree of depth 4.) In some cases, we observe that for higher values of $L$ the performance of the leaf-based method can deteriorate slightly relative to lower values of $L$. Overall, these results suggest that heuristic column generation with the leaf-based top-down induction method of Section~\ref{sec:appendix_leaf_HCG} is also a viable method for learning the decision forest model from data.

\begin{table}
	{\SingleSpacedXI
		\begin{tabular}{lcccccccccc} \toprule
			Product Category & $d = 3$ & $L = 4$ & $d = 4$ & $L = 8$ & $d = 5$ & $L = 16$ & $d = 6$ & $L = 32$ & $d = 7$ & $L = 64$ \\ \midrule
			Beer & 1.36 & 0.79 & 0.85 & 2.54 & 1.98 & 2.22 & 2.88 & 2.63 & 3.05 & 2.34 \\ 
			Blades & 0.43 & 0.39 & 0.36 & 0.68 & 0.49 & 1.05 & 0.83 & 0.71 & 1.18 & 0.74 \\ 
			Carbonated Beverages & 1.60 & 0.79 & 0.86 & 1.72 & 1.30 & 1.88 & 12.53 & 1.77 & 1.31 & 1.77 \\ 
			Cigarettes & 1.37 & 0.72 & 0.78 & 0.89 & 1.06 & 1.35 & 0.86 & 1.58 & 2.15 & 1.79 \\ 
			Coffee & 1.95 & 1.71 & 1.80 & 1.93 & 2.34 & 3.02 & 2.68 & 4.00 & 3.74 & 6.34 \\ 
			Cold Cereal & 0.93 & 0.67 & 0.75 & 0.93 & 0.82 & 1.39 & 1.20 & 1.22 & 1.12 & 1.22 \\ 
			Deodorant & 0.44 & 0.60 & 0.64 & 0.56 & 0.47 & 1.46 & 1.22 & 0.85 & 1.88 & 1.29 \\ 
			Diapers & 0.82 & 1.36 & 1.37 & 1.58 & 1.22 & 1.69 & 1.40 & 1.70 & 1.63 & 1.70 \\ 
			Facial Tissue & 0.78 & 0.73 & 0.75 & 0.84 & 0.76 & 0.98 & 1.05 & 1.16 & 0.96 & 1.15 \\ 
			Frozen Dinners & 1.76 & 1.81 & 2.37 & 3.54 & 4.12 & 2.88 & 3.34 & 1.90 & 2.34 & 1.87 \\ 
			Frozen Pizza & 1.15 & 1.01 & 0.93 & 1.17 & 1.05 & 1.53 & 1.47 & 1.66 & 1.55 & 1.94 \\ 
			Hotdogs & 3.06 & 2.80 & 2.81 & 2.39 & 2.55 & 2.63 & 2.88 & 3.31 & 4.35 & 4.28 \\ 
			Household Cleaners & 0.68 & 0.37 & 0.55 & 1.40 & 1.24 & 5.57 & 2.71 & 5.72 & 4.10 & 5.72 \\ 
			Laundry Detergent & 2.13 & 2.27 & 2.27 & 2.25 & 2.54 & 2.27 & 3.00 & 2.74 & 3.21 & 3.10 \\ 
			Margarine/Butter & 1.37 & 0.67 & 0.85 & 1.07 & 1.22 & 1.50 & 1.54 & 1.50 & 1.46 & 1.50 \\ 
			Mayonnaise & 0.84 & 0.82 & 0.83 & 0.89 & 0.93 & 0.94 & 0.88 & 0.93 & 1.00 & 1.01 \\ 
			Milk & 1.29 & 1.27 & 1.77 & 1.99 & 2.37 & 2.53 & 2.53 & 3.25 & 3.03 & 3.40 \\ 
			Mustard/Ketchup & 0.72 & 0.81 & 0.68 & 0.67 & 0.62 & 0.84 & 0.88 & 0.99 & 1.05 & 0.91 \\ 
			Paper Towels & 1.03 & 1.18 & 1.45 & 1.66 & 1.77 & 1.76 & 1.52 & 1.58 & 1.82 & 1.68 \\ 
			Peanut Butter & 1.49 & 1.55 & 3.09 & 2.46 & 4.85 & 1.69 & 1.72 & 1.68 & 1.82 & 1.65 \\ 
			Photo & 1.29 & 1.32 & 1.43 & 1.38 & 1.45 & 1.27 & 1.41 & 1.49 & 1.44 & 1.35 \\ 
			Salty Snacks & 1.72 & 1.69 & 1.76 & 1.67 & 1.66 & 1.84 & 1.89 & 1.98 & 2.34 & 2.13 \\ 
			Shampoo & 0.93 & 0.64 & 0.66 & 0.56 & 0.65 & 0.72 & 1.04 & 0.80 & 1.29 & 1.00 \\ 
			Soup & 0.96 & 1.39 & 1.61 & 1.23 & 1.10 & 1.48 & 1.39 & 1.66 & 1.59 & 1.66 \\ 
			Spaghetti/Italian Sauce & 2.69 & 3.12 & 3.93 & 3.02 & 2.40 & 3.32 & 3.44 & 2.86 & 3.88 & 3.09 \\ 
			Sugar Substitutes & 0.77 & 0.80 & 0.78 & 0.72 & 0.72 & 0.85 & 0.66 & 0.84 & 0.95 & 0.93 \\ 
			Toilet Tissue & 1.37 & 1.56 & 1.63 & 1.85 & 1.91 & 1.99 & 2.01 & 1.91 & 1.90 & 1.92 \\ 
			Toothbrush & 1.00 & 0.64 & 0.68 & 1.01 & 1.19 & 1.67 & 1.66 & 1.44 & 1.97 & 3.67 \\ 
			Toothpaste & 0.35 & 0.36 & 0.37 & 0.35 & 0.40 & 0.48 & 0.61 & 0.55 & 0.47 & 0.61 \\ 
			Yogurt & 2.78 & 2.31 & 2.87 & 3.53 & 6.42 & 3.88 & 4.85 & 4.34 & 3.80 & 4.19 \\ \bottomrule
		\end{tabular}
		\caption{ Comparison of leaf-based and depth-based heuristic column generation for the IRI data set. All values correspond to KL divergences (in units of $10^{-2}$), averaged over the five assortment folds. \label{table:IRI_assLL}}
	}
\end{table}

\subsection{Results for randomized tree sampling}
\label{subsec:appendix_IRI_assRTS}

In this additional experiment, we compare the randomized tree sampling approach from Section~\ref{subsec:estimation_by_sampling_approach} against the heuristic column generation approach. We test values of the depth $d$ in $\{3,4,5,6,7\}$ and we test different values of the number of sampled trees $K$ from the set $\{100, 200, 500, 1000, 2000\}$. For both the heuristic column generation and the randomized tree sampling approaches, we warm start them using the independent demand model (as in Figure~\ref{fig:independent_demand_model}). For the heuristic column generation approach, the value $d$ is used as the depth limit for the top-down induction procedure (Algorithm~\ref{alg:TDI}). For the randomized tree sampling approach, the base collection of trees $F$ is specified as the set of all balanced trees of depth $d$ that satisfy Requirements 1-3, and the distribution $\xib$ is chosen as the uniform distribution over $F$. 

In Table~\ref{table:IRI_assRTS}, we compare the performance of the heuristic column generation method and the randomized tree sampling method with $K = 2000$, as measured by the out-of-sample KL divergence, for different values of $d$. We can see that in general, the randomized tree sampling method is comparable in performance to the heuristic column generation approach; for larger values of $d$, HCG exhibits an edge over RTS. As an additional comparison, Table~\ref{table:IRI_assRTS_d} shows, for depth $d = 5$, how the performance of the RTS method varies as $K$ varies from 100 to 2000; for comparison, the performance of the HCG method is shown for the same depth. From this table, we can see that the out-of-sample performance of the RTS method improves as the number of sampled trees $K$ increases. Lastly, Table~\ref{table:IRI_assRTS_d_runtime} shows the runtime of the RTS method for different values of $K$ for depth $d = 5$, compared to heuristic column generation with $d = 5$. In the largest case, the runtime of the RTS method is no greater than about 3 minutes. In general, the runtime of the HCG method is smaller than the RTS method because the number of trees that the HCG method generates is small (on average about 270 trees). Comparing these runtime results to those for the $L_1$ estimation problem in Section~\ref{subsec:appendix_CG_runtime}, we note that the runtimes here for the randomized tree sampling method are larger. This is because the estimation problem here is solved using the EM algorithm; for a fixed number of trees, the runtime for EM is in general much larger than solving the $L_1$ estimation problem with Gurobi, as we do in Section~\ref{subsec:appendix_CG_runtime}. 

\begin{table}
	{\SingleSpacedXI
		\begin{center}
			\begin{tabular}{lrrrrrrrrrr}
				\toprule
				& \multicolumn{2}{c}{$d = 3$} & \multicolumn{2}{c}{$d = 4$} & \multicolumn{2}{c}{$d = 5$} & \multicolumn{2}{c}{$d = 6$} & \multicolumn{2}{c}{$d = 7$} \\
				Product Category & HCG & RTS & HCG & RTS & HCG & RTS & HCG & RTS & HCG & RTS \\ \midrule
				Beer & 1.10 & 1.38 & 4.40 & 1.67 & 7.42 & 0.80 & 6.84 & 1.60 & 35.77 & 3.30 \\ 
				Blades & 0.80 & 0.93 & 0.57 & 1.15 & 0.89 & 1.59 & 1.08 & 2.97 & 1.17 & 7.48 \\ 
				Carbonated Beverages & 1.83 & 2.14 & 1.31 & 1.18 & 1.73 & 1.44 & 1.52 & 1.84 & 2.94 & 3.50 \\ 
				Cigarettes & 1.37 & 1.21 & 0.85 & 1.12 & 1.13 & 1.59 & 2.65 & 2.71 & 1.76 & 11.91 \\ 
				Coffee & 2.62 & 2.45 & 2.35 & 1.84 & 3.20 & 1.72 & 3.41 & 2.30 & 6.69 & 3.39 \\ 
				Cold Cereal & 0.90 & 0.68 & 1.33 & 1.00 & 2.20 & 1.18 & 6.62 & 1.85 & 1.77 & 3.02 \\ 
				Deodorant & 0.62 & 0.75 & 0.58 & 0.84 & 1.33 & 1.12 & 1.07 & 1.07 & 1.74 & 1.99 \\ 
				Diapers & 9.38 & 1.61 & 12.38 & 2.37 & 8.62 & 3.09 & 4.41 & 4.97 & 5.22 & 3.14 \\ 
				Facial Tissue & 1.67 & 1.41 & 1.31 & 1.67 & 1.28 & 2.46 & 4.21 & 3.30 & 1.89 & 5.28 \\ 
				Frozen Dinners & 1.46 & 1.49 & 2.77 & 1.86 & 5.99 & 1.68 & 16.86 & 1.92 & 18.25 & 3.18 \\ 
				Frozen Pizza & 1.56 & 1.29 & 1.38 & 1.12 & 1.08 & 1.50 & 1.68 & 1.87 & 3.52 & 4.93 \\ 
				Hotdogs & 2.98 & 3.03 & 2.83 & 2.65 & 2.67 & 2.80 & 2.61 & 3.09 & 3.08 & 4.66 \\ 
				Household Cleaners & 0.58 & 0.55 & 0.87 & 0.92 & 2.82 & 1.15 & 2.34 & 1.57 & 5.82 & 1.79 \\ 
				Laundry Detergent & 2.67 & 2.42 & 2.35 & 2.23 & 2.70 & 2.43 & 3.89 & 3.98 & 2.94 & 9.43 \\ 
				Margarine/Butter & 1.32 & 0.83 & 2.23 & 1.05 & 3.65 & 1.66 & 2.78 & 2.74 & 2.62 & 5.01 \\ 
				Mayonnaise & 1.06 & 1.02 & 0.99 & 1.04 & 1.16 & 1.25 & 1.46 & 2.18 & 2.40 & 3.81 \\ 
				Milk & 1.24 & 1.30 & 1.72 & 1.31 & 3.79 & 1.50 & 3.11 & 2.27 & 5.04 & 4.49 \\ 
				Mustard/Ketchup & 0.93 & 0.89 & 0.84 & 0.76 & 2.91 & 0.90 & 3.72 & 1.17 & 6.05 & 2.26 \\ 
				Paper Towels & 1.38 & 1.37 & 1.73 & 1.58 & 2.79 & 1.88 & 2.80 & 2.72 & 3.09 & 3.60 \\ 
				Peanut Butter & 1.92 & 1.53 & 1.73 & 1.85 & 1.52 & 2.23 & 2.87 & 3.90 & 2.55 & 6.39 \\ 
				Photo & 11.06 & 8.70 & 5.27 & 6.77 & 4.25 & 7.79 & 5.81 & 9.84 & 12.94 & 15.99 \\ 
				Salty Snacks & 1.82 & 1.73 & 1.88 & 1.65 & 2.85 & 1.74 & 2.69 & 2.23 & 5.94 & 4.04 \\ 
				Shampoo & 0.63 & 1.09 & 1.55 & 1.22 & 1.47 & 1.41 & 1.08 & 1.42 & 1.87 & 1.54 \\ 
				Soup & 1.40 & 1.71 & 1.51 & 1.61 & 1.55 & 2.02 & 2.67 & 2.70 & 2.86 & 4.28 \\ 
				Spaghetti/Italian Sauce & 2.86 & 2.91 & 4.12 & 2.32 & 3.52 & 2.03 & 8.33 & 2.85 & 4.41 & 4.47 \\ 
				Sugar Substitutes & 1.01 & 1.09 & 0.64 & 0.82 & 0.90 & 1.11 & 1.16 & 2.05 & 1.09 & 3.83 \\ 
				Toilet Tissue & 3.21 & 1.84 & 1.90 & 2.34 & 2.57 & 3.06 & 3.32 & 3.83 & 1.72 & 5.01 \\ 
				Toothbrush & 0.94 & 0.90 & 1.21 & 0.88 & 1.67 & 1.24 & 3.74 & 1.29 & 3.54 & 1.81 \\ 
				Toothpaste & 0.44 & 0.42 & 0.69 & 0.57 & 0.75 & 0.70 & 1.54 & 1.57 & 1.93 & 4.24 \\ 
				Yogurt & 2.44 & 2.00 & 3.36 & 1.99 & 4.23 & 2.96 & 5.92 & 3.74 & 6.42 & 7.64 \\ \bottomrule
			\end{tabular}
		\end{center}
		\caption{Comparison of out-of-sample KL divergence (in units of $10^{-2}$) for heuristic column generation and randomized tree sampling with $K = 2000$ sampled trees, for different values of the depth $d$. \label{table:IRI_assRTS}}
	}
\end{table}

\begin{table}
	{\SingleSpacedXI
		
		\begin{center}
			\begin{tabular}{lrrrrrr} \toprule
				& HCG & RTS & RTS & RTS & RTS & RTS \\
				Product Category & & ($K = 100$) & ($K = 200$) & ($K = 500$) & ($K = 1000$) & ($K = 2000$) \\ \midrule
				Beer & 7.42 & 7.71 & 3.87 & 3.20 & 2.49 & 0.80 \\ 
				Blades & 0.89 & 16.40 & 6.53 & 2.85 & 1.49 & 1.59 \\ 
				Carbonated Beverages & 1.73 & 6.72 & 2.91 & 2.22 & 1.66 & 1.44 \\ 
				Cigarettes & 1.13 & 11.61 & 3.76 & 2.26 & 2.03 & 1.59 \\ 
				Coffee & 3.20 & 5.83 & 2.94 & 2.06 & 2.12 & 1.72 \\ 
				Cold Cereal & 2.20 & 5.72 & 2.14 & 1.36 & 1.52 & 1.18 \\ 
				Deodorant & 1.33 & 3.41 & 2.47 & 1.00 & 0.77 & 1.12 \\ 
				Diapers & 8.62 & 3.58 & 2.94 & 7.55 & 4.67 & 3.09 \\ 
				Facial Tissue & 1.28 & 6.88 & 5.79 & 3.51 & 2.24 & 2.46 \\ 
				Frozen Dinners & 5.99 & 2.92 & 3.32 & 2.89 & 2.44 & 1.68 \\ 
				Frozen Pizza & 1.08 & 7.90 & 3.96 & 1.94 & 1.65 & 1.50 \\ 
				Hotdogs & 2.67 & 4.92 & 4.56 & 3.23 & 2.83 & 2.80 \\ 
				Household Cleaners & 2.82 & 2.78 & 0.57 & 0.63 & 1.56 & 1.15 \\ 
				Laundry Detergent & 2.70 & 6.98 & 9.72 & 3.27 & 3.35 & 2.43 \\ 
				Margarine/Butter & 3.65 & 4.14 & 3.81 & 2.45 & 1.58 & 1.66 \\ 
				Mayonnaise & 1.16 & 6.35 & 3.06 & 1.53 & 1.82 & 1.25 \\ 
				Milk & 3.79 & 4.63 & 2.63 & 2.14 & 2.04 & 1.50 \\ 
				Mustard/Ketchup & 2.91 & 4.52 & 1.92 & 1.15 & 1.04 & 0.90 \\ 
				Paper Towels & 2.79 & 3.12 & 3.49 & 2.58 & 2.47 & 1.88 \\ 
				Peanut Butter & 1.52 & 13.80 & 6.06 & 3.70 & 2.14 & 2.23 \\ 
				Photo & 4.25 & 18.91 & 17.59 & 12.59 & 8.55 & 7.79 \\ 
				Salty Snacks & 2.85 & 5.98 & 3.43 & 2.05 & 1.72 & 1.74 \\ 
				Shampoo & 1.47 & 1.88 & 0.85 & 1.80 & 1.15 & 1.41 \\ 
				Soup & 1.55 & 4.49 & 2.49 & 2.56 & 2.63 & 2.02 \\ 
				Spaghetti/Italian Sauce & 3.52 & 3.77 & 4.41 & 2.60 & 2.47 & 2.03 \\ 
				Sugar Substitutes & 0.90 & 4.49 & 6.23 & 1.28 & 1.12 & 1.11 \\ 
				Toilet Tissue & 2.57 & 4.46 & 4.40 & 3.32 & 3.33 & 3.06 \\ 
				Toothbrush & 1.67 & 2.68 & 2.05 & 1.64 & 1.40 & 1.24 \\ 
				Toothpaste & 0.75 & 6.48 & 1.49 & 1.43 & 0.75 & 0.70 \\ 
				Yogurt & 4.23 & 10.33 & 3.78 & 3.69 & 2.45 & 2.96 \\ \bottomrule
			\end{tabular}
		\end{center}
		\caption{Comparison of out-of-sample KL divergence (in units of $10^{-2}$) for heuristic column generation and randomized tree sampling for varying values of $K$ and depth $d = 5$. \label{table:IRI_assRTS_d}}
	}
\end{table}

\begin{table}
	{\SingleSpacedXI
		
		\begin{center}
			\begin{tabular}{lrrrrrr} \toprule
				& HCG & RTS & RTS & RTS & RTS & RTS \\
				Product Category & & ($K = 100$) & ($K = 200$) & ($K = 500$) & ($K = 1000$) & ($K = 2000$) \\ \midrule
				Beer & 3.9 & 1.9 & 3.9 & 9.8 & 27.8 & 71.1 \\ 
				Blades & 4.0 & 1.8 & 4.8 & 17.6 & 70.7 & 211.3 \\ 
				Carbonated Beverages & 1.5 & 1.0 & 2.5 & 5.7 & 12.3 & 36.6 \\ 
				Cigarettes & 3.2 & 1.6 & 5.0 & 20.1 & 71.1 & 158.4 \\ 
				Coffee & 3.2 & 1.4 & 3.6 & 9.0 & 22.3 & 61.9 \\ 
				Cold Cereal & 0.5 & 0.3 & 0.5 & 1.0 & 2.3 & 6.2 \\ 
				Deodorant & 2.7 & 1.5 & 2.0 & 4.7 & 10.8 & 34.0 \\ 
				Diapers & 2.9 & 1.0 & 2.5 & 6.3 & 14.2 & 33.1 \\ 
				Facial Tissue & 2.0 & 1.6 & 3.8 & 12.1 & 32.7 & 99.5 \\ 
				Frozen Dinners & 2.0 & 0.6 & 1.1 & 2.1 & 5.3 & 17.6 \\ 
				Frozen Pizza & 2.7 & 1.3 & 3.6 & 10.6 & 34.5 & 89.3 \\ 
				Hotdogs & 6.1 & 2.4 & 5.5 & 17.7 & 63.3 & 118.3 \\ 
				Household Cleaners & 0.8 & 0.3 & 0.3 & 0.7 & 1.3 & 4.8 \\ 
				Laundry Detergent & 4.8 & 2.0 & 4.3 & 16.6 & 56.9 & 154.6 \\ 
				Margarine/Butter & 0.9 & 0.3 & 0.5 & 1.2 & 2.5 & 6.0 \\ 
				Mayonnaise & 2.4 & 1.2 & 2.8 & 8.3 & 18.9 & 86.2 \\ 
				Milk & 3.4 & 1.5 & 3.2 & 7.6 & 17.8 & 50.6 \\ 
				Mustard/Ketchup & 1.6 & 1.0 & 2.2 & 5.3 & 12.6 & 63.5 \\ 
				Paper Towels & 2.4 & 1.2 & 3.2 & 9.8 & 18.9 & 63.5 \\ 
				Peanut Butter & 2.2 & 1.1 & 2.6 & 8.9 & 24.1 & 101.1 \\ 
				Photo & 4.5 & 1.5 & 4.2 & 15.3 & 84.1 & 182.3 \\ 
				Salty Snacks & 2.1 & 1.2 & 2.9 & 8.6 & 19.2 & 56.2 \\ 
				Shampoo & 1.7 & 0.9 & 1.7 & 4.2 & 11.7 & 39.1 \\ 
				Soup & 1.2 & 0.4 & 0.9 & 1.9 & 4.9 & 14.8 \\ 
				Spaghetti/Italian Sauce & 5.0 & 1.6 & 3.5 & 10.1 & 22.3 & 68.3 \\ 
				Sugar Substitutes & 3.0 & 1.7 & 4.3 & 16.0 & 44.9 & 100.1 \\ 
				Toilet Tissue & 2.1 & 0.7 & 1.6 & 3.8 & 8.2 & 30.4 \\ 
				Toothbrush & 4.9 & 1.8 & 3.6 & 10.6 & 39.0 & 69.9 \\ 
				Toothpaste & 0.7 & 0.9 & 2.2 & 5.3 & 12.6 & 33.7 \\ 
				Yogurt & 3.6 & 1.1 & 3.2 & 10.3 & 26.2 & 76.9 \\ \bottomrule
			\end{tabular}
		\end{center}
		\caption{Comparison of runtime (in seconds) for heuristic column generation and randomized tree sampling for varying values of $K$ and depth $d = 5$. For each product category and method, the runtime is averaged over the five assortment folds of the data. \label{table:IRI_assRTS_d_runtime}}
	}
\end{table}

\subsection{Runtime and model size results}
\label{subsec:appendix_IRI_ass_runtime_modelsize}

Table~\ref{table:IRI_ass_runtime} shows the average runtime over the five folds for each of the methods. To simplify the exposition, we focus on the LC-MNL, ranking and decision forest models (see Section~\ref{subsec:experiment_IRI_asst} for the details of the estimation for each model); for the ordinary MNL and HALO-MNL models, the average runtime over all product categories was less than 0.01 and 2 seconds, respectively. Note that for the LC-MNL, ranking and DF models, this time includes the time to perform $k$-fold cross-validation in order to tune the number of classes, the maximum consideration set size and the depth, respectively. For the DF model, we also note that the runtime includes the time required to estimate the ranking model as a warm start. 

\begin{table}
	{\SingleSpacedXI
		\centering 
		\begin{tabular}{lrrrrrr} \toprule
			Product Category & $|\Scal|$ & $|\Tcal|$ & LC-MNL & RM & DF & DF \\ 
			& & & & & (HCG) & (RTS) \\ \midrule
			Beer & 55 & 380,932 & 577.8 & 4054.8 & 207.6 & 385.2 \\ 
			Blades & 57 & 92,404 & 651.9 & 3725.8 & 112.5 & 531.5 \\ 
			Carbonated Beverages & 31 & 721,506 & 680.6 & 1826.4 & 88.6 & 165.4 \\ 
			Cigarettes & 68 & 249,668 & 1206.3 & 5188.8 & 211.5 & 544.1 \\ 
			Coffee & 47 & 372,536 & 1365.3 & 2508.7 & 159.2 & 282.8 \\ 
			Cold Cereal & 15 & 577,236 & 347.3 & 613.1 & 23.3 & 48.2 \\ 
			Deodorant & 45 & 271,286 & 144.4 & 3366.2 & 124.4 & 204.3 \\ 
			Diapers & 18 & 143,055 & 353.4 & 452.5 & 45.1 & 112.0 \\ 
			Facial Tissue & 43 & 73,806 & 867.0 & 2155.8 & 107.6 & 547.8 \\ 
			Frozen Dinners & 30 & 979,936 & 348.9 & 1765.3 & 88.0 & 128.1 \\ 
			Frozen Pizza & 61 & 292,878 & 1648.3 & 4081.2 & 257.9 & 529.8 \\ 
			Hotdogs & 100 & 101,624 & 2013.7 & 5733.0 & 361.0 & 640.6 \\ 
			Household Cleaners & 19 & 282,981 & 286.5 & 684.2 & 32.8 & 52.6 \\ 
			Laundry Detergent & 56 & 238,163 & 1500.3 & 5136.1 & 240.6 & 614.5 \\ 
			Margarine/Butter & 18 & 140,969 & 585.2 & 1563.2 & 37.6 & 58.8 \\ 
			Mayonnaise & 48 & 97,282 & 741.0 & 2308.6 & 111.3 & 367.3 \\ 
			Milk & 49 & 240,691 & 1568.0 & 2570.2 & 139.8 & 258.9 \\ 
			Mustard/Ketchup & 44 & 134,800 & 872.3 & 2565.3 & 116.1 & 295.8 \\ 
			Paper Towels & 40 & 82,636 & 701.2 & 2605.3 & 130.0 & 284.1 \\ 
			Peanut Butter & 51 & 108,770 & 1109.4 & 1839.3 & 109.1 & 392.3 \\ 
			Photo & 80 & 17,047 & 999.6 & 3298.5 & 109.3 & 507.9 \\ 
			Salty Snacks & 39 & 736,148 & 1047.4 & 2501.0 & 114.9 & 241.6 \\ 
			Shampoo & 66 & 290,429 & 313.7 & 3638.0 & 200.9 & 363.2 \\ 
			Soup & 24 & 905,541 & 337.7 & 1507.7 & 62.4 & 114.0 \\ 
			Spaghetti/Italian Sauce & 38 & 276,860 & 1144.6 & 3581.3 & 188.5 & 273.6 \\ 
			Sugar Substitutes & 64 & 53,834 & 841.3 & 3816.8 & 184.3 & 511.0 \\ 
			Toilet Tissue & 27 & 112,788 & 534.7 & 2333.7 & 98.2 & 185.9 \\ 
			Toothbrush & 114 & 197,676 & 1013.9 & 13652.5 & 670.2 & 850.9 \\ 
			Toothpaste & 42 & 238,271 & 273.0 & 1516.2 & 40.2 & 173.1 \\ 
			Yogurt & 43 & 499,203 & 1493.8 & 3274.7 & 145.0 & 310.7 \\ \midrule
			(Mean) &   &  & 852.3 & 3128.8 & 150.6 & 332.5 \\ 
			(Median) &   &  & 791.2 & 2567.7 & 115.5 & 290.0 \\ 
			(Maximum) &   &  & 2013.7 & 13652.5 & 670.2 & 850.9 \\ \bottomrule
		\end{tabular}
		\caption{Runtime (in seconds) for the estimation of each predictive model for each of the thirty product categories in the IRI data set. Each runtime is the average value over the five folds.  \label{table:IRI_ass_runtime}}
	}
\end{table}

From this table, we can see that the ranking-based model requires the most amount of time -- on average 3128 seconds (almost one hour) -- due to the use of $k$-fold cross-validation to tune the maximum consideration set size and the use of integer programming to solve the subproblem at each step of the algorithm of \cite{van2014market}. The LC-MNL model is the second highest, requiring just under 15 minutes on average, due to the EM algorithm which requires the estimation problem for MNL to be solved repeatedly and the use of $k$-fold cross validation. The decision forest model requires on average 150 seconds using the HCG approach and on average 333 seconds using the RTS approach. Note that as mentioned earlier, this includes the time needed to estimate the ranking-based model and for the cross-validation to choose the depth $d$ for the decision forest model (the depth limit for the HCG approach, or the depth of the base forest for the RTS approach). The main takeaway from these results is that the estimation of the decision forest model can be accomplished with manageable computation times. 

Lastly, we also compare the size of the models. Table~\ref{table:IRI_ass_modelsize} reports several metrics of model size for the LC-MNL, ranking and decision forest models. For the LC-MNL model, we report the average number of segments $K$ chosen by cross-validation; for the ranking-based model, we report the average number of rankings $K$ and the average maximum consideration set size chosen by cross-validation; and for the decision forest models, we report the average number of trees $K$, the average cross-validated depth $d$ (either the depth limit for HCG or the depth of the base forest for RTS) and the average number of leaves per tree $L$. From this table, we can see that the number of trees in the decision forest model obtained via the heuristic column generation method is comparable to the number of rankings in the ranking-based model. (For the randomized tree sampling method, the number of trees is slightly over 2000, as this number includes the 2000 trees that are randomly sampled, and the additional rankings that were used for warm-starting.) For the decision forest model, the model size varies by product category. For the HCG-based decision forest model, the average cross-validated depth can be as low as 3.0 and as high as 5.0, and the average number of leaves per tree varies from 4.0 to 9.5. For the RTS-based decision forest model, the average cross-validated depth similarly ranges between 3.0 and 4.4, while the average number of leaves per tree varies from 4.1 to 18.2 (note that the number of leaves is generally larger than it is for HCG, as the base forest was specified as the collection of balanced trees of the chosen depth, whereas the HCG method is allowed to estimate unbalanced trees).

\begin{table}
	{\SingleSpacedXI
		
		\centering
		\begin{tabular}{lccccccccc} \toprule
			Product Category & LC-MNL & \multicolumn{2}{c}{--- RM ---} & \multicolumn{3}{c}{--- DF (HCG) ---} & \multicolumn{3}{c}{--- DF (RTS) ---}   \\ 
			& $K$ & $K$ & $\CSsize$ & $K$ & $d$ & $L$ & $K$ & $d$ & $L$ \\ \midrule
			Beer & 8.8 & 202.0 & 3.4 & 331.6 & 3.8 & 7.7 & 2208.0 & 3.8 & 8.1 \\ 
			Blades & 8.4 & 139.2 & 4.4 & 218.6 & 4.2 & 8.6 & 2185.2 & 3.6 & 9.6 \\ 
			Carbonated Beverages & 10.0 & 118.4 & 2.4 & 274.2 & 3.8 & 5.6 & 2123.2 & 4.2 & 11.7 \\ 
			Cigarettes & 12.0 & 162.2 & 4.4 & 252.2 & 3.8 & 6.8 & 2169.8 & 3.6 & 6.5 \\ 
			Coffee & 13.0 & 142.4 & 3.4 & 298.4 & 3.2 & 5.6 & 2145.4 & 4.2 & 9.5 \\ 
			Cold Cereal & 4.0 & 70.0 & 2.8 & 149.2 & 3.2 & 5.7 & 2072.8 & 3.2 & 4.9 \\ 
			Deodorant & 6.2 & 150.8 & 4.6 & 200.8 & 4.2 & 9.5 & 2148.0 & 3.0 & 4.4 \\ 
			Diapers & 5.0 & 140.8 & 6.2 & 219.0 & 3.6 & 6.8 & 2148.2 & 3.0 & 4.3 \\ 
			Facial Tissue & 13.0 & 258.4 & 4.2 & 319.2 & 3.4 & 7.8 & 2278.4 & 3.2 & 5.2 \\ 
			Frozen Dinners & 5.4 & 110.0 & 2.4 & 230.4 & 3.8 & 7.3 & 2111.4 & 4.2 & 17.8 \\ 
			Frozen Pizza & 14.0 & 210.0 & 5.4 & 351.4 & 3.6 & 7.0 & 2216.0 & 3.2 & 5.1 \\ 
			Household Cleaners & 4.6 & 65.0 & 4.8 & 144.8 & 3.2 & 4.6 & 2065.0 & 3.2 & 4.8 \\ 
			Hotdogs & 11.6 & 152.4 & 3.8 & 449.8 & 3.4 & 5.5 & 2145.4 & 4.4 & 18.2 \\ 
			Laundry Detergent & 7.0 & 157.8 & 5.2 & 284.6 & 3.0 & 6.0 & 2156.4 & 3.6 & 9.4 \\ 
			Margarine/Butter & 9.8 & 168.8 & 2.8 & 240.8 & 4.4 & 7.6 & 2173.8 & 3.2 & 5.0 \\ 
			Mayonnaise & 9.4 & 191.0 & 4.4 & 263.0 & 3.0 & 6.6 & 2187.6 & 3.0 & 4.3 \\ 
			Milk & 6.2 & 89.2 & 2.0 & 254.8 & 3.0 & 4.0 & 2116.8 & 3.2 & 4.8 \\ 
			Mustard/Ketchup & 9.6 & 170.8 & 3.4 & 321.2 & 3.6 & 6.5 & 2176.8 & 3.4 & 5.8 \\ 
			Paper Towels & 5.6 & 266.6 & 3.0 & 291.4 & 3.6 & 7.7 & 2244.8 & 3.0 & 4.4 \\ 
			Peanut Butter & 9.4 & 134.0 & 5.0 & 204.4 & 3.0 & 6.5 & 2141.2 & 3.2 & 5.0 \\ 
			Photo & 2.8 & 164.6 & 7.8 & 184.0 & 3.2 & 8.0 & 2159.8 & 3.4 & 5.8 \\ 
			Salty Snacks & 7.8 & 125.8 & 3.2 & 276.2 & 4.0 & 6.6 & 2127.0 & 3.6 & 7.1 \\ 
			Shampoo & 6.6 & 206.8 & 4.8 & 288.0 & 4.0 & 9.0 & 2208.4 & 4.0 & 11.0 \\ 
			Soup & 8.2 & 102.2 & 5.8 & 223.0 & 3.6 & 6.7 & 2103.4 & 3.4 & 5.7 \\ 
			Spaghetti/Italian Sauce & 6.2 & 213.4 & 3.6 & 414.4 & 3.4 & 6.4 & 2213.2 & 4.2 & 12.3 \\ 
			Sugar Substitutes & 11.0 & 170.2 & 4.0 & 244.8 & 3.6 & 7.7 & 2190.6 & 3.4 & 5.8 \\ 
			Toilet Tissue & 4.6 & 245.0 & 2.4 & 310.0 & 5.0 & 8.6 & 2231.4 & 3.0 & 4.4 \\ 
			Toothbrush & 10.6 & 294.4 & 2.8 & 429.2 & 4.2 & 9.0 & 2303.6 & 3.8 & 10.1 \\ 
			Toothpaste & 5.4 & 84.0 & 3.0 & 133.4 & 3.0 & 6.5 & 2083.0 & 3.0 & 4.2 \\ 
			Yogurt & 14.0 & 167.2 & 4.8 & 267.6 & 3.2 & 5.4 & 2144.0 & 3.0 & 4.2 \\ \bottomrule
		\end{tabular}
		\caption{   Model size metrics for the LC-MNL, ranking and decision forest models for each of the thirty product categories in the IRI data set. In the column headings, $K$ denotes either the number of segments (for LC-MNL) or the number of trees/rankings (for the ranking and forest models); for the decision forest models, $d$ denotes the depth chosen via cross-validation and $L$ denotes the average number of leaves in the forest. For each model and category, the value is the average over the five folds. \label{table:IRI_ass_modelsize}}
	}
\end{table}

\clearpage
\newpage

\section{Temporal splitting}
\label{sec:appendix_IRI_temporal}

In this section, we compare the decision forest model against the other benchmarks using the experimental scheme described in Section~\ref{subsec:experiment_IRI_temporal}, in which split the data by time. In particular, we use the transaction data in the IRI Academic Data Set in the following way. We treat the transactions corresponding to the first two weeks of data in 2007 as training data for all of the methods. We then used the next four weeks (weeks 3 to 6) as testing data. We note that this type of splitting approach has been used previously in the literature; see, for example, \cite{alptekinoglu2016exponomial} and \cite{aouad2018exponomial}. 

We compare the decision forest model (estimated using both the HCG and RTS approaches), the single-class MNL model, the LC-MNL model, the ranking-based model and the HALO-MNL model. We use five-fold cross-validation to tune the models, where the folds correspond to the five assortment folds used in Section~\ref{subsec:experiment_IRI_asst}. We use this cross-validation to tune the values of the following hyperparameters: the depth limit $d$ for the HCG approach for the decision forest model; the depth $d$ of the base forest for the RTS approach for the decision forest model; the number of classes $K$ for the latent-class MNL model; and the consideration set size $\CSsize$ for the ranking-based model. We estimate all of the models using the same methods as in Sections~\ref{subsec:experiment_IRI_asst}. As in the earlier experiments, we set the number of sampled trees $K$ for the RTS method to 2000. Note that unlike the assortment-splitting experiment in Section~\ref{subsec:experiment_IRI_asst}, there is only one form of cross-validation done in this experiment, which is to tune the hyperparameters; the out-of-sample performance of the final model of each class is then evaluated using the testing data for weeks 3 to 6, without any further cross-validation. 

Table~\ref{table:R2_temporal_splitting} shows the out-of-sample KL divergence for each of the five models, on each of the 30 product categories. In addition, the table also summarizes the number of transactions and unique assortments in the training data ($|\Tcal_{1:2}|$ and $|\Scal_{1:2}|$, respectively, where the subscript 1:2 indicates weeks 1 to 2); the number of transactions and unique assortments in the test data ($|\Tcal_{3:6}|$ and $|\Scal_{3:6}|$, respectively, where the subscript 3:6 indicates weeks 3 to 6); and lastly, how many new assortments exist in the test data (i.e., how many assortments in the test data are not present in the training data; this is indicated by $|\Scal_{3:6} \setminus \Scal_{1:2}|$ ).

\begin{sidewaystable}
	\vspace{45em}
	\centering
	\begin{tabular}{lrrrrrcccccc}
		\toprule
		Category & $|\Scal_{1:2}|$ & $|\Tcal_{1:2}|$ & $|\Scal_{3:6}|$ & $|\Tcal_{3:6}|$ & $| \Scal_{3:6} \setminus \Scal_{1:2}|$ & MNL & HALO-MNL & LC-MNL & RM & DF & DF \\ 
		& & & & & & & & & & (HCG) & (RTS) \\ \midrule
		Beer & 55 & 380,932 & 57 & 772,629 & 11 & 0.83 & 0.10 & 0.66 & 0.66 & \bfseries 0.09 & 0.21 \\ 
		Blades & 57 & 92,404 & 74 & 183,906 & 23 & 0.42 & \bfseries 0.18 & 0.26 & 0.34 & 0.29 & 0.35 \\ 
		Carbonated Beverages & 31 & 721,506 & 32 & 1,452,900 & 1 & 1.57 & 0.25 & 1.25 & 1.18 & 0.29 & \bfseries 0.24 \\ 
		Cigarettes & 68 & 249,668 & 87 & 499,296 & 25 & 1.62 & 0.39 & 1.26 & 1.08 & \bfseries 0.31 & 0.42 \\ 
		Coffee & 47 & 372,536 & 50 & 759,294 & 10 & 2.22 & 0.54 & 1.23 & 1.14 & \bfseries 0.26 & 0.45 \\ 
		Cold Cereal & 15 & 577,236 & 17 & 1,163,581 & 2 & 1.41 & 0.09 & 1.34 & 1.18 & \bfseries 0.09 & 0.20 \\ 
		Deodorant & 45 & 271,286 & 66 & 543,687 & 36 & 0.37 & \bfseries 0.31 & 0.36 & 0.35 & 0.33 & 0.37 \\ 
		Diapers & 18 & 143,055 & 19 & 284,854 & 3 & 0.15 & \bfseries 0.06 & 0.07 & 0.09 & 0.09 & 0.19 \\ 
		Facial Tissue & 43 & 73,806 & 46 & 146,165 & 4 & 1.03 & \bfseries 0.29 & 0.64 & 0.64 & 0.36 & 0.40 \\ 
		Frozen Dinners & 30 & 979,936 & 29 & 1,972,097 & 3 & 0.46 & \bfseries 0.14 & 0.36 & 0.32 & 0.18 & 0.26 \\ 
		Frozen Pizza & 61 & 292,878 & 70 & 584,388 & 14 & 2.50 & 0.44 & 1.62 & 1.68 & \bfseries 0.35 & 0.67 \\ 
		Hotdogs & 100 & 101,624 & 112 & 203,573 & 23 & 2.83 & 1.32 & 2.13 & 1.89 & \bfseries 0.33 & 0.84 \\ 
		Household Cleaners & 19 & 282,981 & 28 & 561,774 & 13 & 0.07 & \bfseries 0.05 & 0.07 & 0.06 & 0.05 & 0.06 \\ 
		Laundry Detergent & 56 & 238,163 & 73 & 459,055 & 23 & 1.78 & \bfseries 0.66 & 1.46 & 0.97 & 0.71 & 0.79 \\ 
		Margarine/Butter & 18 & 140,969 & 19 & 283,612 & 5 & 1.18 & \bfseries 0.08 & 0.70 & 0.70 &  0.08 & 0.20 \\ 
		Mayonnaise & 48 & 97,282 & 49 & 194,607 & 7 & 1.19 & 0.33 & 0.59 & 0.56 & \bfseries 0.28 & 0.44 \\ 
		Milk & 49 & 240,691 & 56 & 472,009 & 9 & 3.38 & \bfseries 0.67 & 1.87 & 1.86 & 0.71 & 0.72 \\ 
		Mustard/Ketchup & 44 & 134,800 & 57 & 265,669 & 16 & 1.13 & 0.23 & 0.71 & 0.65 & \bfseries 0.12 & 0.28 \\ 
		Paper Towels & 40 & 82,636 & 49 & 164,366 & 15 & 0.84 & \bfseries 0.33 & 0.41 & 0.46 & 0.36 & 0.48 \\ 
		Peanut Butter & 51 & 108,770 & 53 & 220,186 & 7 & 1.90 & \bfseries 0.56 & 1.23 & 1.17 & 0.81 & 0.81 \\ 
		Photo & 80 & 17,047 & 91 & 33,769 & 18 & 0.74 & 0.90 & 0.45 & 0.54 & \bfseries 0.40 & 0.43 \\ 
		Salty Snacks & 39 & 736,148 & 39 & 1,491,938 & 3 & 1.40 & 0.50 & 0.85 & 0.83 & \bfseries 0.09 & 0.33 \\ 
		Shampoo & 66 & 290,429 & 82 & 568,503 & 28 & 0.44 & \bfseries 0.16 & 0.38 & 0.36 & 0.16 & 0.26 \\ 
		Soup & 24 & 905,541 & 29 & 1,819,666 & 5 & 0.55 & \bfseries 0.10 & 0.40 & 0.29 & 0.23 & 0.27 \\ 
		Spaghetti/Italian Sauce & 38 & 276,860 & 40 & 550,463 & 7 & 2.51 & 0.73 & 1.83 & 1.80 & \bfseries 0.22 & 0.50 \\ 
		Sugar Substitutes & 64 & 53,834 & 69 & 108,422 & 8 & 0.53 & 0.32 & 0.36 & 0.38 & \bfseries 0.15 & 0.28 \\ 
		Toilet Tissue & 27 & 112,788 & 34 & 226,410 & 8 & 1.01 & \bfseries 0.30 & 0.70 & 0.64 & 0.45 & 0.46 \\ 
		Toothbrush & 114 & 197,676 & 143 & 390,089 & 52 & 0.58 & 0.24 & 0.37 & 0.35 & \bfseries 0.22 & 0.37 \\ 
		Toothpaste & 42 & 238,271 & 48 & 474,519 & 8 & 0.42 & \bfseries 0.15 & 0.39 & 0.39 & 0.18 & 0.25 \\ 
		Yogurt & 43 & 499,203 & 49 & 1,028,179 & 9 & 3.98 & \bfseries 0.79 & 3.12 & 2.33 & 1.40 & 0.88 \\  \midrule
		(Mean) & -- & -- & -- & -- & -- & 1.30 & 0.37 & 0.90 & 0.83 & \bfseries 0.32 & 0.41 \\ 
		(Median) & -- & -- & -- & -- & -- & 1.08 & 0.31 & 0.68 & 0.65 & \bfseries 0.27 & 0.37 \\ 
		(Maximum) & -- & -- & -- & -- & -- & 3.98 & 1.32 & 3.12 & 2.33 & 1.40 & \bfseries 0.88 \\ \bottomrule
	\end{tabular}
	\caption{Comparison of out-of-sample KL divergence (in units of $10^{-2}$) for the temporal splitting experiment. \label{table:R2_temporal_splitting}}
\end{sidewaystable}

From this table, we can see that the decision forest model and the HALO-MNL model are essentially tied for the best performance. The decision forest model, using either the HCG or RTS methods, delivers the lowest KL divergence out of all of the models on 14 out of 30 product categories, while the HALO-MNL provides the lowest KL divergence out of all the models on 16 out of 30 categories. At the same time, when comparing the average, median and maximum out-of-sample KL divergence over the 30 product categories, we can see that the decision forest model achieves the best performance, with the HALO-MNL model being very slightly higher. Overall, this experiment indicates the potential of the decision forest model to be used for making predictions prospectively. 

\clearpage
\newpage

\section{Additional Numerical Results with Synthetic Data}
\label{sec:appendix_synthetic}

In this section, we complement the numerical results on real transaction data reported in Section~\ref{sec:experiment_IRI} with numerical results using synthetic data.

For the two experimental setups that we will describe shortly, we consider a set of $N = 9$ candidate products. We consider four classes of models for the ground truth choice model: the LC-MNL, HALO-MNL, ranking-based and decision forest models. We randomly generate five ground truth models for each model class, as follows:
\begin{enumerate}
	
	\item \textbf{HALO-MNL model}: The HALO-MNL model is characterized by utility parameters $u_1,\dots,u_{N-1}$ and pairwise utility parameters $w_{i,j}$ for $i,j \in \{1,\dots,N-1\}, i \neq j$. We generate each ground truth model by drawing each $u_i$ from an independent Uniform(0,1) distribution and each $w_{i,j}$ from an independent Uniform(-0.5,0.5) distribution. 
	\item \textbf{LC-MNL model}: The LC-MNL model with $K$ customer classes is characterized by $K$ probabilities $p_1,\dots, p_K$ and utility parameters $u_{k,1}, \dots, u_{k,N-1}$. We generate each ground truth model by setting $K = 5$, drawing each utility $u_{k,i}$ independently from a Uniform(0,1) distribution, and drawing the probability vector $(p_1,\dots,p_K)$ from the uniform distribution on the $(K-1)$-dimensional unit simplex.
	\item \textbf{Ranking-based model}: The ranking-based model is characterized by a set of rankings $\Sigma$ and the associated probability distribution $\lambdab$, as described in Section~\ref{subsec:model_ranking}. We generate each ground truth model by independently generating $K = 50$ rankings uniformly at random, and drawing the probability distribution $\lambdab$ from the uniform distribution on the $(K-1)$-dimensional unit simplex.
	\item \textbf{Decision forest model}: The decision forest model is characterized by a set of trees $F$ and the associated probability distribution $\lambdab$. We generate each ground truth model by independently generating $K = 50$ trees uniformly at random from the set of all balanced trees of depth $d = 4$, and drawing the probability distribution $\lambdab$ from the uniform distribution on the $(K-1)$-dimensional unit simplex.
\end{enumerate}

We examine the performance of each predictive model in two experiments. In the first experiment (Section~\ref{subsec:appendix_synthetic_ass_kfold}), we generate a dataset of synthetic transactions according to each ground truth model and split the data by assortments. In the second experiment (Section~\ref{subsec:appendix_synthetic_L1}), we consider the problem of learning the ground truth choice model when the true choice probabilities are known for a limited set of assortments.

\subsection{Out-of-sample performance under assortment-based splitting}
\label{subsec:appendix_synthetic_ass_kfold}

We randomly generate $|\Scal| = 50$ assortments for each of the 20 ground truth choice models described previously. We generate $|\Tcal| = 100,000$ transactions, where the assortment for each transaction is randomly chosen from the 50 assortments. We divide the set of assortments $\Scal$ into five sets $\Scal_1, \dots, \Scal_5$ consisting of 10 assortments each, and define $\Tcal_1,\dots \Tcal_5$ to be the sets of transactions corresponding to $\Scal_1,\dots, \Scal_5$ respectively. We then evaluate each predictive model using 5-fold cross validation, where for a given $i \in \{1,\dots, 5\}$, we use the transaction sets $\Tcal_1,\dots, \Tcal_{i-1}, \Tcal_{i+1}, \dots, \Tcal_5$ for training, and use the transaction set $\Tcal_i$ for testing. We estimate the LC-MNL model with $K = 5$, the HALO-MNL model, the ranking-based model, and the decision forest model with depth $d = 4$ using the HCG method. We warm start the decision forest model using the ranking-based model.

Table~\ref{table:synthetic_ass_kfold} shows the out-of-sample KL divergence of each predictive model for each of the 20 data sets. The bottom-most row reports the worst-case KL divergence of each predictive model over the 20 ground truth models. We find that the decision forest model is able to obtain good performance even when the ground truth model belongs to the RUM class. For example, when the ground truth model is the LC-MNL model, the KL divergence of the fitted LC-MNL model ranges from 0.15 to 0.19, whereas the decision forest model is only slightly higher, ranging from 0.37 - 0.43. We also find that decision forest model attains a lower worst-case KL divergence than the other models, suggesting that the decision forest model has the potential in some cases to offer better predictive performance in the presence of model misspecification. 

\begin{table}   
	\SingleSpacedXI
	\centering
	\begin{tabular}{lccccc} \toprule
		Ground Truth & Instance & LC-MNL & HALO-MNL & RM & DF \\ \midrule
		LC-MNL & 1 & 0.19 & 0.28 & 0.51 & 0.37 \\ 
		LC-MNL & 2 & 0.19 & 0.33 & 0.61 & 0.42 \\ 
		LC-MNL & 3 & 0.19 & 0.22 & 0.61 & 0.43 \\ 
		LC-MNL & 4 & 0.15 & 0.23 & 0.64 & 0.39 \\ 
		LC-MNL & 5 & 0.16 & 0.20 & 0.58 & 0.40 \\ \midrule
		HALO-MNL & 1 & 4.30 & 0.19 & 2.31 & 1.57 \\ 
		HALO-MNL & 2 & 4.71 & 0.22 & 2.43 & 1.99 \\ 
		HALO-MNL & 3 & 4.07 & 0.29 & 2.78 & 1.66 \\ 
		HALO-MNL & 4 & 4.61 & 0.26 & 3.09 & 2.45 \\ 
		HALO-MNL & 5 & 4.66 & 0.22 & 3.11 & 2.04 \\ \midrule
		RM & 1 & 1.79 & 1.81 & 0.89 & 0.78 \\ 
		RM & 2 & 2.22 & 1.49 & 0.80 & 0.81 \\ 
		RM & 3 & 1.06 & 1.31 & 0.85 & 0.74 \\ 
		RM & 4 & 2.07 & 1.55 & 0.97 & 0.83 \\ 
		RM & 5 & 1.44 & 1.45 & 0.97 & 0.81 \\ \midrule
		DF & 1 & 5.95 & 4.14 & 4.47 & 2.67 \\ 
		DF & 2 & 5.20 & 5.14 & 4.85 & 2.79 \\ 
		DF & 3 & 5.76 & 3.29 & 4.53 & 2.67 \\ 
		DF & 4 & 6.08 & 6.58 & 5.08 & 4.53 \\ 
		DF & 5 & 5.42 & 5.32 & 5.29 & 2.80 \\ \midrule
		(Maximum) &  & 6.08 & 6.58 & 5.29 & 4.53 \\ \bottomrule
	\end{tabular}
	
	\caption{   Out-of-sample KL divergence (in units of $10^{-2}$) for the four predictive models and the 20 ground truth models, under assortment-based splitting. For each ground truth model and method, the reported value is the average over the five folds. \label{table:synthetic_ass_kfold}}
\end{table}

\subsection{Out-of-sample performance with known choice probabilities}
\label{subsec:appendix_synthetic_L1}

In our second experiment, we consider the same ground truth models as in our previous experiment and for each ground truth model, we consider the same set of $M = 50$ assortments used in the preceding experiment. For a fixed ground truth model and each assortment $S \in \Scal = \{S_1, \dots, S_M\}$, we compute the purchase distribution $\vb_S$ under the ground truth model. We then use the collection of assortments and purchase distributions $(S, \vb_S)_{S \in \Scal}$ as training data to estimate each predictive model.

We use two metrics to examine the performance of predictive models. The first metric is the $L_1$ error between the predicted and true purchase probabilities. We define the $L_1$ error as
\begin{align*}
\frac{1}{2^N}\cdot \sum_{ S \subseteq \Ncal } \left\Vert   \hat{\vb}_S - \vb_S \right\Vert_1,
\end{align*}
where $\hat{\vb}_S$ is the predicted purchase probability distribution for assortment $S$; in words, it is the $L_1$ norm of the difference between the predicted and actual purchase distributions, averaged over all $2^N$ possible assortments. A value of zero for the $L_1$ error implies a perfect recovery of the ground truth choice probabilities. 

The second metric is the revenue gap. Let $r_i$ be the marginal revenue of product $i$, and let $\rb = (r_0, r_1,\dots, r_N)$ be the vector of revenues, where $r_i$ is the revenue of option $i \in \Ncal^+$ (note that for the no-purchase option, $r_0 = 0$). Let $S^* \in \arg \max_{S \subseteq \Ncal} \rb^T \vb_{S^*}$ be an optimal assortment under the ground truth choice model, and let $\hat{S} \in \arg \max_{S \subseteq \Ncal} \rb^T \hat{\vb}_S$ be an optimal assortment under the estimated choice model. We define the revenue gap as 
\begin{align}
\frac{ \rb^T \vb_{\hat{S}}}{ \rb^T \vb_{S^*}}. \label{eq:approx_rate}
\end{align}
In \eqref{eq:approx_rate}, the numerator is the expected revenue of the optimal assortment returned by the estimated/learned model and the denominator is the true optimal expected revenue; both expected revenues are calculated with respect to the ground truth model. This metric quantifies how useful the model is for the purpose of deriving an optimal or near-optimal assortment; a value of one for the approximation rate implies that the learned model leads to optimal assortments. For simplicity, in our experiments we set $r_i = 100 - 10i$ for each $i \in \Ncal$, so that $\rb = (0, 90, 80, \dots, 10)$ for each ground truth model. 

We estimate both the decision forest model and the ranking-based model by minimizing the average $L_1$ error as the objective. We learn the decision forest model by solving the linear optimization problem~\eqref{prob:master_primal} described in Section~\ref{sec:model_estimation_methods}. We learn the ranking-based model using the heuristic column generation method described in \cite{misic2016data}; we use this alternate method, as opposed to the likelihood-based method of \cite{van2014market}, as it is tailored to directly minimizing the $L_1$ error. To the best of our knowledge, there do not exist standard methods for estimating the HALO-MNL and LC-MNL models so as to minimize the $L_1$ error of the predicted choice probabilities; thus, we continue to use likelihood based methods (for the HALO-MNL model we solve the maximum likelihood estimation problem directly as a concave optimization problem, while for the LC-MNL model we use the EM algorithm).

Table~\ref{table:synthetic_L1} shows the $L_1$ error of each predictive model on each of the 20 ground truth models, while Table~\ref{table:synthetic_gap} show the revenue gap. From Table~\ref{table:synthetic_L1}, we see again that the DF model still attains good performance when the ground truth model is an RUM model, and we again see that the decision forest model achieves a better worst-case predictive performance, as measured by the maximum $L_1$ error, than the other three models. Table~\ref{table:synthetic_gap} shows similar behavior with respect to the revenue gap: the decision forest model leads to assortments that in the worst case achieve 94\% of the optimal revenue, whereas for the other models, the worst-case revenue gap ranges from 68.4\% to 72.1\%.

\begin{table}[ht]
	\SingleSpacedXI
	
	\centering
	\begin{tabular}{lccccc} \toprule
		Ground Truth & Instance & LC-MNL & HALO-MNL & RM & DF \\ \midrule
		LC-MNL & 1 & 1.56 & 0.13 & 3.37 & 3.37 \\ 
		LC-MNL & 2 & 1.74 & 0.13 & 3.30 & 3.30 \\ 
		LC-MNL & 3 & 1.96 & 0.24 & 3.25 & 3.25 \\ 
		LC-MNL & 4 & 1.79 & 0.16 & 3.37 & 3.37 \\ 
		LC-MNL & 5 & 1.78 & 0.15 & 3.32 & 3.32 \\ \midrule
		HALO-MNL & 1 & 21.26 & 0.00 & 16.32 & 11.48 \\ 
		HALO-MNL & 2 & 23.03 & 0.00 & 15.94 & 12.09 \\ 
		HALO-MNL & 3 & 24.07 & 0.00 & 18.85 & 14.36 \\ 
		HALO-MNL & 4 & 24.58 & 0.00 & 17.37 & 12.94 \\ 
		HALO-MNL & 5 & 21.54 & 0.00 & 16.54 & 11.31 \\ \midrule
		RM & 1 & 11.39 & 10.78 & 2.86 & 2.87 \\ 
		RM & 2 & 12.31 & 9.29 & 2.82 & 2.92 \\ 
		RM & 3 & 10.64 & 9.09 & 3.66 & 3.81 \\ 
		RM & 4 & 12.62 & 10.29 & 4.78 & 4.69 \\ 
		RM & 5 & 11.41 & 9.50 & 4.16 & 4.20 \\ \midrule
		DF & 1 & 23.27 & 17.30 & 18.28 & 8.37 \\ 
		DF & 2 & 22.61 & 16.69 & 20.72 & 10.21 \\ 
		DF & 3 & 20.42 & 15.31 & 18.61 & 8.10 \\ 
		DF & 4 & 21.79 & 18.77 & 19.30 & 10.61 \\ 
		DF & 5 & 22.66 & 17.45 & 23.34 & 8.68 \\ \midrule
		(Maximum) &  & 24.58 & 18.77 & 23.34 & 14.36 \\ \bottomrule
	\end{tabular}
	\caption{    $L_1$ error (in units of $10^{-2}$) for each predictive model and each ground truth model. \label{table:synthetic_L1}}
\end{table}

\begin{table}
	\SingleSpacedXI
	
	\centering
	\begin{tabular}{lccccc} \toprule
		Ground Truth & Instance & LC-MNL & HALO-MNL & RM & DF \\ \midrule
		LC-MNL & 1 & 100.0 & 100.0 & 100.0 & 100.0 \\ 
		LC-MNL & 2 & 100.0 & 100.0 & 100.0 & 100.0 \\ 
		LC-MNL & 3 & 100.0 & 100.0 & 100.0 & 100.0 \\ 
		LC-MNL & 4 & 100.0 & 100.0 & 100.0 & 100.0 \\ 
		LC-MNL & 5 & 100.0 & 100.0 & 100.0 & 100.0 \\ \midrule
		HALO-MNL & 1 & 100.0 & 100.0 & 100.0 & 94.9 \\ 
		HALO-MNL & 2 & 100.0 & 100.0 & 100.0 & 97.3 \\ 
		HALO-MNL & 3 & 100.0 & 98.3 & 100.0 & 94.2 \\ 
		HALO-MNL & 4 & 97.2 & 100.0 & 100.0 & 97.2 \\ 
		HALO-MNL & 5 & 100.0 & 100.0 & 100.0 & 100.0 \\ \midrule
		RM & 1 & 97.9 & 100.0 & 98.2 & 98.2 \\ 
		RM & 2 & 96.5 & 99.5 & 100.0 & 100.0 \\ 
		RM & 3 & 100.0 & 96.2 & 100.0 & 100.0 \\ 
		RM & 4 & 95.1 & 97.1 & 95.1 & 95.1 \\ 
		RM & 5 & 100.0 & 99.3 & 99.3 & 99.3 \\ \midrule
		DF & 1 & 98.3 & 86.2 & 98.3 & 95.9 \\ 
		DF & 2 & 72.1 & 72.1 & 87.8 & 100.0 \\ 
		DF & 3 & 83.8 & 81.4 & 77.0 & 100.0 \\ 
		DF & 4 & 81.7 & 79.0 & 81.2 & 99.9 \\ 
		DF & 5 & 80.7 & 69.1 & 68.4 & 97.3 \\ \midrule
		(Minimum) &  & 72.1 & 69.1 & 68.4 & 94.2 \\ \bottomrule
	\end{tabular}
	\caption{   Revenue gap (in \%) for each predictive model and each ground truth model. \label{table:synthetic_gap}}
\end{table}

\end{document}